\documentclass{article}

\usepackage{arxiv}

\usepackage[utf8]{inputenc} 
\usepackage[T1]{fontenc}    
\usepackage{hyperref}       
\usepackage{url}            
\usepackage{booktabs}       
\usepackage{amsfonts}       
\usepackage{nicefrac}       
\usepackage{microtype}      
\usepackage{lipsum}		
\usepackage{graphicx}
\usepackage{natbib}
\usepackage{doi}
\usepackage{bbding}
\usepackage{array}
 \usepackage{array,multirow,graphicx}

\usepackage{amsmath}
\usepackage{diagbox}
\usepackage{color}
\usepackage{bm}
\usepackage{array}
\usepackage[utf8]{inputenc}
\usepackage{graphicx}
\usepackage{float}
\usepackage{amsmath}
\usepackage{caption}
\usepackage{subcaption}
\usepackage{multirow}

\usepackage{color}
\usepackage{booktabs}
\usepackage[ruled,vlined]{algorithm2e}
\usepackage{makecell}
\newcolumntype{I}{!{\vrule width 1.5pt}}
\usepackage{booktabs}

\usepackage{graphicx}  
\usepackage{soul}
\usepackage{subcaption}
\usepackage{mwe}
\usepackage{xcolor}
\usepackage{amssymb}

\usepackage{tikz}
\definecolor{fc}{HTML}{1E90FF}
\tikzset{fc/.style={black,draw=black,fill=fc,rectangle,minimum height=1cm}}
\definecolor{h}{HTML}{228B22}
\definecolor{bias}{HTML}{87CEFA}
\tikzset{h/.style={black,draw=black,fill=h,rectangle,minimum height=1cm}}
\tikzset{bias/.style={black,draw=black,fill=bias,rectangle,minimum height=1cm}}
\definecolor{anti-flashwhite}{rgb}{0.95, 0.95, 0.96}
\definecolor{almond}{rgb}{0.98, 0.91, 0.71}

\title{Physics-Informed Geometry-Aware  Neural Operator}

\author{
  Weiheng Zhong \\
  Department of Civil and Environmental Engineering\\
  University of Illinois at Urbana-Champaign\\
  Champaign, Illinois\\
  \texttt{weiheng4@illinois.edu} \\
   \And
 Hadi Meidani \\
  Department of Civil and Environmental Engineering\\
  University of Illinois at Urbana-Champaign\\
  Champaign, Illinois \\
  \texttt{meidani@illinois.edu} \\
}
 
\begin{document}
\maketitle

\begin{abstract}
Engineering design problems often involve solving parametric Partial Differential Equations (PDEs) under variable PDE parameters and domain geometry. Recently, neural operators have shown promise in  learning  PDE operators and quickly predicting the PDE solutions. However, training these neural operators typically requires large datasets, the acquisition of which can be prohibitively expensive. To overcome this, physics-informed training offers an alternative way of building  neural operators, eliminating the high computational costs associated with Finite Element generation of training data. Nevertheless, current physics-informed neural operators struggle with limitations, either in handling varying domain geometries or varying PDE parameters. In this research, we introduce a novel method, the Physics-Informed Geometry-Aware Neural Operator (PI-GANO), designed to simultaneously generalize across both PDE parameters and domain geometries.  We adopt a geometry encoder to capture the domain geometry features, and design a novel pipeline to integrate this component within the existing Deep Compositional Operator Network architecture. Numerical results  demonstrate the accuracy and efficiency of the proposed method.  {All the  codes and data related to this work are available on GitHub: \textcolor{blue}{\href{https://github.com/WeihengZ/Physics-informed-Neural-Foundation-Operator}{https://github.com/WeihengZ/Physics-informed-Neural-Foundation-Operator}}}.
\end{abstract}

\keywords{Physics-informed deep learning, Neural operator, Geometry generalization}

\section{Introduction \label{Sec.Intro}}

 {The finite element method (FEM) as the predominant high fidelity numerical approach for solving partial differential equations (PDEs) involves discretizing a continuous function space using a discrete mesh and solving a high-dimensional linear system~\cite{fem}. The cost associated with this solution  can be prohibitively large particularly  in tasks requiring repetitive solutions.} An example of such tasks is engineering design which necessitates solving parameterized PDEs over a wide range of PDE parameters and domain geometries for design evaluation \cite{mechanic_design}.

Recently, machine learning techniques have been introduced to accelerate the process of solving PDEs by learning a neural operator as a mapping from variable PDE parameters and/or domain geometry to the PDE solution \cite{DON}. Once a neural operator model is successfully trained on a dataset, it can generalize to new, unseen parameters and domain geometries. This is done by a single forward pass through the trained neural network, with minimal computational cost. 

Research on neural operators initially was focused on developing various neural operator architectures for operator learning of parametric PDEs in a data-driven way \cite{DON, FNO, NO, GNO}. As the first neural operator architecture, Deep Operator Networks (DeepONet) \cite{DON} utilize the universal approximation theorem for operators, introducing an architecture that approximates nonlinear operators through learning a collection of basis functions and coefficients. The Fourier Neural Operator (FNO) \cite{FNO} leverages the Fourier transform to model mappings in the spectral domain, capturing global dependencies and showing superior performance in square domain shapes. Geo-FNO \cite{geo_FNO} extends FNO to irregular meshes by learning a mapping from an irregular mesh to a uniform mesh. Inspired by FNO, Wavelet Neural Operator (WNO) \cite{WNO} is also proposed by using wavelet transform instead of Fourier transform to better handle signals with discontinuity and spikes in an irregular domain geometry. While each of these methods has demonstrated potential in specific applications, a common limitation is that each trained model is restricted to a specific domain geometry.

To overcome the limitations of these ``geometry-specific" methods, a number of ``geometry aware" methods have been developed. The Graph Neural Operator (GNO) \cite{GNO} employs graph neural networks (GNNs)  for operator learning by treating inputs and outputs as graphs, enabling generalization across different domain geometries. Similarly, PointNet \cite{PointNet} facilitates PDE operator learning across diverse geometries by using a set of collocation points to represent the domain geometry. As an extension of FNO, geometry-aware FNO \cite{geo_aware_FNO} was  proposed for PDE solution prediction using a signed distance function (SDF) and point-cloud representations of the geometry. Diffeomorphism Neural Operator \cite{diff_NO} combines harmonic mapping approach and the FNO to offer geometry generalization ability to FNO. Additionally, the General Neural Operator Transformer (GNOT) \cite{GNOT} encodes the PDE parameters and domain geometries into tokens and adopt the architecture of Transformer \cite{attn} for PDE solution predictions. Using the same idea of GNOT, other transformer-based methods \cite{transolver, pde_transformer, point_transformer} were also proposed to address PDEs in varying geometrical settings.

Even though these methods offer the ability of generalizing to various PDE parameters and geometries, they all are data-intensive \cite{PI-deepOnet}. To bypass the need for training datasets, which are typically expensive to obtain, physics-informed neural operators, which are inspired by physics-informed neural networks (PINNs) \cite{pinn}, have emerged as an effective approach. This approach  integrates the governing PDEs directly into the training process, making it possible to develop a neural operator without any  training data obtained from high fidelity simulations, such as FEM.
 Among these methods, Physics-informed DeepONet (PI-DeepONet) \cite{PI-deepOnet} extends the original DeepONet framework by embedding physical laws directly into the loss function with automatic differentiation \cite{AD} during training, which can only be applied on a specific mesh. Physics-informed Fourier Neural Operator (PI-FNO) \cite{PI-FNO} builds upon the original FNO framework, offering generalization across different PDE parameter representations in a square domain while also incorporating physics-informed training methods. Similarly, physics-informed Wavelet Neural Operator (PI-WNO) \cite{PI-WNO} is also proposed based on the architecture of WNO, implementing physics-informed training for WNO \cite{WNO} using stochastic projection. Physics-informed Deep Compositional Operator Network (PI-DCON) \cite{PIDCON} adopts a pooling layer to capture the global features of the variable PDE parameters,  {which enables the learning of the PDE operator} on any given irregular domain geometry. However, all these physics-informed methods struggle with generalization to varying geometries. A significant advancement to address this limitation is the Physics-informed PointNet (PI-PointNet) \cite{PI_PointNet}, which integrates a physics-informed training algorithm into PointNet. Nonetheless, PI-PointNet lacks the capability to generalize across PDE parameters, which limits its application in tackling  complex problems in engineering design which also involves varying boundary conditions, etc.

In this study, we tackle the aforementioned challenges by introducing an innovative framework, called Physics-informed Geometry-aware Neural Operator (PI-GANO). This model is inspired by PI-DCON and PI-PointNet and is capable of generalizing across different PDE parameter representations and domain geometry representations simultaneously, including those in irregular domain shapes. The differences between our proposed model versus other existing works are summarized in Table \ref{table.archi_compare}. To the best of our knowledge, our proposed model is the first attempt to develop a neural operator that can simultaneously handle varying PDE parameters and domain geometries without need for any training data.  


\begin{table*}[ht]
\begin{center}
\caption{Differences of Physics-informed Geometry-aware Neural Operator versus other existing models.}
\begin{tabular}{c c c c}
\hline
{} & Data-free & \makecell{Generalize across PDE \\ parameter representations} & \makecell{Generalize across \\ geometries} \\
DeepONet & -- & -- & -- \\
FNO , WNO  , LRNO , geo-FNO  & -- & \CheckmarkBold & --  \\
GNO , GNOT, GA-FNO, Diff-NO  & -- & \CheckmarkBold & \CheckmarkBold  \\
PI-DeepONet & \CheckmarkBold & -- & -- \\
PI-FNO, PI-WNO, PI-DCON & \CheckmarkBold & \CheckmarkBold & -- \\
PI-PointNet & \CheckmarkBold & -- & \CheckmarkBold \\
PI-GANO & \CheckmarkBold & \CheckmarkBold & \CheckmarkBold \\
\hline
\label{table.archi_compare}
\end{tabular}
\end{center}
\end{table*}

The remainder of this paper is organized as follows. In Section~\ref{sec.bg}, we  briefly introduce the problem settings and  technical backgrounds for PI-DeepONet and PI-PointNet. Section~\ref{sec.methodology} introduces our model architecture and explains the 
key points about it. Finally, numerical experiments and detailed performance evaluation of the proposed methods and conclusions are included in Sections~\ref{sec.results} and \ref{sec.conclusions}.

\section{Technical background \label{sec.bg}}

\subsection{Problem Setting \label{Subsec.PS}}

Our goal is to develop an efficient machine learning-based solver for parametric PDEs  which are formulated by: 
\begin{equation}
    \begin{aligned}
            \mathcal{N}_{\bm{x}} [u(\bm{x}), k(\bm{x})] &= 0, \quad &&\bm{x} \in \Omega, \\
    \mathcal{B}_{\bm{x}}[u(\bm{x})] &= g(\bm{x}), \quad &&\bm{x} \in \partial \Omega,
\label{eq.pde}
    \end{aligned}
\end{equation}
where $\Omega$ is a  physical domain in $R^d$, $\bm x$ is a  $d$-dimensional vector of spatial coordinates, $\mathcal{N}_{\bm x}$ is a general differential operator, and $\mathcal{B}_{\bm{x}}$ is a boundary conditiond operator acting on the domain boundary $\partial \Omega$.  Also, $k(\bm{x})$ refers to the parameters of the PDE, which can include the  coefficients and forcing terms in the governing equation, $g(\bm{x})$  denotes the boundary conditions, and  $u(\bm{x})$ is the solution of this PDE for a given set of parameters and boundary conditions. 

We seek to solve this PDE across different parameters and  geometries. Let  $\bm{\Omega} = \{\Omega_1, \Omega_2, ..., \Omega_n\}$ denote the set of variable geometries. For each geometry $\Omega_i$, we have a corresponding set of PDE parameters, i.e. coefficients denoted by $k_i(\bm x)$ and boundary conditions denoted by $g_i(\bm x)$. Therefore, we employ a neural network model to approximate the operator $\mathcal{M}$ defined by:
\begin{equation}
    \mathcal{M}: \{k_i(\bm{x}), g_i(\bm{x}), \Omega_i\} \rightarrow u_i(\bm{x}), \quad \bm{x} \in \Omega_i, \quad \Omega_i \in \bm{\Omega},
\label{eq.geo_op}
\end{equation}
where the inputs to the operator include information on both PDE parameters and domain geometries. Existing physics-informed neural operators primarily focus on learning the PDE operator with variations in either the PDE parameters or  geometry. Our proposed framework handles both of these variation types. This is done by using two model ingredients: (1) one  based on PI-DeepONet to handle  varying PDE parameters and (2) one inspired by PI-PointNet to encode  varying domain geometries. In the next two sections, we present the background for PI-DeepONet and PI-PointNet. 

\subsection{Parameter encoding with physics-informed DeepONet \label{PI-DeepONet}}

DeepONet was proposed to solve parametric PDEs by approximating a domain-specific operator $\mathcal{M}$ given by
\begin{equation}
    \mathcal{M}: \{k_i(\bm{x}),g_i(\bm{x})\} \rightarrow u_i(\bm{x}), \quad \bm{x} \in \Omega
\label{eq.deepOnet_op}
\end{equation}
where $\Omega$ is a given domain geometry. Specifically, DeepONet approximates $\mathcal{M}$ by a neural network model $\mathcal{U}_{\theta}(\bm{x}, k(\bm{x}'), g(\bm{x}''))$, where $\bm{x}$ denotes  the coordinates at which the solution $u$ is to be calculated, and $\bm{x}'$ and $\bm{x}''$ denote the coordinates at which the values of coefficients and boundary conditions, respectively, are available. 

\begin{figure}[ht]
    \centering
    \includegraphics[width=6cm]{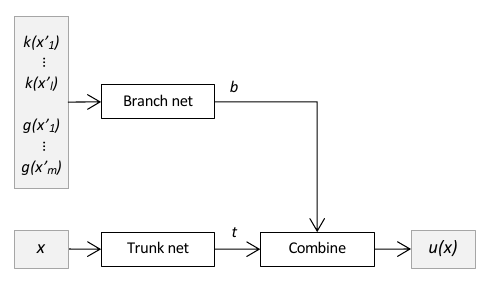}
    \caption{\footnotesize The architecture of the DeepONet is shown. The input to the branch net is the sampled discrete representations of variable parameters $k(\bm x)$  and $g(\bm x)$, with $l$ and $m$ discretization points, respectively.}
    \label{fig.don}
\end{figure}

As shown in Figure ~\ref{fig.don}, the architecture of DeepONet is composed of two separate neural networks referred to as the ``branch net" and ``trunk net", respectively. Both the branch net and trunk net are simply multilayer perceptrons (MLP).  The DeepONet prediction is given by
\begin{equation}
    \mathcal{U}_{\theta}(\bm{x}, k(\bm{x}'), g(\bm{x}'')) = \sum_i^q b_i t_i.
    \label{eq.DeepONet}
\end{equation}
where the $q$-dimensional features embedding $\bm{b}=[b_1,b_2,..., b_q]$ is the output of the branch net, and the $q$-dimensional coordinate embedding  $\bm{t} = [t_1,t_2,..., t_q]$ is the output of the trunk neural networks. To calculate $\bm t$, the trunk net takes the coordinates of a collocation point, $\bm{x}$, as input. To calculate $\bm b$, the branch net takes $[k(\bm{x}'_1), k(\bm{x}'_2), ..., k(\bm{x}'_l), g(\bm{x}''_1), g(\bm{x}''_2), \cdots, g(\bm{x}''_m)]$ as input. This input is the parameter function $k(\bm{x}')$ evaluated at a collection of sampled locations $\{\bm{x}'_j\}_{j=1}^l$ and the boundary function $g(\bm{x}'')$ evaluated at a generally different collection of sampled locations $\{\bm{x}''_j\}_{j=1}^m$. 

In order to set up the training loss for the model training, we first obtain   $N$  realizations of branch net input, where each realization is parameter function $k(\cdot)$ obtained at  locations $\bm{x}'$ and  BC function $g(\cdot)$ obtained at locations $\bm{x}''$. Then the physics-informed training loss of the neural operator $\mathcal{L}_{\text{PINO}}$ is given by:
\begin{equation}
    \mathcal{L}_{\text{PINO}} = \mathcal{L}_{\text{PDE}} + \alpha \mathcal{L}_{\text{BC}}
\label{eq.pino_loss}
\end{equation}
where
\begin{equation}
\begin{aligned}
    \mathcal{L}_{\text{PDE}} &= \frac{1}{N} \sum_i^N (\mathcal{N}_{\bm{x}} [\mathcal{U}_{\theta}(\bm{x}, k_i(\bm{x}'), g_i(\bm{x}'')), k_i(\bm{x}')])^2, \quad &&\bm{x} \in \Omega, \\
    \mathcal{L}_{\text{BC}} &= \frac{1}{N} \sum_i^N (\mathcal{B}_{\bm{x}}[\mathcal{U}_{\theta}(\bm{x}, k_i(\bm{x}'), g_i(\bm{x}''))] - g_i(\bm{x}''))^2, \quad &&\bm{x} \in \partial \Omega,
\end{aligned}
\end{equation}
where $\alpha$ is the trade-off coefficient between the PDE residual loss term and the initial and boundary condition loss term. The optimal neural network parameters $\theta$ are found by minimizing the total training loss $\mathcal{L}_{\text{PINO}}$ with exact derivatives computed using automatic differentiation \cite{AD}. After the model training, for any new realization of the parameters, the well-trained DeepONet can predict the corresponding solution directly in a specific domain geometry $\Omega$. To ensure the efficiency of model training, we implement stochastic gradient descent algorithm to update model parameters by random sampling of collocation points in each epoch.

\subsection{Geometry encoding with physics-informed PointNet \label{Sec.PI-PointNet}}

PI-PointNet focus on approximating the PDE operator across a set of domain geometries. In its original form, it does not handle varying PDE parameters. Specifically, let the operator $\mathcal{M}$  be given by
\begin{equation}
    \mathcal{M}: \{\Omega_i\} \rightarrow u_i(\bm{x}), \quad \bm{x} \in \Omega_i, \quad \Omega_i \in \bm{\Omega}.
\label{eq.PointNet_op}
\end{equation}
Each domain geometry $\Omega_i$ is represented by a set of $s_i$ collocation points inside the domain, denoted by $\{\bm{x}_j\}_{j=1}^{s_i}$. PI-PointNet takes these coordinates as input, and predicts the PDE solution   on the input collocation points $\{u(\bm{x}_i)\}_{i=1}^{s_i}$. The model parameters are estimated  in a physics-informed training using the same loss function shown in Equation \ref{eq.pino_loss}.

The architecture of the PointNet consists of two multilayer perceptrons $\mathcal{U}_1(\cdot,\theta_1)$ and  $\mathcal{U}_2(\cdot,\theta_2)$, where $\theta_1$ and $\theta_2$ are trainable parameters. For each geometry input $\Omega_i$, it first generates the hidden embeddings $\bm{h}$ of the input collocation points by:
\begin{equation}
    \{\bm{h}_j\}_{j=1}^{s_i} = \{\mathcal{U}_1 (\bm{x}_j, \theta_1)\}_{j=1}^{s_i}.
\end{equation}
Then a Maxpooling layer \cite{maxpooling} is applied to hidden embeddings to obtain a global feature $\bm{G}^i$ of the $i$-th geometry:
\begin{equation}
     \bm{G}^i = \text{Maxpool}(\{\bm{h}_j\}_{j=1}^{s_i}),
\end{equation}
where $\text{Maxpool}$ refers to the max-pooling operation which calculates the maximum value in each embedding dimension over all the $s_i$ points. The global feature is then concatenated to each of the hidden embeddings, providing a unique embedding $\bm H^i_j$ of coordinate $\bm{x}_j$ in the $i$-th geometry by
\begin{equation}
     \bm{H}^i_j = [\bm{h}_j \| \bm{G}^i].
\end{equation}
The final PDE solution over the whole domain $\{u(\bm{x}_i)\}_{i=1}^{s_i}$ is computed by
\begin{equation}
    \{u_i(\bm{x}_j)\}_{j=1}^{s_i} = \{\mathcal{U}_2 (\bm{H}^i_j, \theta_2)\}_{j=1}^{s_i}.
\end{equation}

\section{Methodology \label{sec.methodology}}

In this section, we introduce our proposed Physics-informed Geometry-aware Neural Operator (PI-GANO) for PDE operator learning. Our approach is primarily based on the PI-DCON architecture and draws inspiration from PI-PointNet.  {Our main goal in the model architecture design is to enable discretization-independent generalizability  for both variable parameters and variable geometry. To do so, we seek to integrate the parameter generalizability of PI-DCON with the geometry generalizability of the PI-PointNet architecture, in a way that facilitates physics-informed training.}

 {In the following sections, we  discuss how our architecture is inspired from PI-PointNet and present the mathematical representations of the geometry-coordinate embedding. Then, we discuss the architecture of DCON and its limitation in handling varying geometries in Section \ref{sec.DCON}, followed by a detailed discussion of our proposed model architecture in Section \ref{Sec.model_arch}.}

\subsection{Geometry-coordinate encoding \label{Subsec.geometry_encoder}}

 {Inspired by PI-PointNet, our approach aims to integrate the coordinate of collocation points  with  information on the domain geometry to compute the PDE solution. This enables the model to generalize across different domain geometries. We first present the mathematical representation for the geometry-coordinate embedding that encapsulate both local coordinate features and global geometric features.}

We first propose a multi-layer perception layers (MLP) $\mathcal{U}_G(\cdot, \theta_G)$ to capture the feature of the domain geometry.  Specifically, we represent a given domain geometry, denoted by $\Omega_i$, using a number of collocation points, with coordinates $\{\bm{x}^i_j\}_{j=1}^{s_i}$. We compute the high-dimensional features of each collocation point using the MLP. Then, similarly to PI-PointNet, we use a pooling layer to extract the global feature of the geometry in each embedding dimension. Hence, the global features of the $i$-th domain geometry $\bm G^i$ is computed by:
\begin{equation}
    \bm G^i = \text{AvgPool}(\{\bm{h}_j^i\}_{j=1}^{s_i}),
    \label{eq.geo_emb_cal}
\end{equation}
where $  \{\bm{h}_j^i\}_{j=1}^{s_i} = \{\mathcal{U}_G (\bm{x}_j^i, \theta_G)\}_{j=1}^{s_i}$, and  $\theta_G$ are trainable parameters of the MLP.  We consider three options for selecting the collocation points: (1) selecting points only on the boundary segments that vary, (2) selecting points on the entire boundary, and (3) selecting points inside the entire domain. As discussed in Section \ref{Subsec.geo_embed_study}, selecting  points only on the variable boundary segments  led to the best performance. Additionally, we select the Average-pooling layer instead of other pooling layers as it slightly enhanced the model performance in our experiments.  {The detailed numerical comparison of the effects of  different pooling layers are presented in Section \ref{Subsec.geo_embed_study}}.

Next, we compute the local coordinate feature of the collocation point on which the PDE solution is to be calculated. Specifically, for $\bm{x}_j^i$ in domain geometry $\Omega_i$, this is done via a linear layer parameterized by $W_t$ and $B_t$ and an activation function $\sigma$:
\begin{equation}
    \bm{h}_j^i = \sigma (W_t \bm{x}_j^i + B_t).
\end{equation}
Then we concatenate the geometry global feature $\bm{G}^i$ with each local coordinate feature $\bm{h}_j^i$ to obtain the geometry-coordinate embedding $\bm{H}^i_j$ of each collocation point $\bm{x}_j^i$ in the geometry $\Omega_i$ by:
\begin{equation}
    \bm{H}_j^i = [\bm{h}_j^i \| \bm{G}^i].
\end{equation}

\begin{figure}[ht]
    \centering
    \includegraphics[width=10cm]{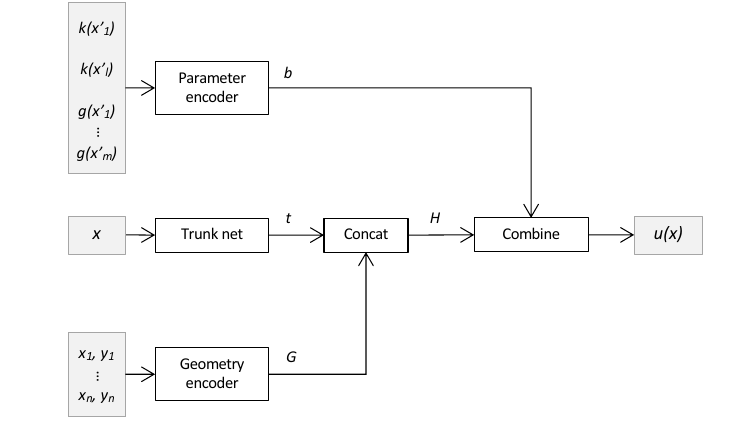}
    \caption{\footnotesize A schematic of how information on the domain geometry can be integrated with the location and parameter embeddings. The variable inputs include  $l$ and $m$ discrete values of parameter functions, together with the coordinates of $n$ points on the domain boundary. }
    \label{fig.high_level_idea_GANO}
\end{figure}

 {Using these geometry-coordinate embeddings $\bm H$ and the PDE parameter features, we can employ various novel model architectures to predict the PDE solutions. A schematic of our proposed model is shown in Figure \ref{fig.high_level_idea_GANO}. In this work we focus on the case where PDE parameters are not required to be represented using  the same number of discretization points, which is especially limiting for variable geometries.   Therefore, the numerical experiments are conducted using the PI-DCON architecture, as it is, to the best of the authors' knowledge, the only physics-informed model capable of generalizing to different function discretizations.}

\subsection{Deep Compositional Operator Network \label{sec.DCON}}

Our model is built upon on the  architecture of Deep Compositional Operator Network \cite{PIDCON}.  {Without loss of generality, let us consider a simplistic case where the variability is only in boundary conditions,  denoted by $g(\bm{x})$.} Our goal is then to find an approximation for the following operator,
\begin{equation}
    \mathcal{M}: \{g(\bm{x}), \Omega\} \rightarrow u(\bm{x}), \quad \bm{x} \in \Omega. 
\end{equation}

We can approximate this PDE operator with the architecture of Deep Compositional Operator Network (DCON). Using DCON, the PDE parameters are represented in a discrete form by evaluating function $g$  on a finite number of sampled coordinates $[\bm{x}'_1, \bm{x}'_2, ..., \bm{x}'_m]$, i.e. $\bm{g}=[g(\bm{x}'_1), g(\bm{x}'_2), \cdots, g(\bm{x}'_m)]$. A multi-layer perceptron, $\mathcal{U}_b(\cdot,\theta_b)$, is applied to it with a Max-pooling layer to derive the low-dimensional feature embedding $\bm{b}$ by:
\begin{equation}
    \bm{b} = \text{Maxpool}(\{\mathcal{U}_b(g(\bm{x}'_j),\theta_b)\}_{j=1}^{m}),
\end{equation}
where $\theta_b$ are  trainable parameters. After computing the  parameter embedding $\bm{b}$, the PDE operator $\mathcal{M}$ is approximated by a compositional version of DeepONet \cite{DON} with multiple stacked operator layers $\{\mathcal{O}_j\}$
\begin{equation}
    \mathcal{M} \approx \mathcal{M}_{\theta} =  \text{sum} \{\bm b \odot ... \ \mathcal{O}_2(\bm b \odot \mathcal{O}_1(\bm b \odot \bm t(\bm{x})))\},
    \label{eq.theorem_DCON}
\end{equation}
where $\theta$ denotes all the trainable parameters of the operator layers, described later, $\odot$ refers to component-wise multiplication, $\bm{t}: \mathbb{R}^d \rightarrow \mathbb{R}^q$ is a learnable nonlinear mapping from coordinates to hidden embedding, $\mathcal{O}: \mathbb{R}^q \rightarrow \mathbb{R}^q$ is a learnable nonlinear mapping of the operator layers, and $\text{sum}(\cdot)$ returns the summation of the components of a vector. In this work, we use a single linear layer with an activation function as the mapping $\mathcal{O}$, leading to the following operator approximation 
\begin{equation}
    \mathcal{M} \approx \mathcal{M}_{\theta} = \text{sum} \{\bm b \odot  ... (\bm b \odot (W^2_O \ \sigma(\bm b \odot (W^1_O \ \sigma (W_{\bm t}  \bm{x} + B_{\bm t}) + B^1_O)) + B^2_O)  )\},
    \label{eq.formula_model_arch}
\end{equation}
where  $\sigma(\cdot)$ represents the nonlinear activation function, $W_{\bm t} \in R^{q \times d}, B_{\bm t} \in R^{q}$ are  trainable parameters (weights and biases) of the mapping  $\bm t(\cdot)$, and $W^j_O \in R^{q \times q}, B^j_O \in R^{q}$ are  trainable parameters of the $j$-th operator layer, $\mathcal{O}_j$.

It should be noted that this operator is expected to have limitation in handling variation in domain geometry. That is, for a problem  with the same parameter function $g(\bm{x})$  but two different domains  $\Omega_1$ and $\Omega_2$, the predictions at a given coordinate $\bm{x}$, that falls within both domains, will be identical, i.e.
\begin{equation}
    \mathcal{M}_{\theta}(g(\bm{x}), \Omega_1) = \mathcal{M}_{\theta}(g(\bm{x}), \Omega_2).
    \label{eq.DCON_limitation}
\end{equation}
This shows that while DCON can still predict solution for any collocation points in a varied or unseen domain, it fails to produce unique predictions for  distinct domains. This suggests that the model may exhibit poor performance especially when changes in domain geometry are substantial.

\subsection{Proposed model architecture \label{Sec.model_arch}}

We seek to develop a model that makes a different prediction at the same spatial coordinate when the domain geometry is modified. To encode the domain geometry into the existing architecture of DCON, we introduce PI-GANO, which combines the representations for geometry-coordinate embedding and DCON.

\begin{figure}[ht]
    \centering
    \includegraphics[width=14cm]{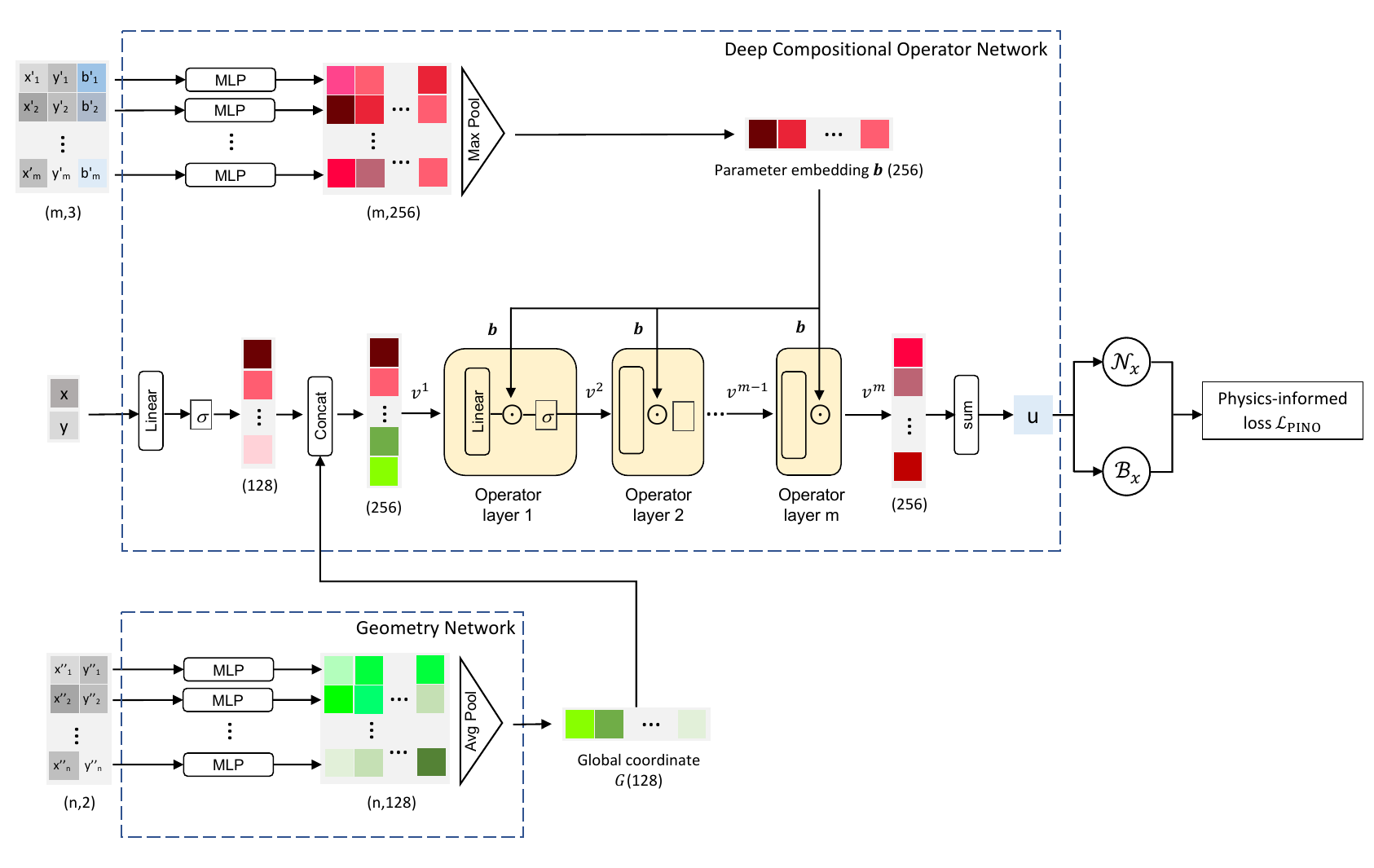}
    \caption{\footnotesize The architecture of the physics-informed Geometry aware Neural Operator is shown. The grey blocks represent the coordinates and the blue blocks represent the function values. The red blocks represent the hidden embeddings. The value of $b'_i$ is the boundary condition value evaluated on $(x'_i, y'_i)$. The value of $u$ is the solution value evaluated on $(x, y)$. It should be noted that no activation function is included in the last operator layer.}
    \label{fig.GADCON}
\end{figure}

In our proposed model architecture, the coordinate-geometry embedding $\bm{H}_j^i$ will be input to a series of operator layers to produce the PDE solution. Finally, the  PDE prediction of our model is given by:
\begin{equation}
    u_i(\bm{x}^i_j) \approx \mathcal{M}_{\theta}(\bm{x}^i_j, g_i(\bm{x}), \Omega_i) = \text{sum} \{\bm b \odot  ... (\bm b \odot (W^2_O \ \sigma(\bm b \odot (W^1_O [\sigma (W_t \bm{x}_j^i + B_t) \| \bm{G}^i] + B^1_O)) + B^2_O)  )\},
\end{equation}
where $\bm{G}^i$ is the geometry embedding of $i$-th geometry $\Omega_i$ computed by Equation \ref{eq.geo_emb_cal}. The details about the proposed model architecture are shown in Figure \ref{fig.GADCON}.

Although our model utilizes a similar architecture to PI-PointNet for generating geometry embeddings and coupling global and local features, there are fundamental differences between our architecture and that of PI-PointNet. In the physics-informed training algorithm, we estimate  the network parameters by minimizing the PDE residual. This residual in general involves spatial and temporal derivatives which means taking the derivative with respect to the input layer. Therefore, the  backward propagation path for derivative calculation using auto differentiation plays a significant role in the quality of physics-informed training. 

\begin{figure}[!ht]
 \centering
 \includegraphics[width=10cm]{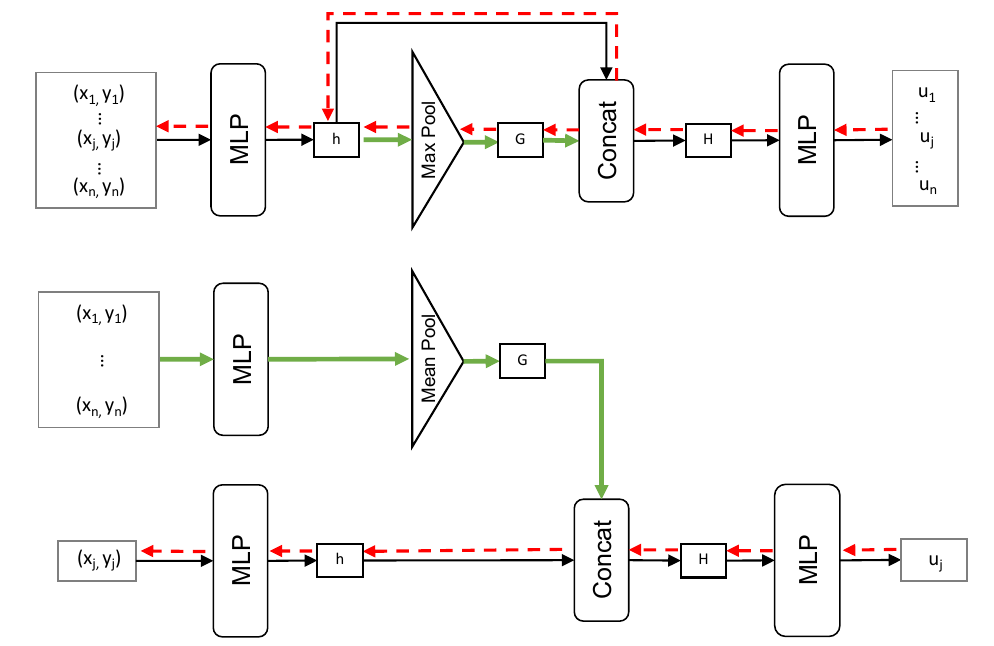}
\caption{\footnotesize The differences between our model architecture and PI-PointNet are shown. The black solid arrows are forward computation path for PDE solution prediction, while the red dashed arrows are backward computation path for derivative calculation of the approximated function.}
 \label{fig.inspiration}
\end{figure}

Figure \ref{fig.inspiration} shows the computation graphs of PI-PointNet (top) and our model (bottom) for PDE solution  in a two-dimensional domain. Solid arrows indicate the forward computation steps, and dashed arrows show the backward propagation steps for derivative calculation. If the two MLPs in our model are identical, it appears that our model and PI-PointNet would share the exact same computation path for PDE solution prediction. However, in our model, the computation of local (collocation point-based) features is independent from that of global features. This results in different computation paths for the derivatives of the model. For instance, in PI-PointNet, the first order derivative of $u_j$ with respect to coordinate $x_j$ is formulated by:
\begin{equation}
    \frac{\partial u_j}{\partial x_j} = \frac{\partial u_j}{\partial H_j} \cdot \left[\frac{\partial H_j}{\partial h_j} \| \frac{\partial H_j}{\partial G} \frac{\partial G}{\partial h_j}\right] \cdot \frac{\partial h_j}{\partial x_j},
    \label{eq.PI-PointNet_derivative}
\end{equation}
where $\|$ represents the operation of concatenation. However, the derivative approximation in our model architecture is simpler than Equation \ref{eq.PI-PointNet_derivative}, which is formulated by:
\begin{equation}
    \frac{\partial u_j}{\partial x_j} = \frac{\partial u_j}{\partial H_j} \cdot \frac{\partial H_j}{\partial h_j} \cdot \frac{\partial h_j}{\partial x_j}.
    \label{eq.PIGADCON_derivative}
\end{equation}

Comparing Equation \ref{eq.PI-PointNet_derivative} to Equation \ref{eq.PIGADCON_derivative}, it is evident that the differentiation process in our proposed architecture involves calculating only the derivatives of MLPs, whereas the PI-PointNet's  differentiation process also involves taking the  derivative of the Max-pooling layer. Based on the Universal Approximation Theorem for neural networks \cite{NN_universal_approx}, an MLP can approximate any continuous function and its derivatives within certain error bounds. This is while automatic differentiation over the Max-pooling layer, negatively impacts the model's performance. The numerical results presented in Section \ref{Subsec.main_results} will further show the differences in performance between these two back-propagation paths.

\section{Numerical results \label{sec.results}} 

In this section, we numerically evaluate the accuracy of the proposed model for solving parametric differential equations. We compare our models with Physics-informed DCON (PI-DCON), Physics-informed PointNet (PI-PointNet). In order to add PDE parameter generalizability to PI-PointNet, we will also introduce a modified version of PI-PointNet, named PI-PointNet* (as explained in Section \ref{Subsec.M_PI-PointNet}). This modified model, as an additional baseline model, offers a more fair comparison of effectiveness in generalization to variations in both PDE parameters and domain geometry.

In the numerical experiments, we use the following default settings unless mentioned otherwise. We use the hyperbolic tangent function (Tanh) \cite{tanh} as our activation function to ensure the smoothness in high-order derivatives of the models. For PI-DCON and our model architecture, we use three operator layers of width 512. For the architecture of PI-PointNet and PI-PointNet*, we use three hidden layers of width 512. Adam is the default optimizer with the following default hyper-parameters: $\beta_1$ = 0.9 and $\beta_2$ = 0.999 \cite{adam}. Each model is trained on 70\% of the sampled PDE parameters and validated on 10\% of the sampled PDE parameters. Well-trained models predict the PDE solution for the remaining 20\% of the sampled PDE parameters. There are two hyper-parameters used in model training: learning rate and coordinate sampling size ratio in each epoch for computing the PDE residual. We implemented grid search \cite{grid-search} to obtain the best set of hyper-parameters and report the corresponding model performance.  The possible values of learning rate are 0.001, 0.0005, 0.0002, and 0.0001, and the possible values of coordinate sampling size ratio are 0.3,0.2,0.1,0.05.  {For the remaining hyperparameters related to the architecture of DCON, we selected their values based on the hyperparameter studies presented in \cite{PIDCON}.} Model training is performed on an NVIDIA P100 GPU using a batch size of 20. We collect the parameters of PDEs by Monte Carlo Simulation \cite{mcs} of stochastic processes \cite{sp} and derive the solution of PDEs based on  the Finite Element Method using Matlab \cite{matlab}.  {For each of the two problems investigated in our work, we generated 500 PDE samples for the experiments. In contrast to the work in \cite{PIDCON}, which focuses on mesh generalization, we used a consistent resolution across all FEM results in our study, as our primary focus in this study is on generalization across different domain geometries.}

\subsection{A modified PI-PointNet as a baseline \label{Subsec.M_PI-PointNet}}

\begin{figure}[ht]
    \centering
    \includegraphics[width=15cm]{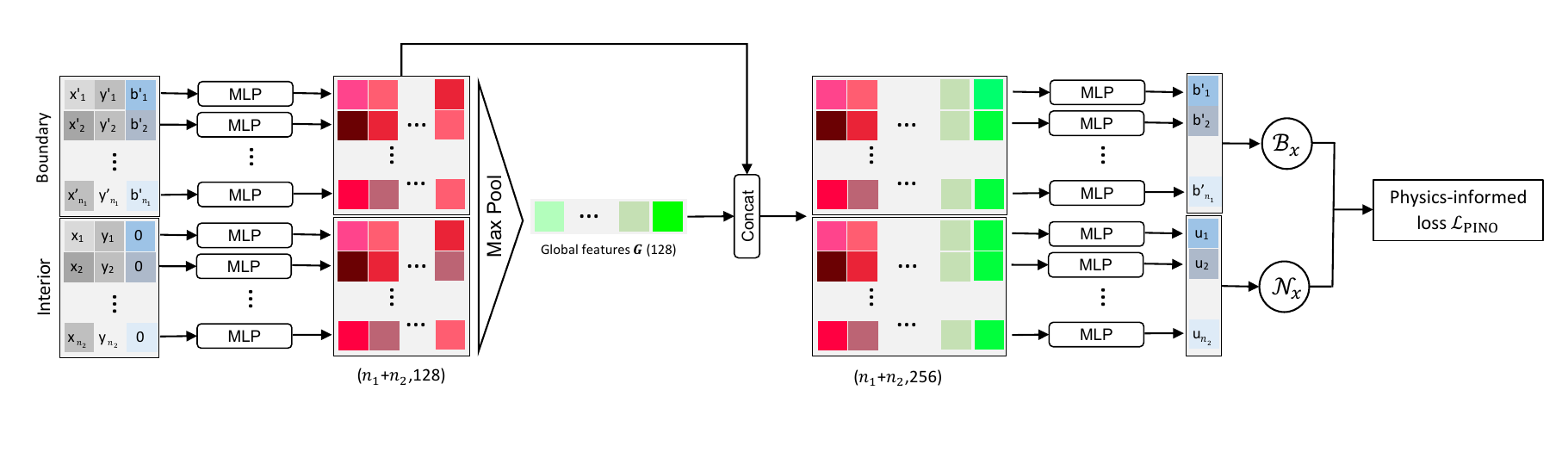}
    \caption{\footnotesize The architecture of the modified physics-informed PointNet (PI-PointNet*).}
    \label{fig.MPI-PointNet}
\end{figure}

In its original form, PI-PointNet is not designed for generalization across different PDE parameters. Therefore, to create a baseline with this generalization capability, we have modified it to include a part which encodes PDE parameter information into the model inputs. In this modified architecture, called PI-PointNet*, if the function values of PDE parameters are available, they are concatenated with the coordinates of the collocation points as the model input. If not available, zeros are concatenated instead. This setup ensures that both the PDE parameters and the domain geometry are provided to the model, allowing the PDE solution to depend on both factors.  Figure \ref{fig.MPI-PointNet} shows  the architecture of PI-PointNet* for the scenario where the only variable PDE parameter is the boundary conditions,  {where $n_1$ is the number of collocation points on the boundary and $n_2$ is the number of collocation points inside the domain.}

\subsection{Experiment Setups}

\subsubsection{A Darcy flow problem}

\begin{figure}[ht]
    \centering
    \includegraphics[width=5cm]{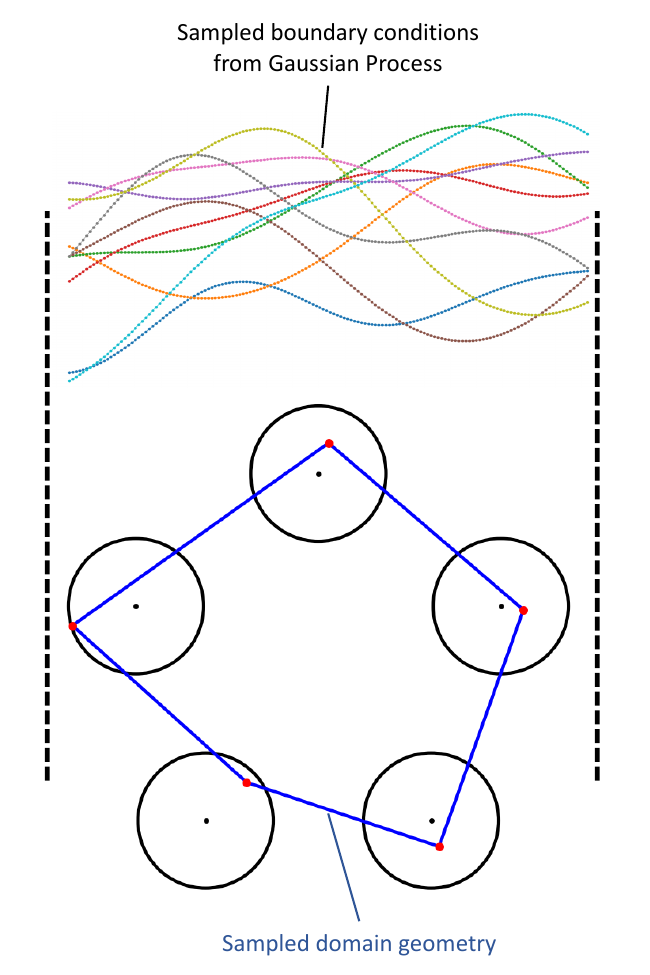}
    \caption{\footnotesize The boundary conditions in our Darcy flow experiment setting shown on a random geometry. The boundary conditions $g(\bm{x})$  are samples from a Gaussian process.}
    \label{fig.darcy_BC}
\end{figure}

 {As the first example, we consider a Darcy flow problem over variable domain geometries and boundary conditions. A motivation for creating a fast geometry-aware solver is to facilitate the engineering design problem that seeks the  optimal  shape of a filtration system that enhances the fluid flow through the porous media.} Let us specifically consider a two-dimensional Darcy flow problem in a five-edge polygon, where the steady state solution is described by,
\begin{equation}
    \begin{aligned}
    - \nabla (k(x,y) \nabla p(x,y)) &= f(x,y), \quad (x,y) \in \Omega,\\
    p(x,y) &= g(x,y), \quad (x,y) \in \partial \Omega.
\end{aligned}
\end{equation}
where $p(x,y)$ is the pressure, $k(x,y)$ is the permeability field,  $f(x,y)$ is the source term, and $g(x,y)$ is the prescribed pressure on the domain boundaries $\partial \Omega$. Following the same setting as \cite{NO_compare}, we consider $k(\bm{x})=1$ and $f(\bm{x})=10$,  {where $\bm{x}=(x,y)$ denotes the coordinates}. To create variable domain geometries, we consider an analytical expressions, where the coordinates of the five vertices of the polygon is varied. To do so, we first draw five random coordinates to be the center points for the five circles of radius of $0.25$ as shown in ~\ref{fig.darcy_BC}. These five center points are uniformly drawn on the perimeter of the circle, with the radius of $1$ centered at $(0,0)$. Inside each of the five  circles, we will randomly generate one point to serve as a vertex of the polygon-shaped domain.

The boundary $\partial \Omega$ in this example is chosen to  be the edges of the polygon, on which we impose the nonzero pressure function $g(x,y)$. The imposed pressure is assumed to follow a zero mean Gaussian process with a covariance kernel that is only dependent on the horizontal component of the distance, i.e., 
\begin{equation}
    \begin{aligned}
    g(x,y) &\sim \mathcal{GP}(0, K(x,x')),\\
    K(x,x') &= \exp\left[-\frac{(x-x')^2}{2l^2}\right], \quad l=1.
\end{aligned}
\label{eq.gp}
\end{equation}

We employ the Gaussian process model of Eq.~\ref{eq.gp} to generate $N=500$ different imposed pressure functions. Then, given each of these $N$ sampled functions, $\{g_i\}_{i=1}^N$, we solve the PDE using the Finite Element Method (FEM) \cite{matlab}. Specifically, for the $i$-th realized domain $\Omega_i$ and BC function $g_i$, let $M_{i}$ be the number of discretization locations on $\partial \Omega_i$, and $\bm{X}_{\text{BC},i}$ be the coordinates of these locations, i.e.
\begin{equation}
    \bm{X}_{\text{BC},i} = \{(x_j^{i}, y_j^{i})\}_{j=1}^{M^{i}},
\end{equation}
where $M^{i}$ is the size of the discretization generated for the $i$-th sampled geometry. We then collect all the boundary conditions information for the $i$-th geometry in  $\bm{G}^i$, i.e.
\begin{equation}
    \bm{G}_i = \{\bm{X}_{\text{BC},i}, g_i(\bm{X}_{\text{BC},i})\},
\end{equation}
where $g_i(\bm{X}_{\text{BC},i})$ is the $i$-th sampled BC function evaluated at the coordinates $\bm{X}_{\text{BC},i}$.

Given this setup, we seek to learn the nonlinear mapping $\mathcal{M}_{\text{Darcy}}$ that transforms a given boundary function $g(x,y)$ on $\partial \Omega$ to the pressure field $p(x,y)$ in the entire domain, i.e.,
\begin{equation}
    \mathcal{M}_{\text{Darcy}}: g(x,y) \rightarrow p(x,y).
\end{equation}
Specifically, for a 2D problem, the neural operator $p_{\theta}(x,y, \bm G)$  
is trained using the loss function
\begin{equation}
    \mathcal{L}(\theta) = \mathcal{L}_{\text{PDE}}(\theta) + \alpha \mathcal{L}_{\text{BC}_1}(\theta),
    \label{eq.darcy_loss}
\end{equation}
where $\alpha$ is the weighing hyperparameters, set to be $\alpha=500$ to ensure similar orders of magnitude; $\mathcal{L}_{\text{PDE}}(\theta)$ and $\mathcal{L}_{\text{BC}}(\theta)$ are the PDE residual and the BC loss function, given by
\begin{equation}
\begin{aligned}
    \mathcal{L}_{\text{PDE}}  (\theta) &= \frac{1}{N} \sum_{i=1}^N \left[\frac{\partial^2 p_{\theta}(x,y, \bm G_i)}{\partial x^2} + \frac{\partial^2 p_{\theta}(x,y, \bm G_i)}{\partial y^2} + 10 \right] ^2,  &&(x, y) \in \Omega \\
    \mathcal{L}_{\text{BC}}(\theta) &= \frac{1}{N} \sum_{i=1}^N \left\{\frac{1}{M_{i}} \sum_{j=1}^{M_{i}} \left[ p_{\theta}(x^i_j,y^i_j, \bm G_i) - 0 \right]^2 \right\},  &&(x^i_j, y^i_j) \in \bm{X}_{\text{BC},i}.
\end{aligned}
\end{equation}

\subsubsection{A 2D plate problem \label{Subsec.2D_plate}}

 {In this example, we consider the  placement of bolt holes in a 2D plate under varying loading conditions. This is} a more complex problem which involves a system of differential equations and  a nonlinear mapping is to be learned between multiple parameter functions and multiple response functions. In particular, let us consider a solid mechanics problem for a two-dimensional  plate with four holes. The static solution for the plate displacements is governed by the following system of partial differential equations
\begin{equation}
\begin{aligned}
    \frac{E}{1-\mu^2}\left[\frac{\partial^2 u(x,y)}{\partial x^2} + \frac{(1-\mu)}{2} \frac{\partial^2 u(x,y)}{\partial y^2} + \frac{(1+\mu)}{2} \frac{\partial^2 v(x,y)}{\partial x \partial y}\right] &= 0, \quad (x,y) \in \Omega,  \\
    \frac{E}{1-\mu^2}\left[\frac{\partial^2 v(x,y)}{\partial y^2} + \frac{(1-\mu)}{2} \frac{\partial^2 v(x,y)}{\partial x^2} + \frac{(1+\mu)}{2} \frac{\partial^2 u(x,y)}{\partial x \partial y}\right] &= 0, \quad (x,y) \in \Omega,
\end{aligned}
\end{equation}
where $u$ and $v$ are the plate displacements in $x$ and $y$ directions, respectively, $E$ is the Young's Modulus, and $\mu$ is the Poisson's Ratio. 

\begin{figure}[ht]
    \centering
    \includegraphics[width=16cm]{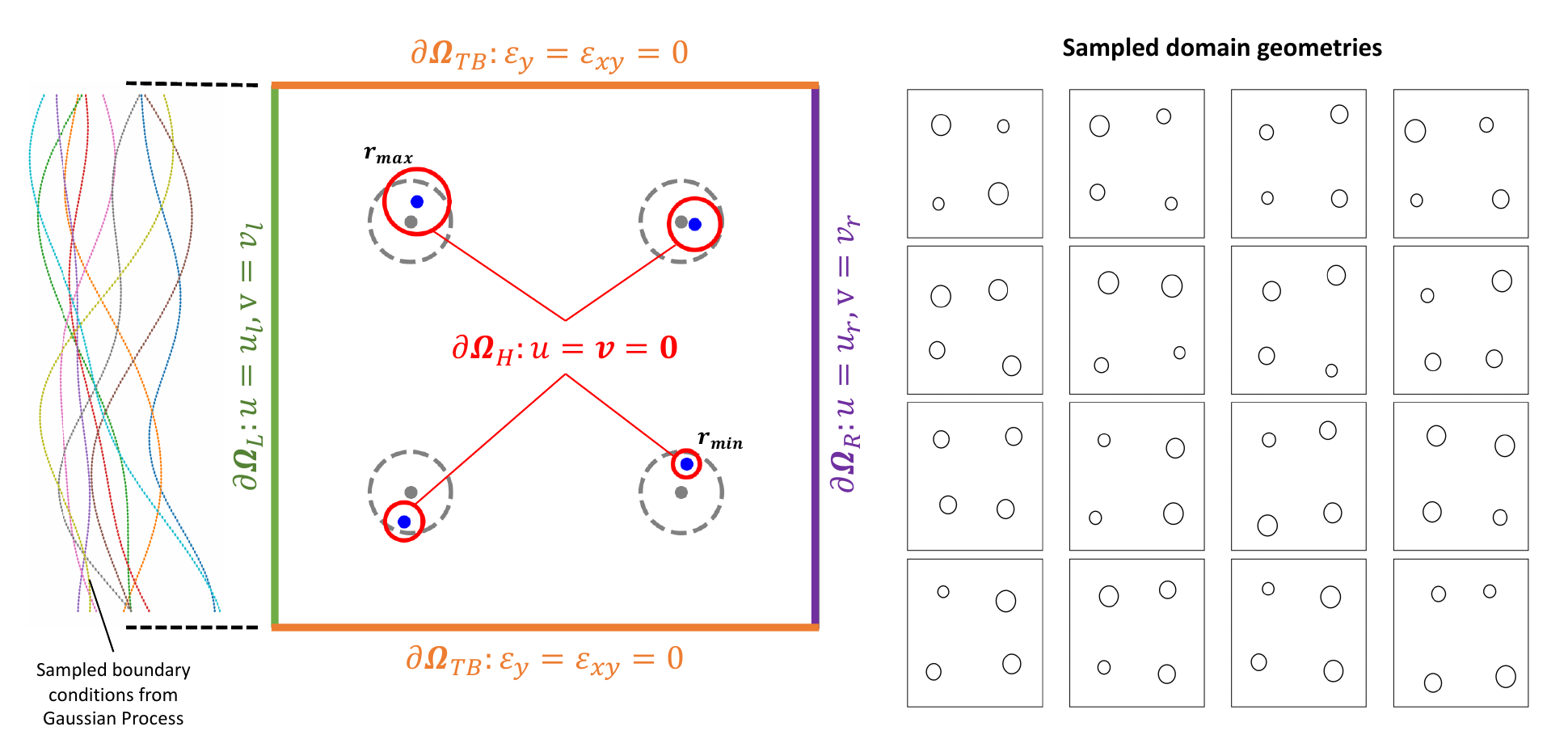}
    \caption{\footnotesize The boundary conditions of 2D Plate experiment, and 2D plates with largest and smallest bolt hole area are shown. The samples of Gaussian processes as the prescribed displacement on leftmost edges and rightmost edges are also shown. The dots on the curve are considered as the representation of the prescribed displacements.}
    \label{fig.plate_BC}
\end{figure}

In this experiment, the plate is considered to be 20 mm $\times$ 20 mm, and the location and  size of the bolts are variable. To generate variable geometries, we first consider four circles of radius of 1.5 centered at $(5,5)$, $(5,-5)$, $(-5,5)$, and $(-5,-5)$ respectively  (shown by dashed lines in Figure~\ref{fig.plate_BC}). In each circle, we randomly select one point to be the center of a bolt hole. The radius of the holes are uniformly drawn from the range $[0.8,1.5]$.  The edge of the holes are subjected to the ``fixed" boundary condition (zero displacements). The left and right sides of the plate are subject to a prescribed imposed displacements (given as variable functions), while the top and bottom sides are assigned the ``free" boundary conditions (zero strains). 

The complete boundary conditions are shown in Figure ~\ref{fig.plate_BC}, and are given by
\begin{equation}
\begin{aligned}
    u(x,y) = 0, \quad v(x,y) = 0, \quad (x,y) \in \partial \Omega_{\text{H}}, \\
    u(x,y) = u_L(x,y), \quad v(x,y) = v_L(x,y) \quad (x,y) \in \partial \Omega_{\text{L}}, \\
    u(x,y) = u_R(x,y), \quad v(x,y) = v_R(x,y) \quad (x,y) \in \partial \Omega_{\text{R}}, \\
    \frac{\partial v(x,y)}{\partial y} = 0, \quad \frac{1}{2} \left[ \frac{\partial u(x,y)}{\partial y} + \frac{\partial v(x,y)}{\partial x} \right] = 0, \quad (x,y) \in \partial \Omega_{\text{TB}},
\end{aligned}
\end{equation}
where the  prescribed displacement functions $u_L, v_L, u_R, v_R$ are the input functions to the operator. Similar to the previous example, these functions are considered to follow a Gaussian Process model. The covariance kernel  is assumed to be dependent only on the vertical component of the distance, i.e.,
\begin{equation}
\begin{aligned}
    u_l, v_l, u_R, v_R &\sim \mathcal{GP}(1, K(y,y')),\\
    K(y,y') &= \exp\left[-\frac{(y-y')^2}{2l^2}\right], \quad l=5.
\end{aligned}
\end{equation}

Similarly to the previous example, we draw $N=500$ realizations of these functions, and obtain FEM solutions using meshes with different sizes.   Figure \ref{fig.plate_BC} shows  the scheme of the variable geometry generation and the boundary conditions settings. Some randomly generated geometries are also shown on the right. The boundary condition information on the left and right sides  for the $i$-th realization is collected into the following four vectors
\begin{equation}
\begin{aligned}
    \bm{X}_{\text{BC}_{\text{TB}},i} = \{(x_j^{1,i}, y_j^{1,i})\}_{j=1}^{M_{1,i}}, \\
    \bm{X}_{\text{BC}_L,i} = \{(x_j^{2,i}, y_j^{2,i})\}_{j=1}^{M_{2,i}}, \\
    \bm{X}_{\text{BC}_R,i} = \{(x_j^{3,i}, y_j^{3,i})\}_{j=1}^{M_{3,i}}, \\
    \bm{X}_{\text{BC}_H,i} = \{(x_j^{4,i}, y_j^{4,i})\}_{j=1}^{M_{4,i}},
\end{aligned}
\end{equation}
where $\bm{X}_{\text{BC}_1,i}$, $\bm{X}_{\text{BC}_2,i}$, $\bm{X}_{\text{BC}_3,i}$, $\bm{X}_{\text{BC}_4,i}$ are the sets of boundary coordinates sampled on $\partial \Omega_1$, $\partial \Omega_2$, $\partial \Omega_3$, $\partial \Omega_4$, respectively. The  discretized version of the input functions to the operator (sampled boundary values on the left and right sides) are then represented by the following matrices
\begin{equation}
\begin{aligned}
    \bm{G}_{L,i} &= \{\bm{X}_{\text{BC}_L,i}, u_{L,i}(\bm{X}_{\text{BC}_L,i}), v_{L,i}(\bm{X}_{\text{BC}_L,i})\} \\
    \bm{G}_{R,i} &= \{\bm{X}_{\text{BC}_R,i}, u_{R,i}(\bm{X}_{\text{BC}_R,i}), v_{R,i}(\bm{X}_{\text{BC}_R,i})\} 
\end{aligned}
\end{equation}

Given this setup, we seek to learn the nonlinear mapping $\mathcal{M}_{\text{Plate}}$ that transforms the given boundary displacement functions $u_L(x,y), v_L(x,y), u_R(x,y), v_R(x,y)$ on $\partial \Omega_2$ and  $\partial \Omega_3$ to the displacement field $u(x,y)$ and $v(x,y)$ in the entire domain, i.e.,
\begin{equation}
    \mathcal{M}_{\text{Plate}}: [u_L(x,y), v_L(x,y), u_R(x,y), v_R(x,y)]  \rightarrow [u(x,y), v(x,y)].
\end{equation}

Based on these boundary condition settings, the training loss function $\mathcal{L}(\theta)$ for training of  neural operators $u_{\theta}(x,y,\bm{G}_L, \bm{G}_R)$ and $v_{\theta}(x,y,\bm{G}_L, \bm{G}_R)$ is formulated as:
\begin{equation}
    \mathcal{L}(\theta) = \alpha_0 \mathcal{L}_{\text{PDE}}(\theta) + \alpha_1 \mathcal{L}_{\text{BC}_{\text{TB}}}(\theta) + \alpha_2 \mathcal{L}_{\text{BC}_L}(\theta) + \alpha_3 \mathcal{L}_{\text{BC}_R}(\theta) + \alpha_4 \mathcal{L}_{\text{BC}_H}(\theta),
\end{equation}
where $\alpha_0$, $\alpha_1$, $\alpha_2$, $\alpha_3$, $\alpha_4$ are the trade-off coefficients, set to be $\alpha_0=0.00001, \alpha_1=\alpha_2=\alpha_3=\alpha_4=1$ to ensure similar orders of magnitude; and $\mathcal{L}_{\text{PDE}}(\theta)$ is the PDE residual given by 
\begin{equation}
    \begin{aligned}
    \mathcal{L}_{\text{PDE}}(\theta) = \frac{1}{N} \sum_{i=1}^N \ & \left\{ \frac{E}{1-\mu^2} \left(\frac{\partial^2 u_{\theta}(x,y,\bm{G}_{L,i}, \bm{G}_{R,i})}{\partial x^2} + \frac{\partial^2 v_{\theta}(x,y,\bm{G}_{L,i}, \bm{G}_{R,i})}{\partial y^2} \right) + \right.\\
    & \left. \frac{E}{2(1+\mu)} \left(\frac{\partial^2 u_{\theta}(x,y,\bm{G}_{L,i}, \bm{G}_{R,i})}{\partial y^2} + \frac{\partial^2 v_{\theta}(x,y,\bm{G}_{L,i}, \bm{G}_{R,i})}{\partial x^2} \right) + \right.\\
    & \left.  \frac{E}{2(1-\mu)} \left(\frac{\partial^2 v_{\theta}(x,y,\bm{G}_{L,i}, \bm{G}_{R,i})}{\partial x \partial y} + \frac{\partial^2 u_{\theta}(x,y,\bm{G}_{L,i}, \bm{G}_{R,i})}{\partial x \partial y} \right) \right\}, \quad (x,y) \in \Omega. \\
\end{aligned}
\end{equation}
The BC loss terms $\mathcal{L}_{\text{BC}_{\text{TB}}}(\theta)$, $\mathcal{L}_{\text{BC}_L}(\theta)$, $\mathcal{L}_{\text{BC}_R}(\theta)$, $\mathcal{L}_{\text{BC}_H}(\theta)$ are given by
\begin{equation}
    \begin{aligned}
    \mathcal{L}_{\text{BC}_{\text{TB}}}(\theta) =  \frac{1}{N} \sum_{i=1}^N \ & \left\{ \frac{1}{M_{1,i}} \sum_{j=1}^{M_{1,i}}   \left[u_{\theta}(x_j^{i},y_j^{i},\bm{G}_{L,i}, \bm{G}_{R,i}) - 0 \right]^2 + \left[v_{\theta}(x_j^{i},y_j^{i},\bm{G}_{L,i}, \bm{G}_{R,i})- 0 \right]^2 \right\}, \quad (x_j^{i},y_j^{i}) \in \bm{X}_{\text{BC}_{\text{TB}},i},\\
    \mathcal{L}_{\text{BC}_L}(\theta) = \frac{1}{N} \sum_{i=1}^N \ & \left\{ \frac{1}{M_{2,i}} \sum_{j=1}^{M_{2,i}}   \left[u_{\theta}(x_j^{i},y_j^{i},\bm{G}_{L,i}, \bm{G}_{R,i}) - u_{L,i}(x_j^{i},y_j^{i}) \right]^2 + \left[v_{\theta}(x_j^{i},y_j^{i},\bm{G}_{L,i}, \bm{G}_{R,i})- v_{L,i}(x_j^{i},y_j^{i}) \right]^2 \right\}, \\
    & (x_j^{i},y_j^{i}) \in \bm{X}_{\text{BC}_L,i},\\
    \mathcal{L}_{\text{BC}_R}(\theta) = \frac{1}{N} \sum_{i=1}^N \ & \left\{ \frac{1}{M_{3,i}} \sum_{j=1}^{M_{3,i}}   \left[u_{\theta}(x_j^{i},y_j^{i},\bm{G}_{L,i}, \bm{G}_{R,i}) - u_{R,i}(x_j^{i},y_j^{i}) \right]^2 + \left[v_{\theta}(x_j^{i},y_j^{i},\bm{G}_{L,i}, \bm{G}_{R,i})- v_{R,i}(x_j^{i},y_j^{i}) \right]^2 \right\}, \\
    & (x_j^{i},y_j^{i}) \in \bm{X}_{\text{BC}_R,i},\\
    \mathcal{L}_{\text{BC}_H}(\theta) =  \frac{1}{N} \sum_{i=1}^N \ & \left\{ \frac{1}{M_{4,i}} \sum_{j=1}^{M_{4,i}}   \left[ \frac{\partial v_{\theta}(x_j^{i},y_j^{i},\bm{G}_{L,i}, \bm{G}_{R,i})}{\partial y_j^{i}} - 0 \right]^2 + \right. \\ 
    & \left. \left[ \frac{1}{2} \left( \frac{\partial u_{\theta}(x_j^{i},y_j^{i},\bm{G}_{L,i}, \bm{G}_{R,i})}{\partial y_j^{i}} + \frac{\partial v_{\theta}(x_j^{i},y_j^{i},\bm{G}_{L,i}, \bm{G}_{R,i})}{\partial x_j^{i}} \right) - 0\right]^2 \right\}, \quad (x_j^{i},y_j^{i}) \in \bm{X}_{\text{BC}_H,i}.
\end{aligned}
\end{equation}

\subsection{Main results \label{Subsec.main_results}}

For both the Darcy flow problem and the 2D Plate problem, we will generate two datasets: one with varying geometry only, and the other one with both varying PDE parameters and varying geometry. The first dataset is used to only investigate geometry generalization and  compare the performance of our proposed model with that of PI-PointNet. The second dataset is utilized to evaluate the parameter and geometry generalization abilities of our proposed model with PI-DCON and PI-PointNet*. It should be noted that for the 2D plate dataset, to properly discretize the boundary around the small holes, we employ a higher resolution compared to the Darcy problem. Typically, the meshes in the Darcy flow dataset contain  approximately 1,000 nodes, whereas the meshes in 2D plate dataset has about 20,000 nodes. We use relative $L_2$ error \cite{L2} as our error measure computed by:
\begin{equation}
    L_2 = \frac{\|\bm u_{\text{pred}}- \bm u_{\text{fem}}\|_2}{\|u_{\text{fem}}\|_2},
\end{equation}
where $u_{\text{pred}}$ refers to a vector of  neural network predictions at all the collocation points, and $u_{\text{fem}}$ is the corresponding  FEM solution vector.  We present the average error calculated across the entire test dataset, along with the standard deviation of the relative errors to demonstrate the model stability.

\begin{figure}[ht]
    \centering
            \begin{subfigure}[b]{0.3\textwidth}
            \centering
            \includegraphics[width=\textwidth]{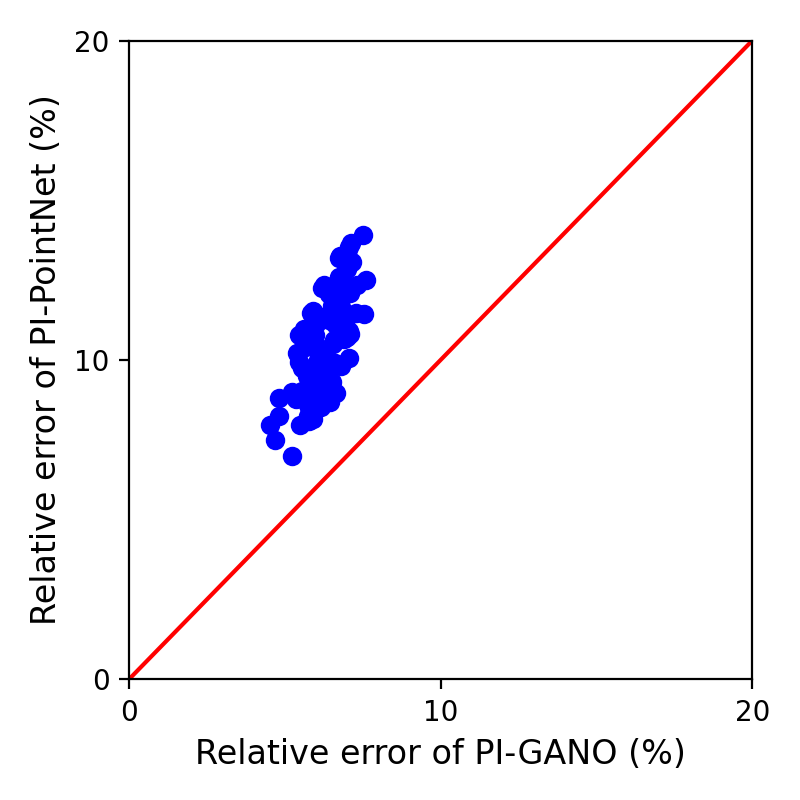}
            \caption{Darcy flow}    
            \label{fig:GA_vs_PI-PointNet_darcy}
        \end{subfigure}
        \hspace{0.2cm}
        \begin{subfigure}[b]{0.3\textwidth}   
            \centering 
            \includegraphics[width=\textwidth]{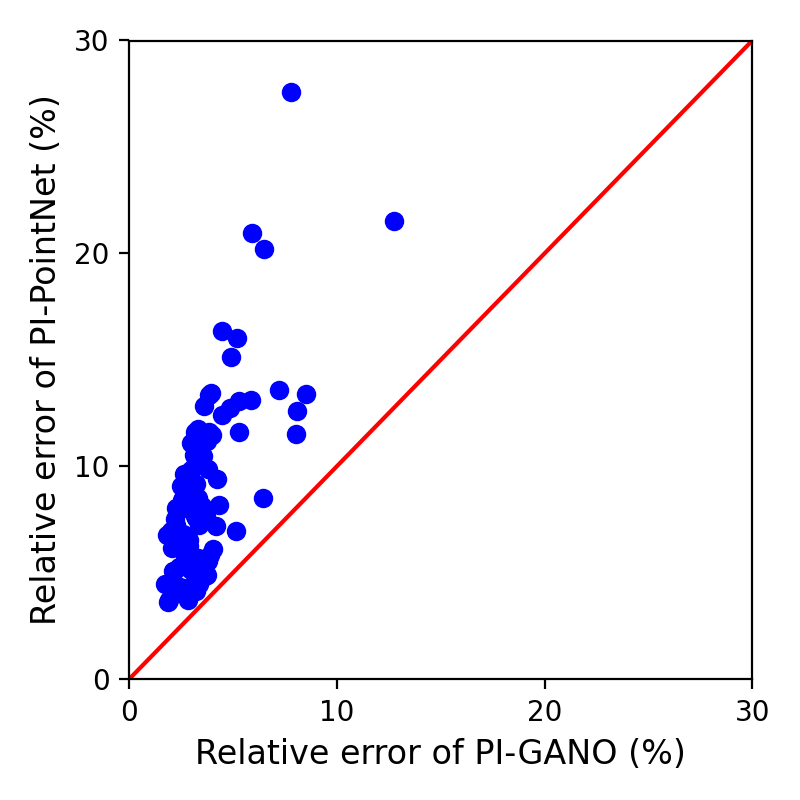}
            \caption{2D plate}  
            \label{fig:GA_vs_PI-PointNet_plate}
        \end{subfigure}
    \caption{\footnotesize Comparison between PI-GANO and PI-PointNet on the Darcy flow and 2D plate problems. It can be seen that in all the cases PI-GANO offers the smaller errors. \label{fig.err_only_geo}}
\end{figure}

Table \ref{table.PI_com_only_geo} shows the results for the case where only geometry generalization is compared with PI-PointNet.  As can be seen, our model achieved 39.1\% accuracy improvement for the Darcy flow problem and 75.0\% accuracy improvement for the 2D Plate problem. Figure \ref{fig.err_only_geo}, in more details, compares the two models, and  show the our method in all the geometry samples offers smaller error compared to PI-PointNet.

\begin{table*}[!ht]
\begin{center}
\caption{Accuracy and efficiency comparison between PI-GANO and PI-PointNet for the dataset of only geometry variation.}
\begin{tabular}{c c c c c c c}
\hline
\multirow{2}{3em}{Dataset} & \multicolumn{2}{c}{\makecell{Mean of relative error}} & \multicolumn{2}{c}{\makecell{Standard deviation of relative error}} & \multicolumn{2}{c}{\makecell{Training time (minutes)}}\\
\cline{2-7}
 & PI-PointNet & PI-GANO & PI-PointNet & PI-GANO & PI-PointNet & PI-GANO\\
\hline
\makecell{Darcy flow} & 11.30\% & \bf{6.70}\% & 3.52\% & \bf{2.09\%} & 32 & \bf{18}\\
\makecell{2D Plate} & 25.72\% & \bf{6.41\%} & 10.13\% & \bf{2.19\%} & 265 & \bf{57}\\
\hline
\label{table.PI_com_only_geo}
\end{tabular}
\end{center}
\end{table*}

Comparing the training times of the two models, we demonstrate that our method surpasses PI-PointNet in terms of efficiency, especially when handling fine meshes, noting the results for the 2D plate problem. In physics-informed training, the computation of PDE residuals is required in each iteration. In our proposed model, we can sample a small subset of the collocation points, in the trunk net, for the approximation calculation of loss and an efficient update of the model parameters. This is because the trunk net, which approximates the solution at a given point, is chosen to be separate from the geometry encoder. In contrast, PI-PointNet can only handle the entire point cloud representing the entire domain. Therefore, stochastic gradient descent algorithm with a small minibatch size cannot work for this model architecture.  As a result, we observe a significantly reduced training time for our model when dealing with fine meshes. To  visually compare  our method with PI-PointNet, the best and worst-case predictions are presented  in Table \ref{fig.best_worst_cases_only_geo}.

\begin{table*}[!ht]
\centering 
\caption{Comparison between the performance of PI-GANO and PI-PointNet. Out of all realizations of boundary conditions, the ones that causes best and worst performances of each model are shown.}
\begin{tabular}{|c|c|c | c c | c c|}
\hline
\multicolumn{2}{|c|}{ } & \multirow{2}{*}{Ground Truth} & \multicolumn{2}{c|}{PI-PointNet} & \multicolumn{2}{c|}{PI-GANO} \\ 
\cline{4-7}
\multicolumn{2}{|c|}{ } &  & Prediction & Absolute Error & Prediction & Absolute Error \\ 
\hline
\multirow{8}{*}{\rotatebox[origin=c]{90}{Darcy flow }} & \rotatebox[origin=l]{90}{\makecell{PI-PointNet \\ Best case}} & {\includegraphics[width=0.16\textwidth]{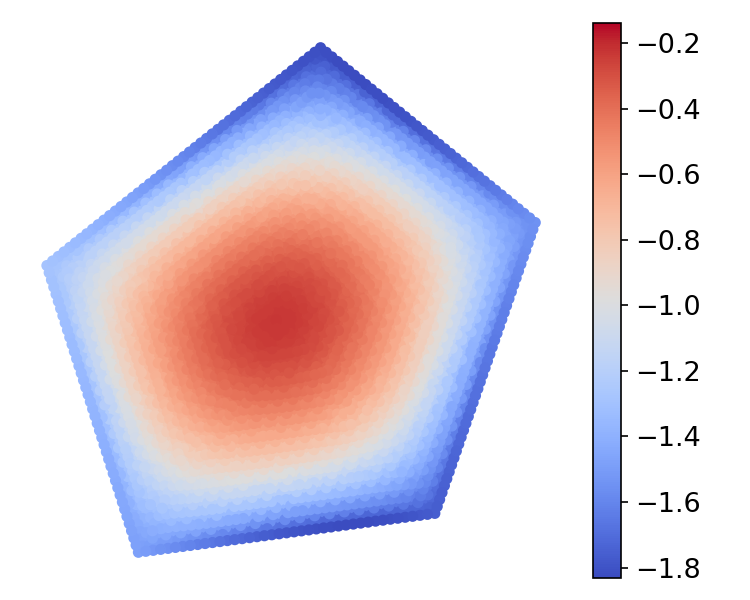}} & {\includegraphics[width=0.16\textwidth]{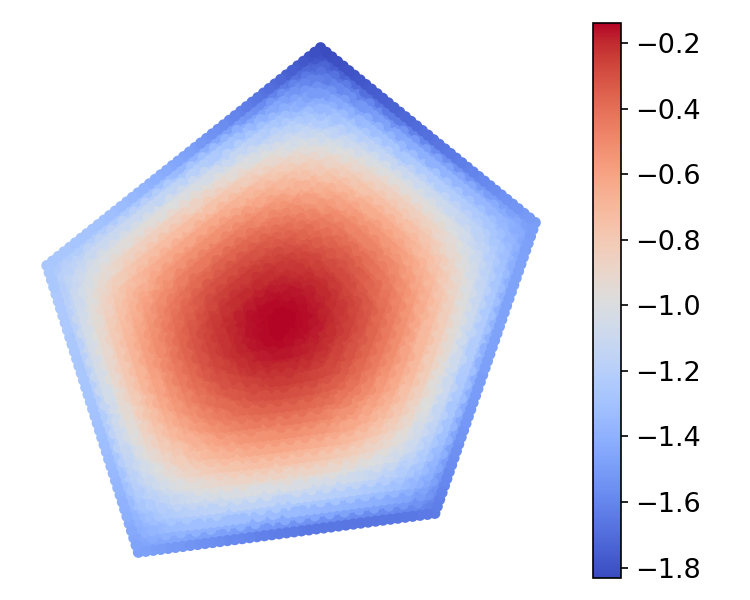}} & {\includegraphics[width=0.16\textwidth]{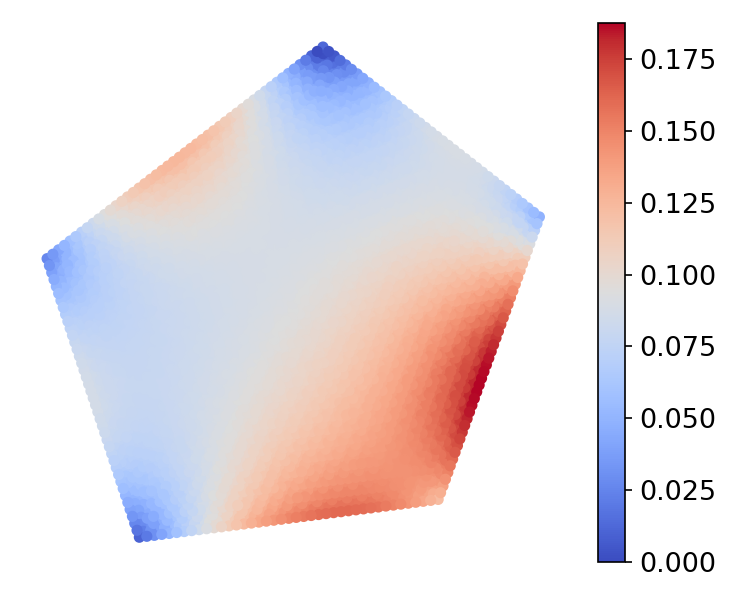}} & {\includegraphics[width=0.16\textwidth]{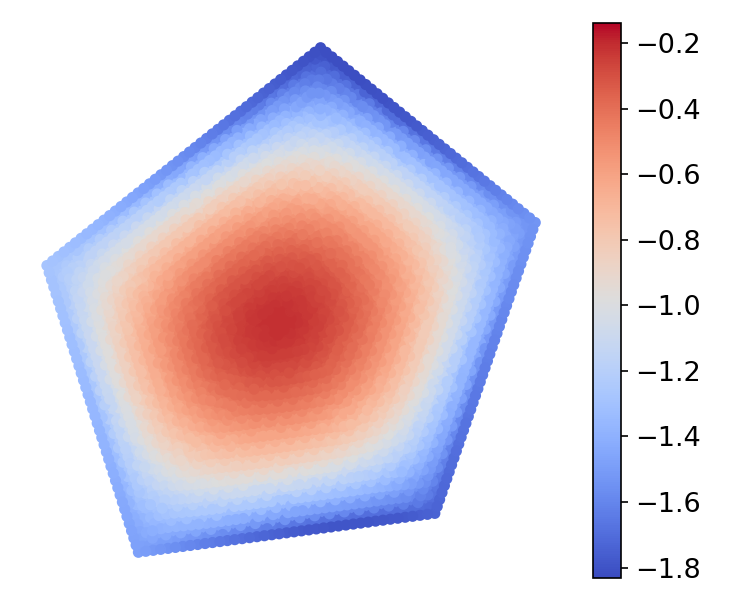}} & {\includegraphics[width=0.16\textwidth]{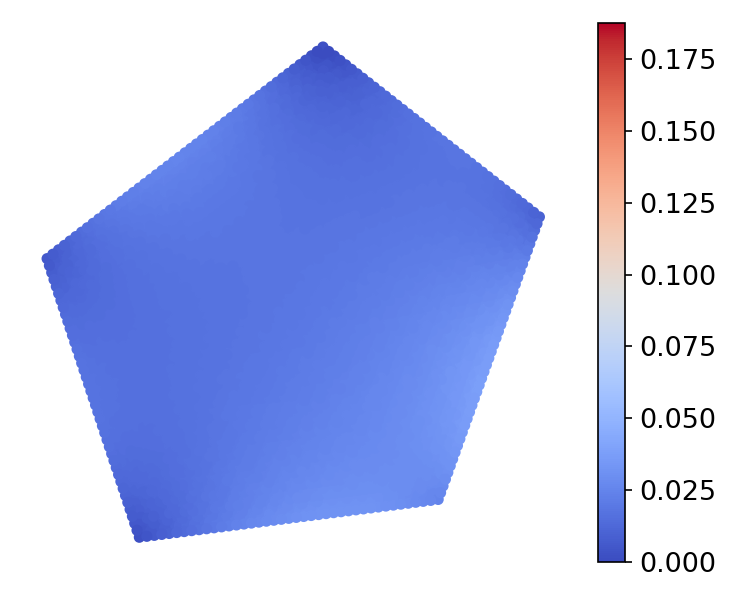}} \\
& \rotatebox[origin=l]{90}{\makecell{PI-GANO \\ Best case}} & {\includegraphics[width=0.16\textwidth]{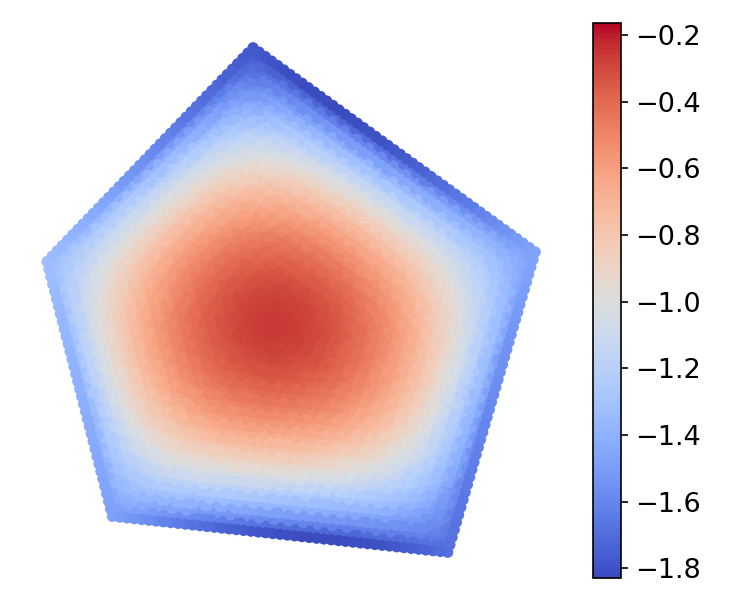}} & 
 {\includegraphics[width=0.16\textwidth]{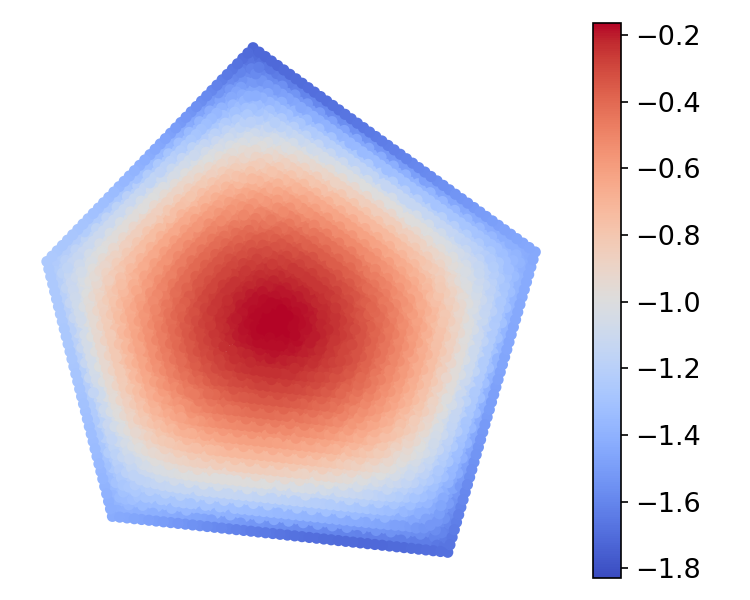}} & {\includegraphics[width=0.16\textwidth]{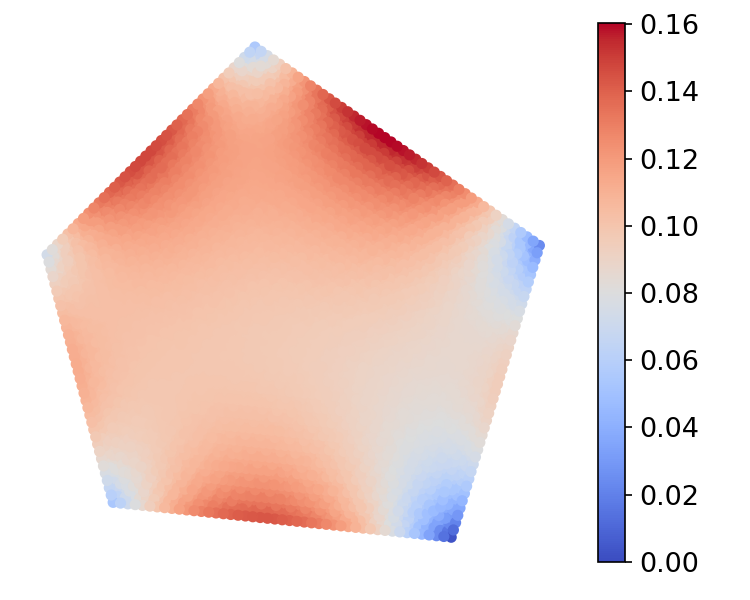}} &{\includegraphics[width=0.16\textwidth] {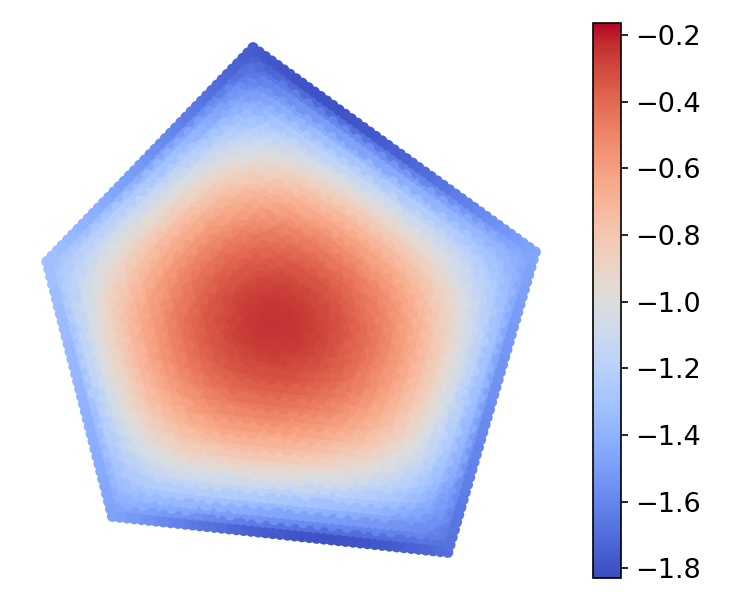}} & {\includegraphics[width=0.16\textwidth]{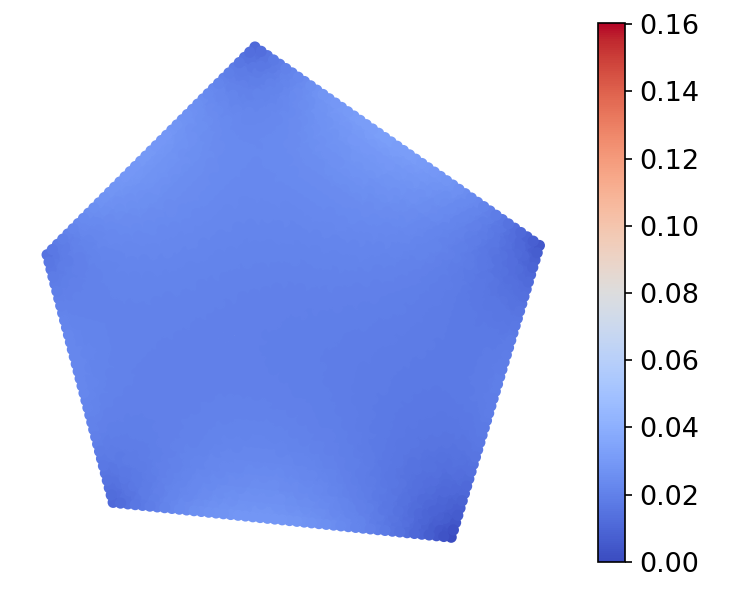}}  \\  
& \rotatebox[origin=l]{90}{\makecell{PI-PointNet \\ Worst case}} & {\includegraphics[width=0.16\textwidth]{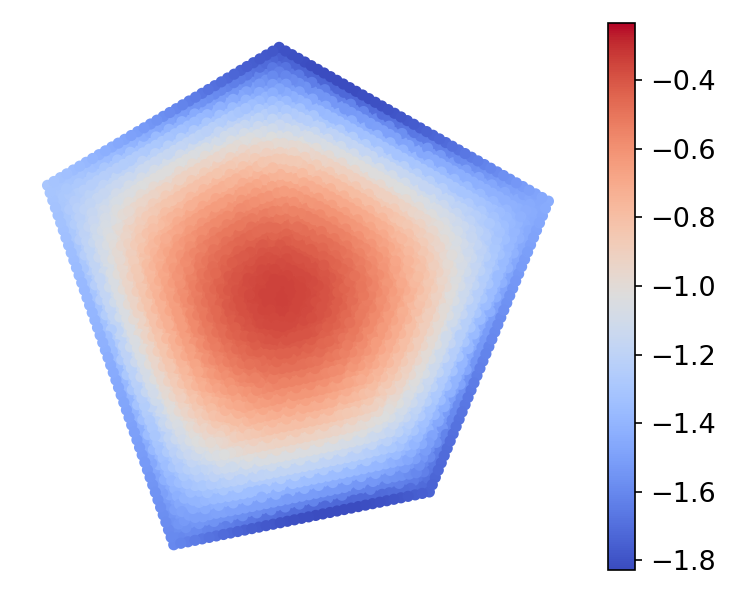}} & {\includegraphics[width=0.16\textwidth]{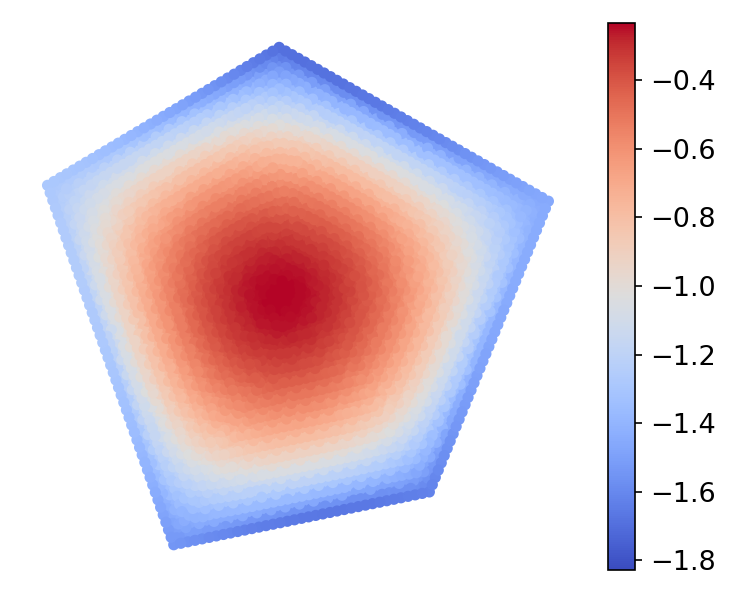}} & {\includegraphics[width=0.16\textwidth]{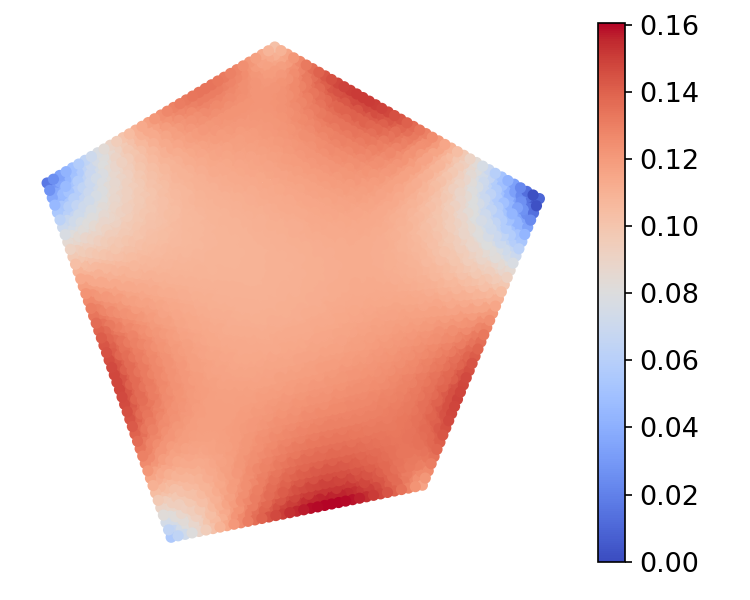}} & {\includegraphics[width=0.16\textwidth]{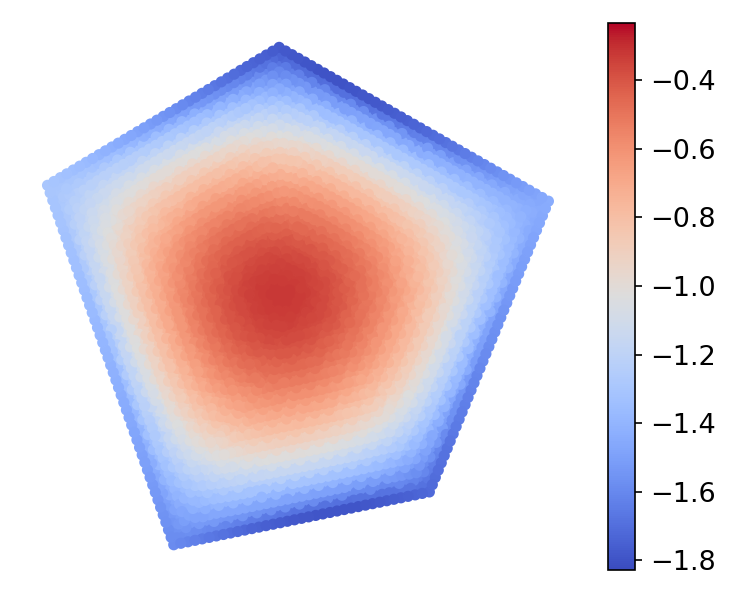}} & {\includegraphics[width=0.16\textwidth]{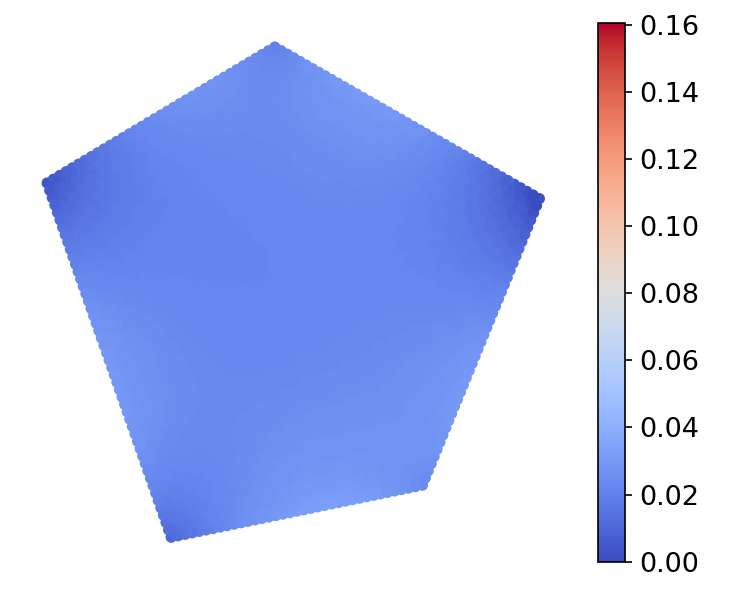}} \\
& \rotatebox[origin=l]{90}{\makecell{PI-GANO \\ Worst case}} & {\includegraphics[width=0.16\textwidth]{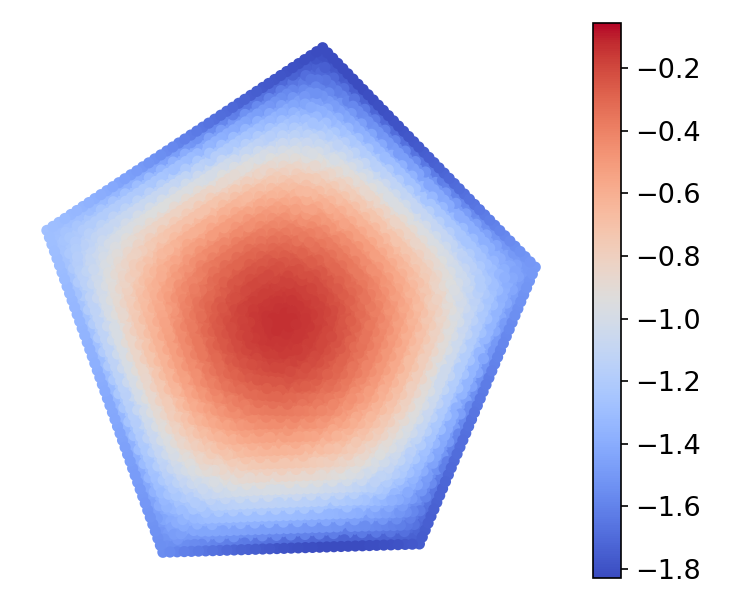}} &
 {\includegraphics[width=0.16\textwidth]{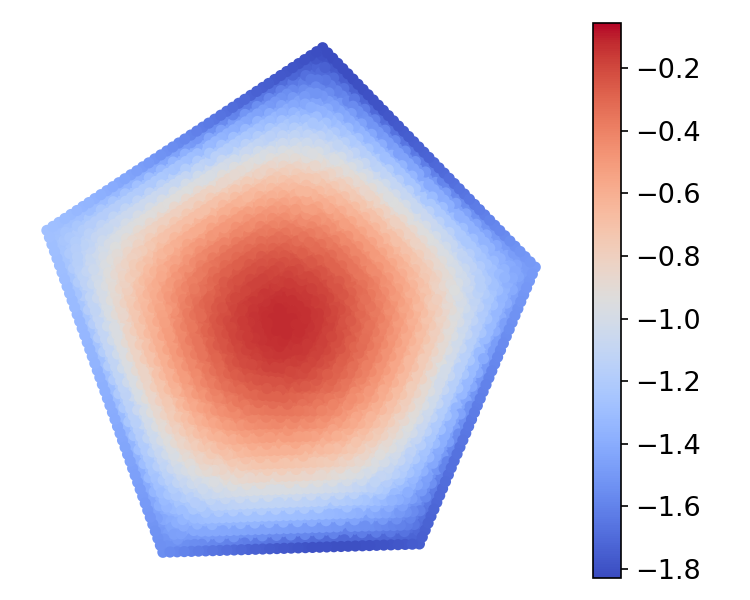}} & {\includegraphics[width=0.16\textwidth]{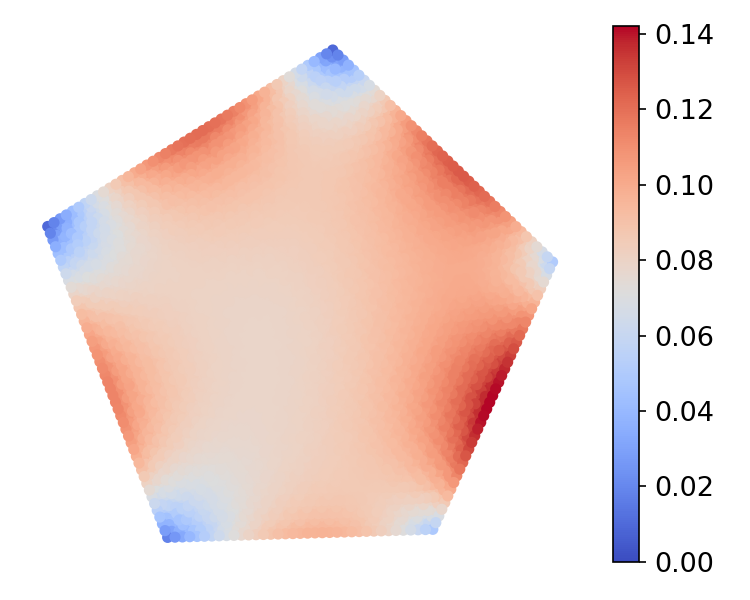}} &{\includegraphics[width=0.16\textwidth]{figs/samples/worst_DGKM_pred_Darcy_star_DG_only_geo.png}} & {\includegraphics[width=0.16\textwidth]{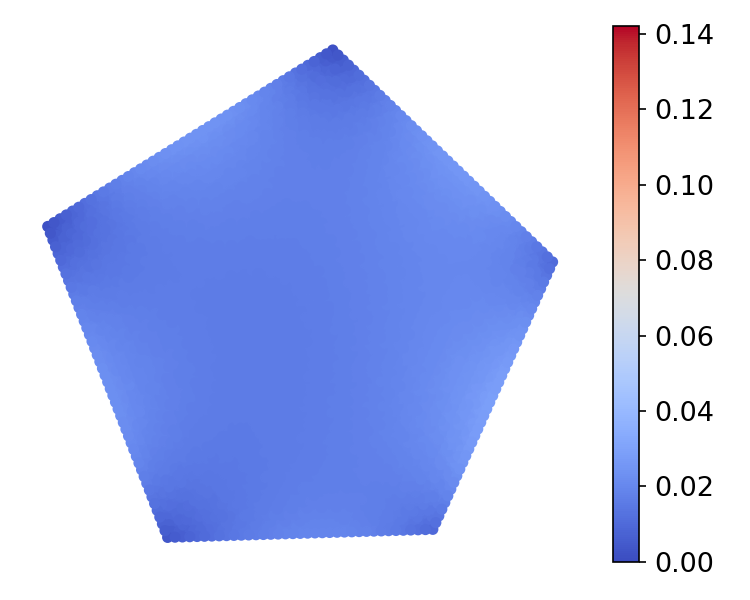}}  \\ 
\hline
\multirow{8}{*}{\rotatebox[origin=c]{90}{2D Plate}} & \rotatebox[origin=l]{90}{\makecell{PI-PointNet \\ Best case}} & {\includegraphics[width=0.16\textwidth]{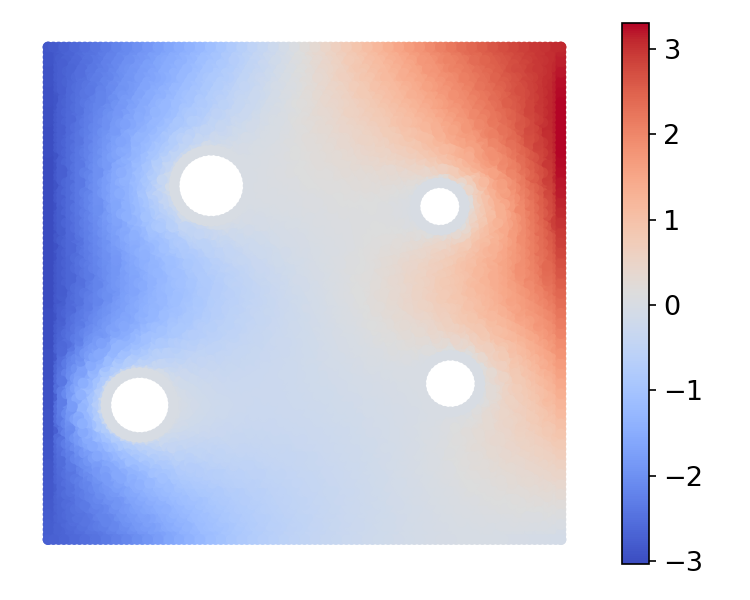}} & {\includegraphics[width=0.16\textwidth]{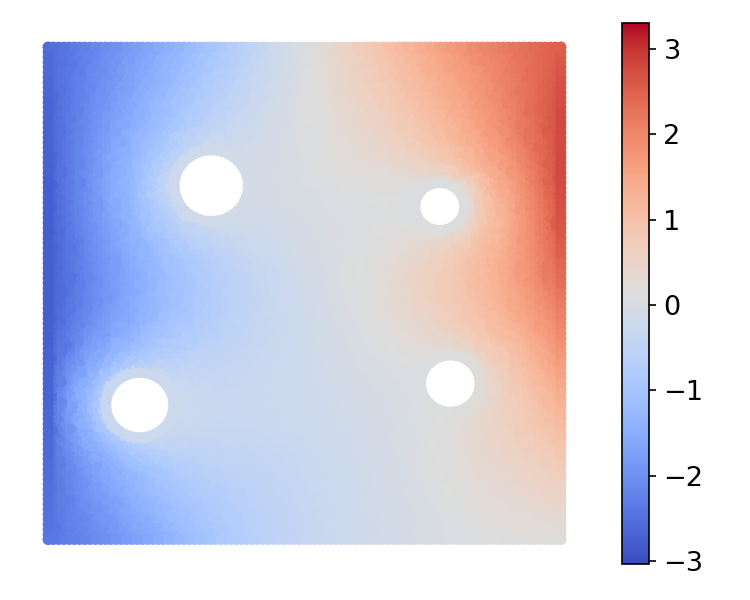}} & {\includegraphics[width=0.16\textwidth]{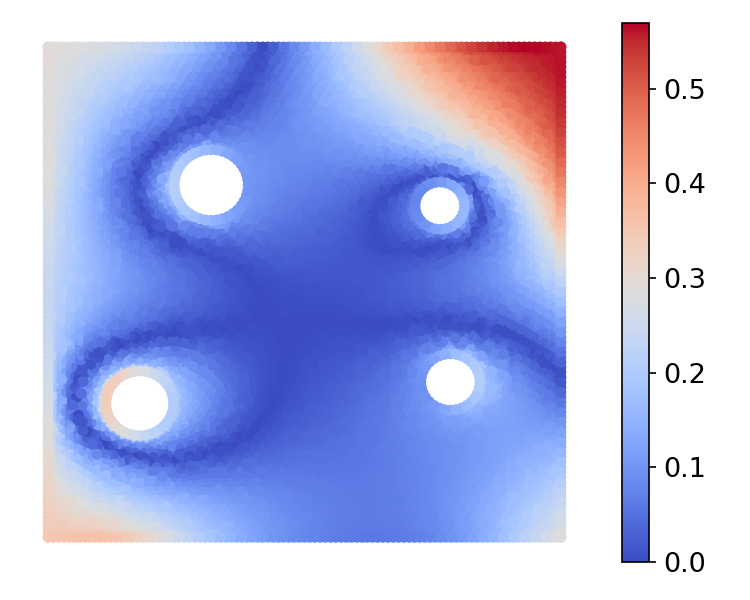}} & 
{\includegraphics[width=0.16\textwidth]{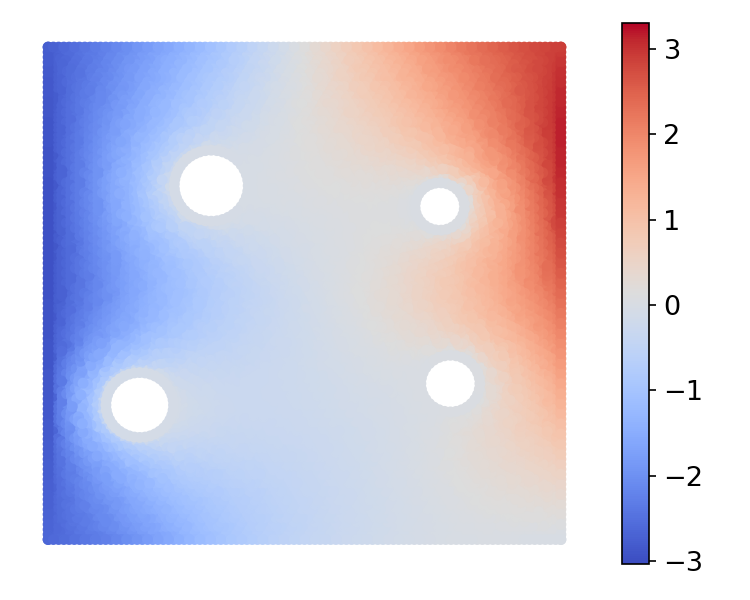}} & {\includegraphics[width=0.16\textwidth]{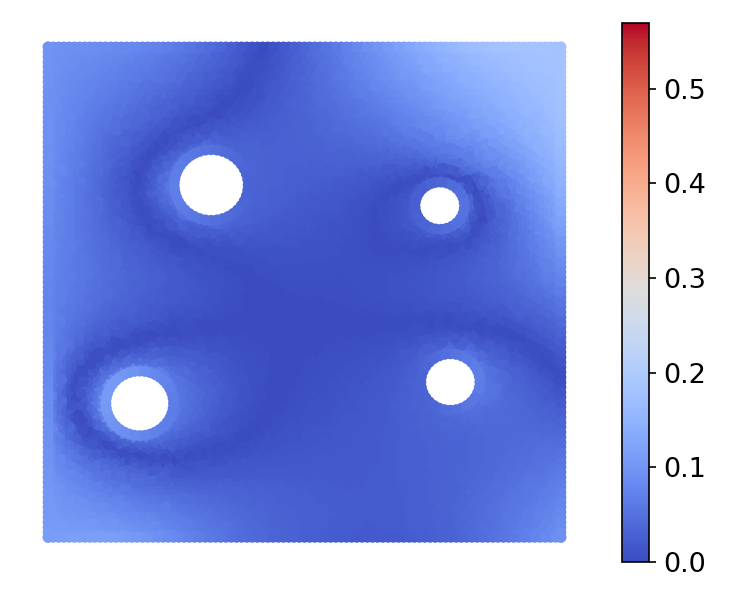}}\\
 & \rotatebox[origin=l]{90}{\makecell{PI-GANO \\ Best case}} & {\includegraphics[width=0.16\textwidth]{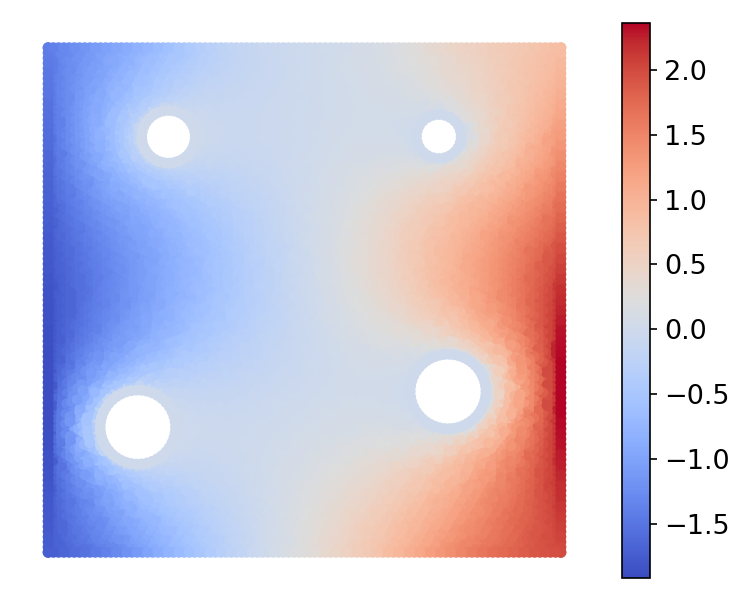}} & 
 {\includegraphics[width=0.16\textwidth]{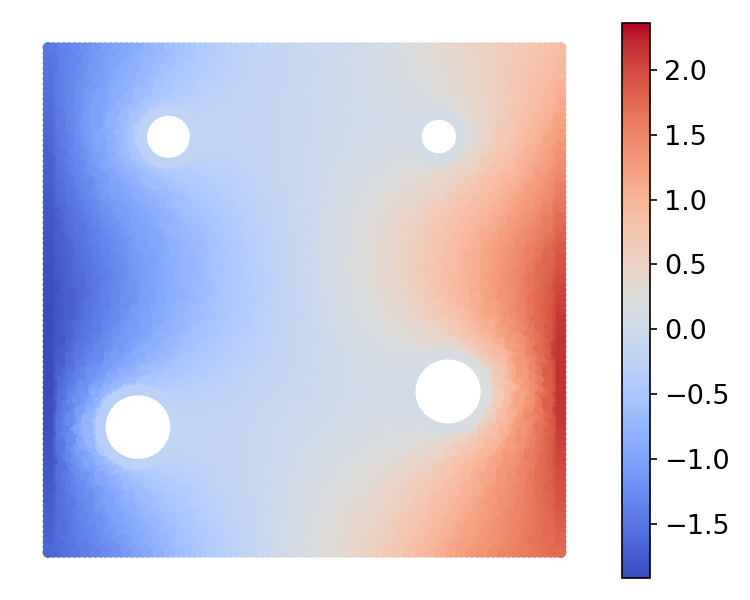}} & {\includegraphics[width=0.16\textwidth]{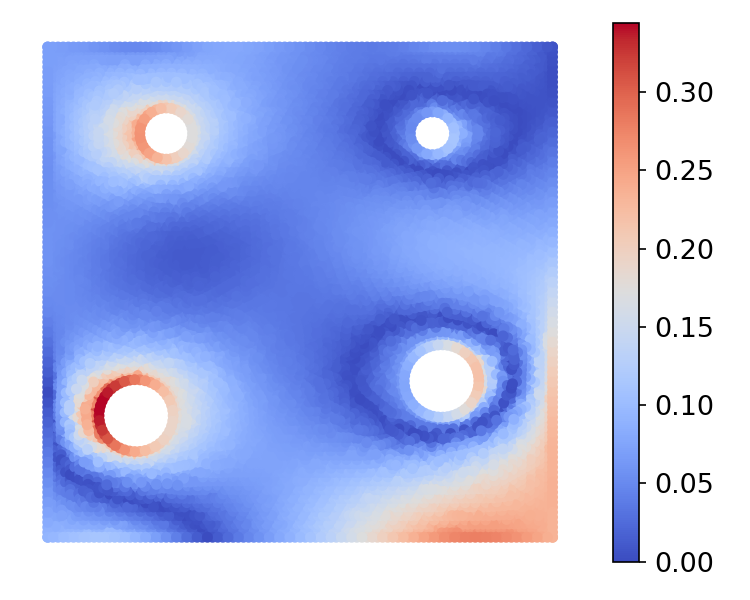}} &{\includegraphics[width=0.16\textwidth]{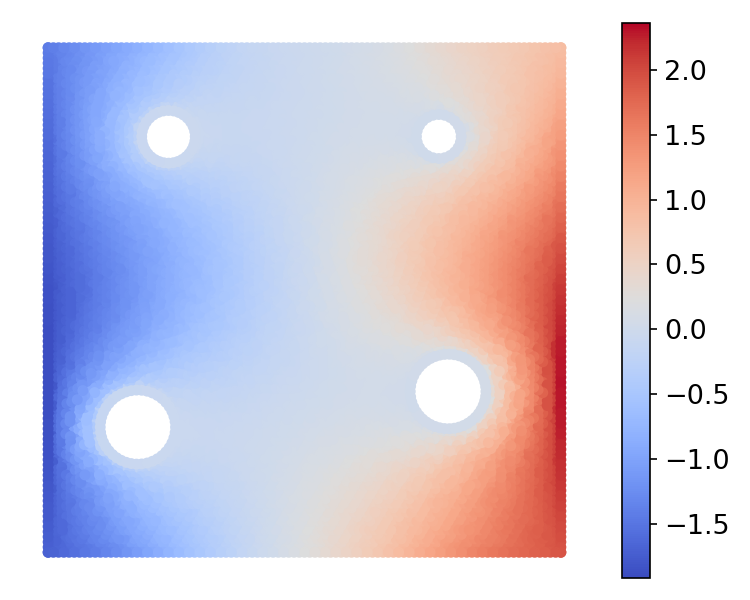}} & {\includegraphics[width=0.16\textwidth]{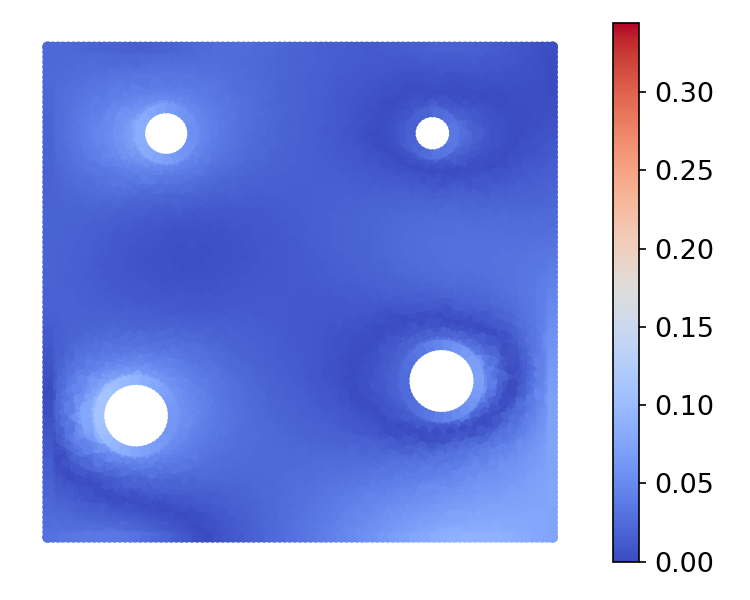}} \\ 
 & \rotatebox[origin=l]{90}{\makecell{PI-PointNet \\ Worst case}}& {\includegraphics[width=0.16\textwidth]{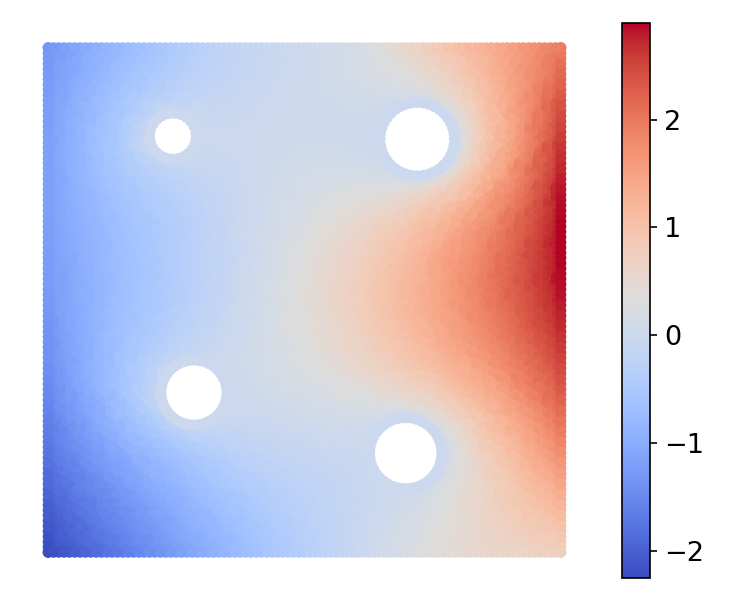}} & {\includegraphics[width=0.16\textwidth]{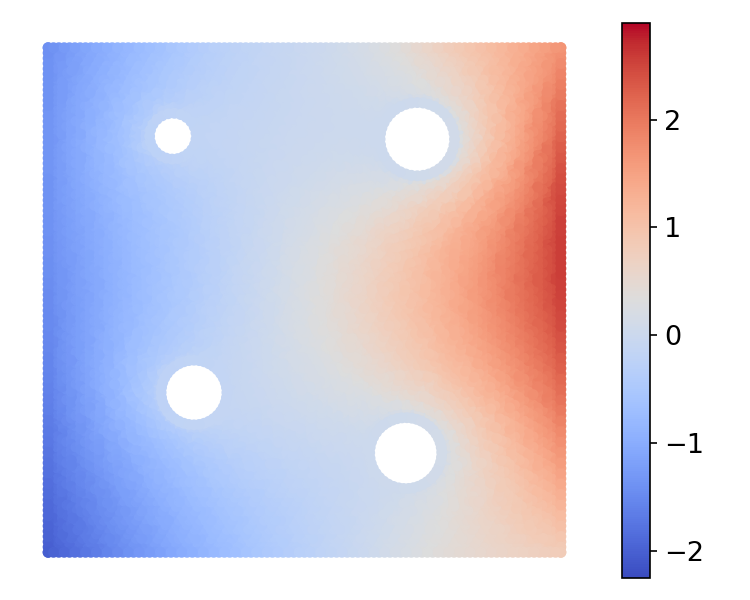}} & {\includegraphics[width=0.16\textwidth]{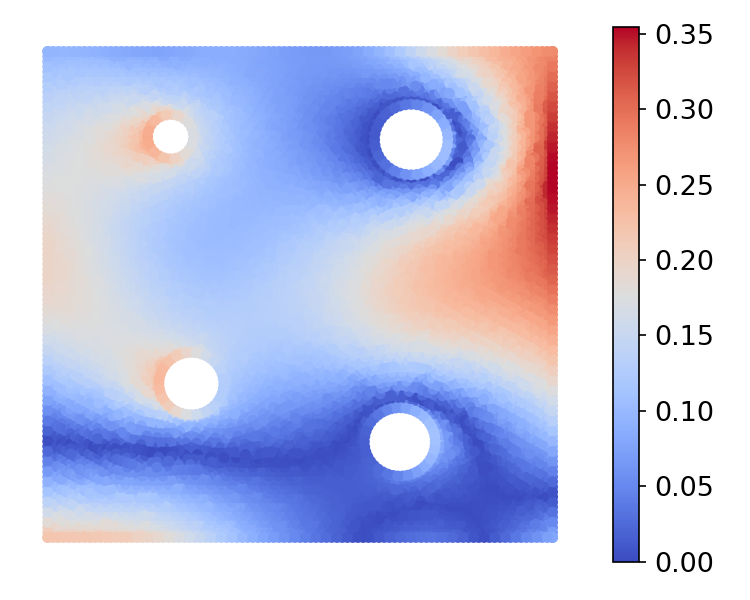}} & {\includegraphics[width=0.16\textwidth]{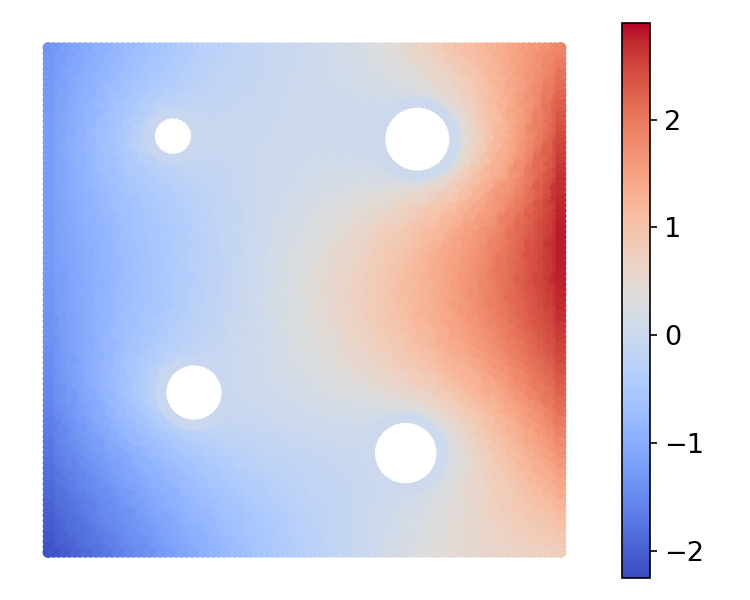}} & {\includegraphics[width=0.16\textwidth]{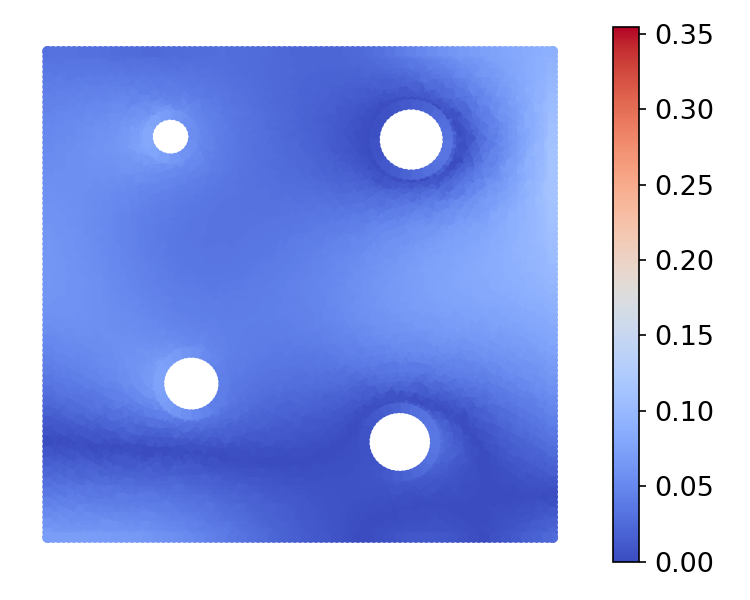}} \\
 & \rotatebox[origin=l]{90}{\makecell{PI-GANO \\ Worst case}} & {\includegraphics[width=0.16\textwidth]{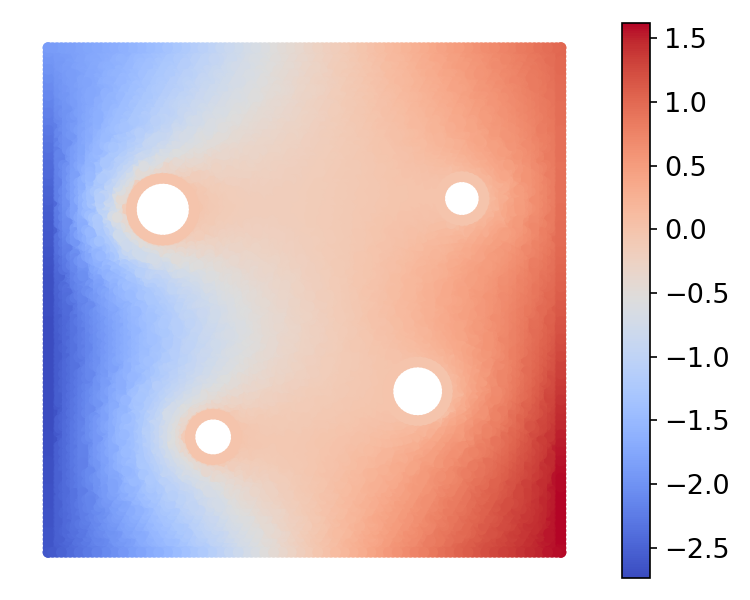}} & 
 {\includegraphics[width=0.16\textwidth]{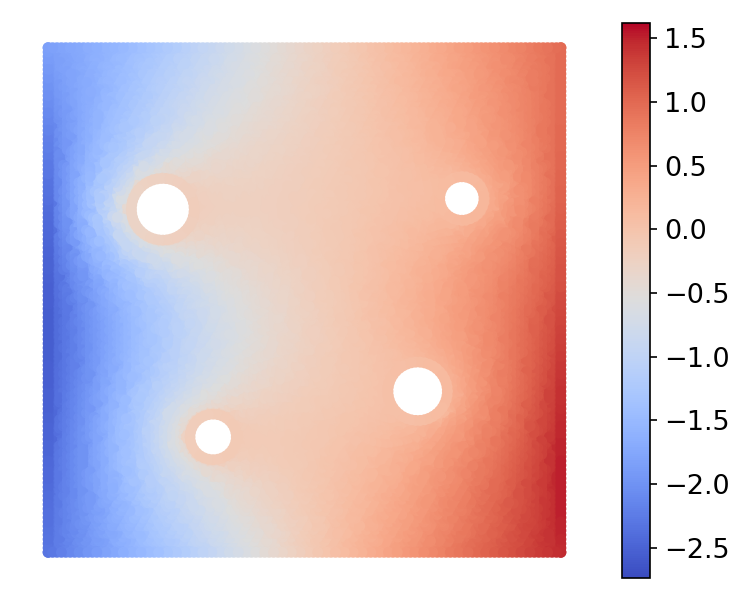}} & {\includegraphics[width=0.16\textwidth]{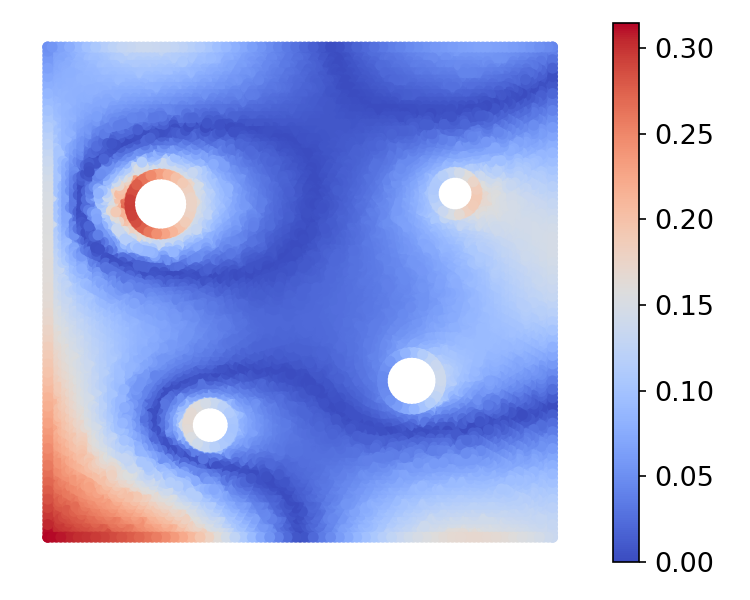}} &{\includegraphics[width=0.16\textwidth]{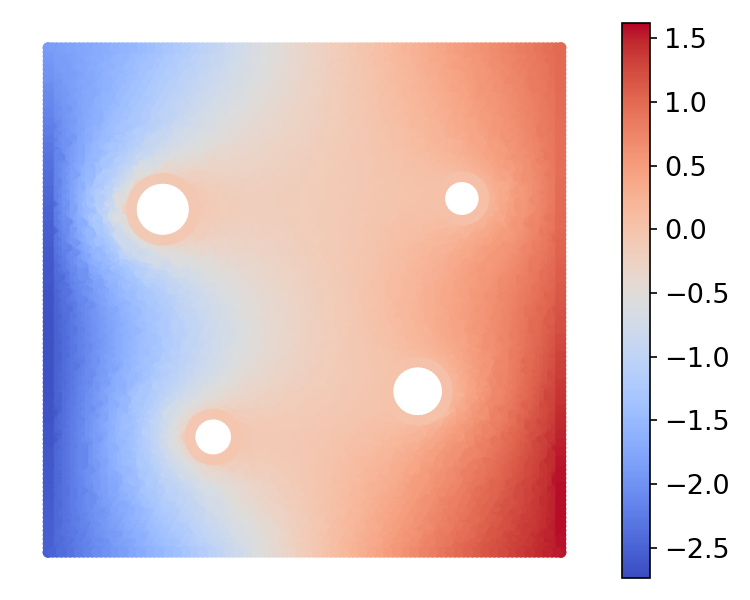}} & {\includegraphics[width=0.16\textwidth]{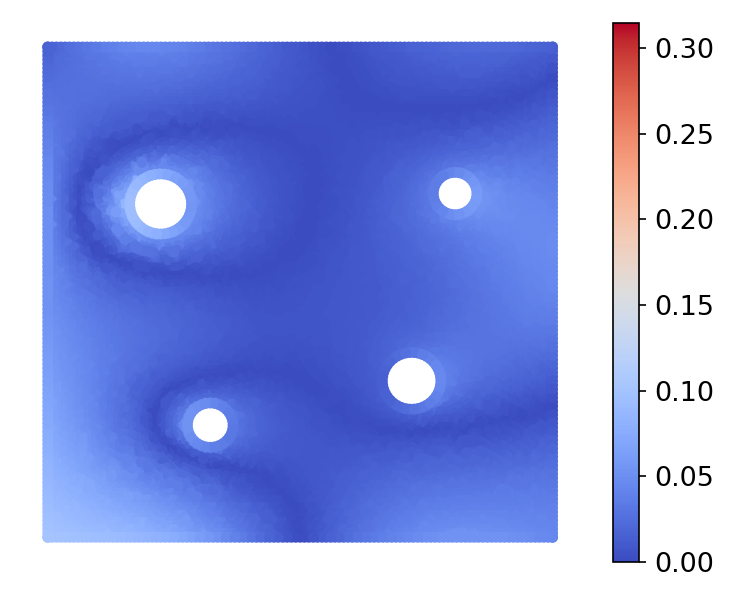}}  \\ 
 \hline
\end{tabular}
\label{fig.best_worst_cases_only_geo}
\end{table*}

\begin{figure}[ht]
    \centering
            \begin{subfigure}[b]{0.3\textwidth}
            \centering
            \includegraphics[width=\textwidth]{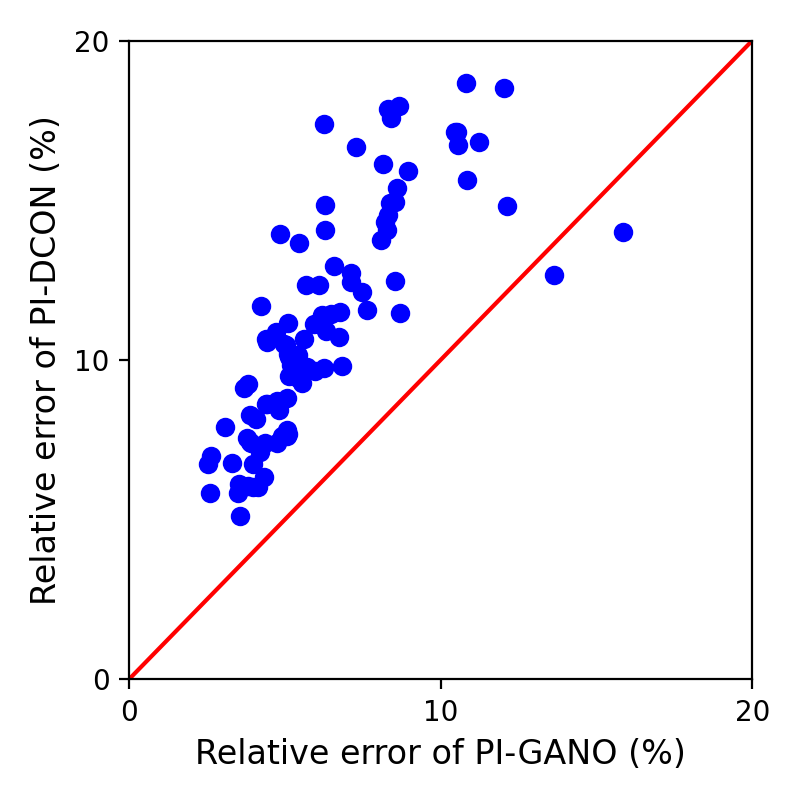}
            \caption{Darcy flow}    
            \label{fig:GA_vs_DCON_darcy}
        \end{subfigure}
        \hspace{0.2cm}
        \begin{subfigure}[b]{0.3\textwidth}   
            \centering 
            \includegraphics[width=\textwidth]{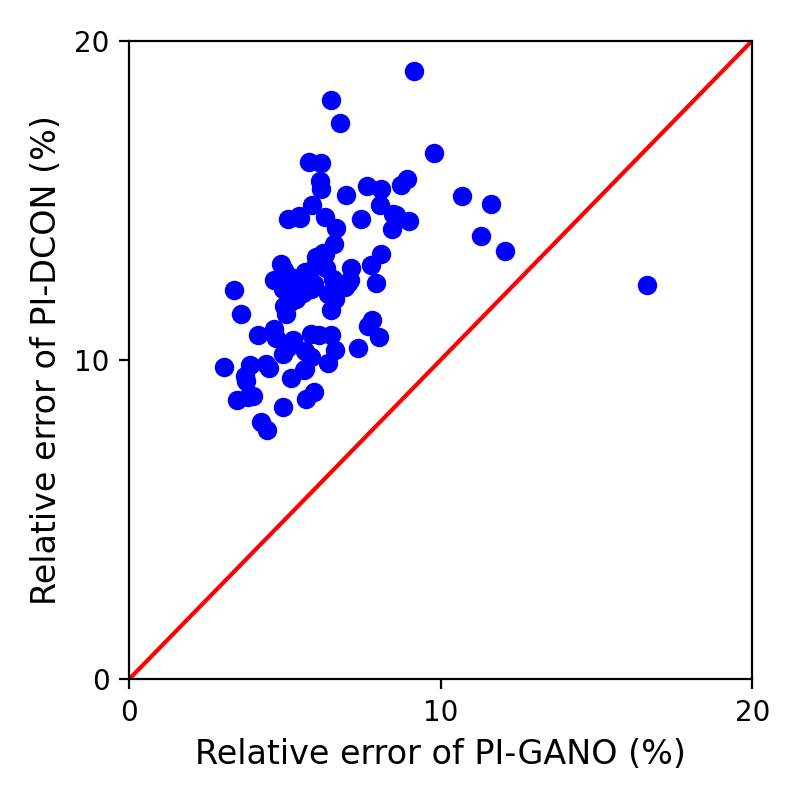}
            \caption{2D plate}  
            \label{fig:GA_vs_DCON_plate}
        \end{subfigure}
    \caption{\footnotesize Comparison between PI-GANO and PI-DCON on the Darcy flow and 2D plate problems. It can be seen that in all the cases PI-GANO offers the smaller errors.  }
    \label{fig.err}
\end{figure}

For cases where both PDE parameters and geometries vary, we compare our method with PI-DCON and PI-PointNet* in Table \ref{table.PI_com}. It can be seen that the PI-PointNet* struggles to achieve satisfactory accuracy, underscoring the challenge of developing a model capable of generalizing to varying  PDE parameters and geometries. Our proposed model demonstrated highest  accuracy in addressing both the Darcy flow problem and the 2D Plate problem. Compared to PI-DCON, our model achieved a 60\% accuracy improvement for the Darcy flow problem and a 50\% accuracy improvement for the 2D Plate problem, illustrating the effectiveness of the geometry encoder in our model. Figure \ref{fig.err} shows more detailed comparison between the errors calculated for  each testing samples, and underlines the superiority of PI-GANO in all the test cases.

\begin{table*}[!ht]
\begin{center}
\caption{Accuracy comparison between PI-GANO and PI-DCON.}
\begin{tabular}{c c c c c c c}
\hline
{ } & \multicolumn{3}{c}{\makecell{Mean of relative error}} & \multicolumn{3}{c}{\makecell{Standard deviation of relative error}} \\
\cline{2-7}
Dataset & PI-PointNet* & PI-DCON  & PI-GANO  & PI-PointNet* & PI-DCON & PI-GANO\\
\hline
\makecell{Darcy flow} & 64.3\% & 13.50\%  & \bf{8.40\%}  & 24.7\% & 5.88\% & \bf{2.53\%}  \\
\makecell{2D Plate} & 74.3\% & 14.06\%  & \bf{7.12\%}  & 27.8\% & 4.81\% & \bf{2.82\%} \\
\hline
\label{table.PI_com}
\end{tabular}
\end{center}
\end{table*}

To better illustrate the performance of our model, Table \ref{fig.best_worst_cases} shows the predictions from PI-DCON and PI-GANO and their errors compared to the FEM prediction, for various best and worst-case scenarios. Overall, our model shows higher accuracy  maintaining acceptable performance even in its worst cases. Furthermore, the relatively small performance gaps between the worst and best cases underscore the robustness of our model.

\begin{table*}[!ht]
\centering 
\caption{Comparison between the performance of PI-GANO and PI-DCON. Out of all realizations of boundary conditions, the ones that causes best and worst performances of each model are shown.}
\begin{tabular}{|c|c|c | c c | c c|}
\hline
\multicolumn{2}{|c|}{ } & \multirow{2}{*}{Ground Truth} & \multicolumn{2}{c|}{PI-DCON} & \multicolumn{2}{c|}{PI-GANO} \\ 
\cline{4-7}
\multicolumn{2}{|c|}{ } &  & Prediction & Absolute Error & Prediction & Absolute Error \\ 
\hline
\multirow{8}{*}{\rotatebox[origin=c]{90}{Darcy flow }} & \rotatebox[origin=l]{90}{\makecell{PI-DCON \\ Best case}} & {\includegraphics[width=0.16\textwidth]{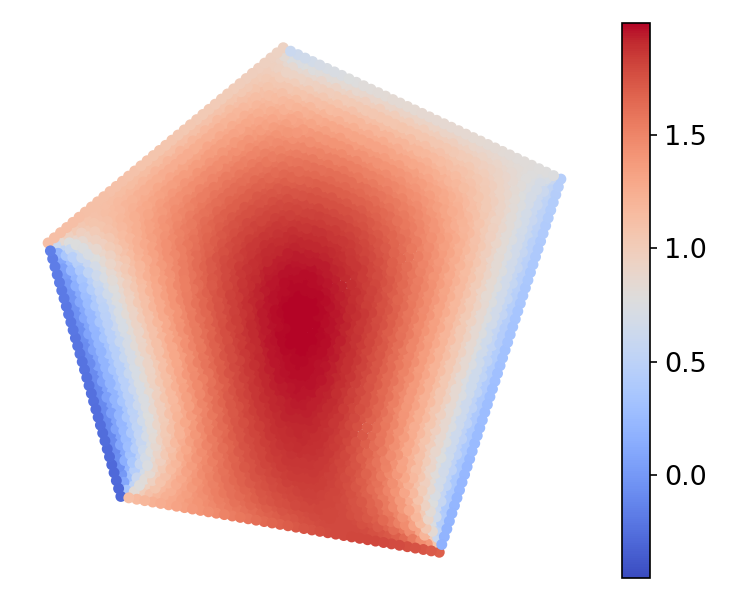}} & {\includegraphics[width=0.16\textwidth]{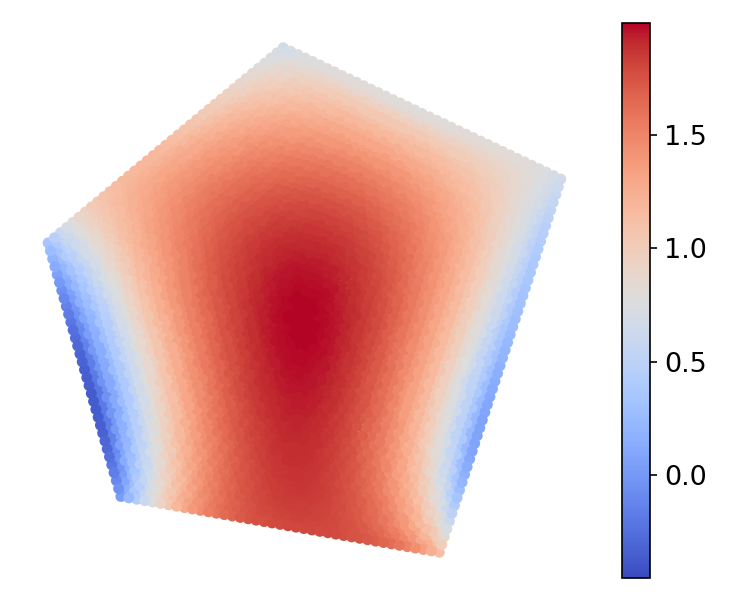}} & {\includegraphics[width=0.16\textwidth]{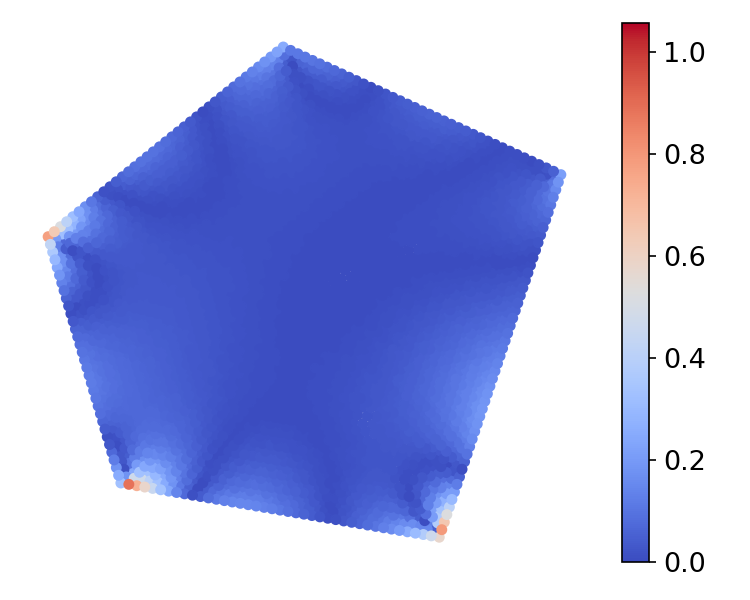}} & {\includegraphics[width=0.16\textwidth]{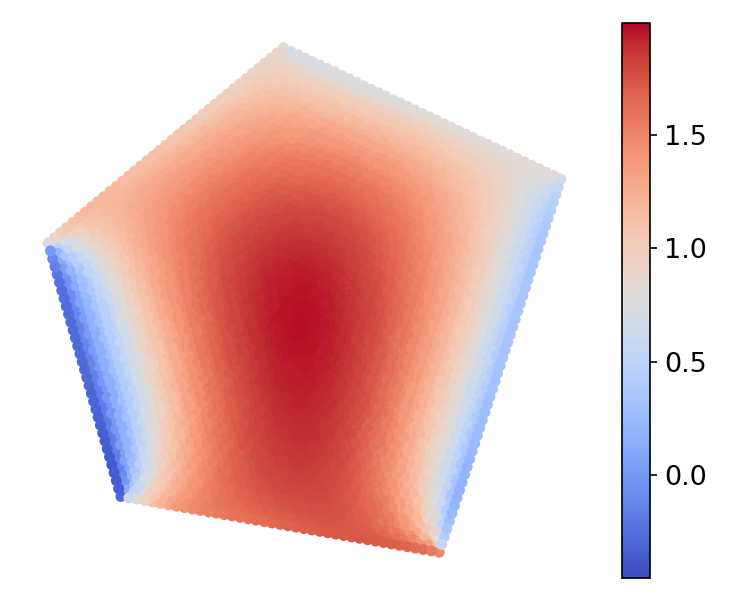}} & {\includegraphics[width=0.16\textwidth]{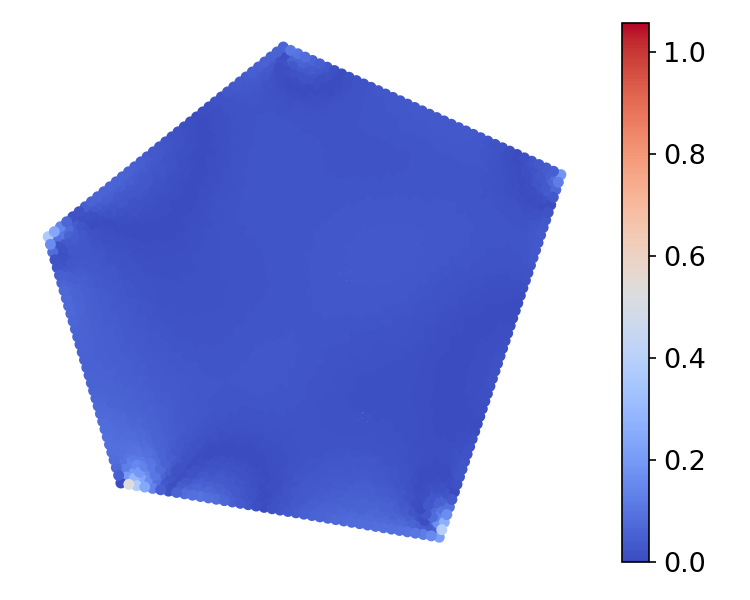}} \\
& \rotatebox[origin=l]{90}{\makecell{PI-GANO \\ Best case}} & {\includegraphics[width=0.16\textwidth]{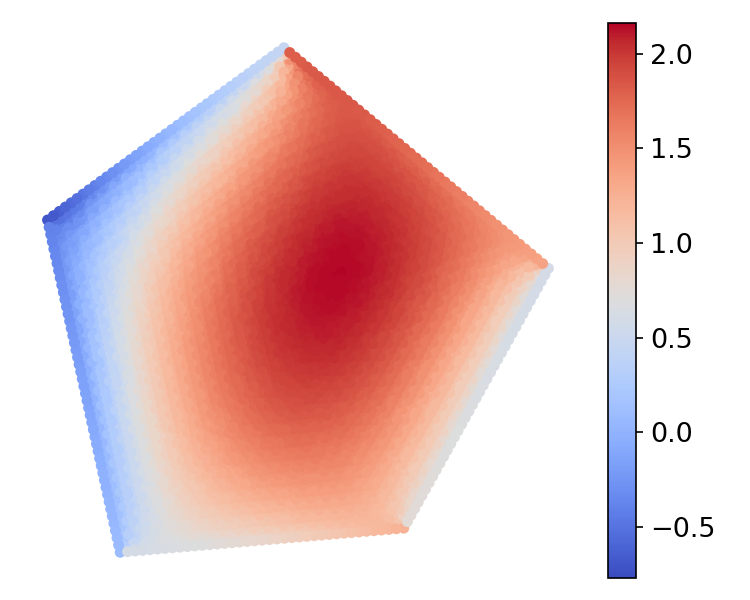}} & 
 {\includegraphics[width=0.16\textwidth]{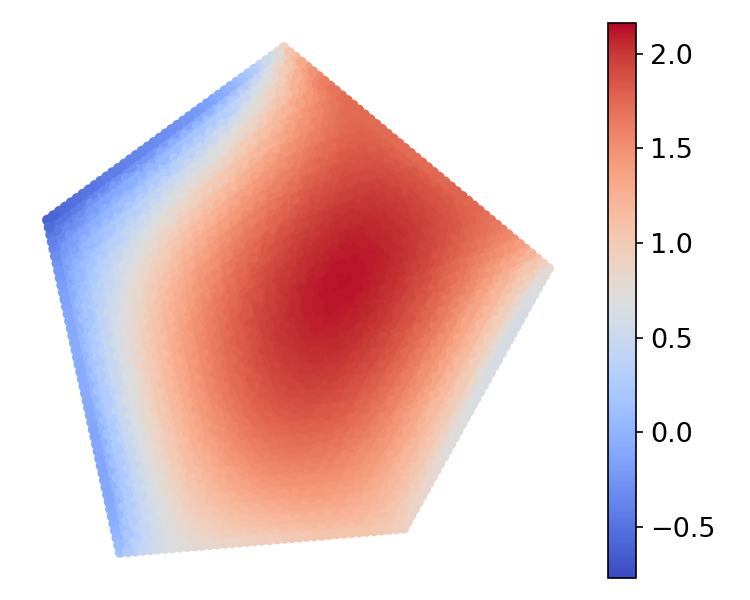}} & {\includegraphics[width=0.16\textwidth]{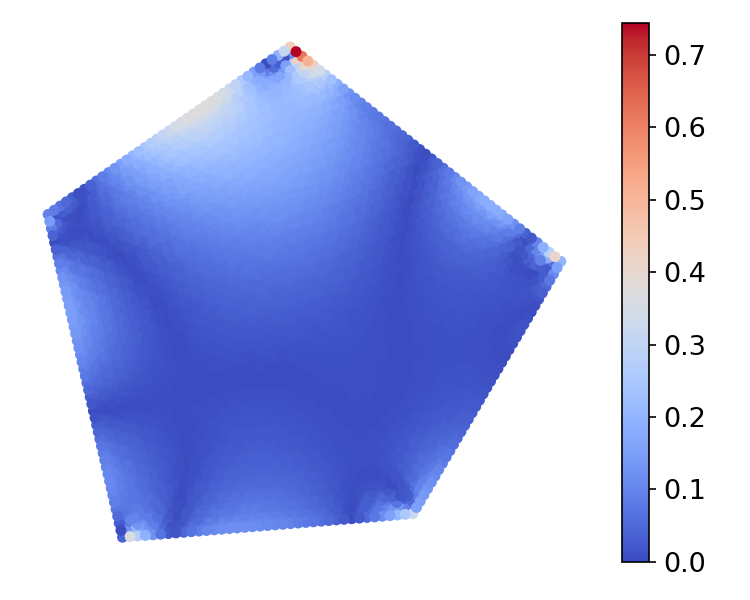}} &{\includegraphics[width=0.16\textwidth] {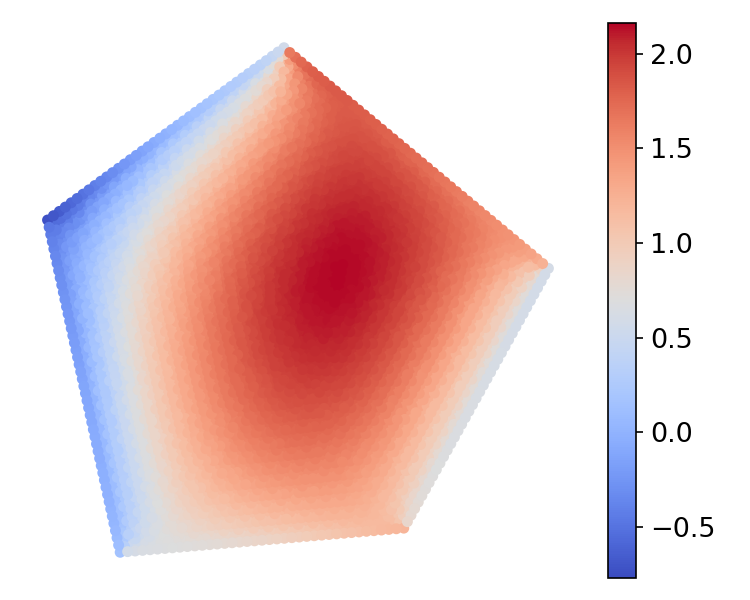}} & {\includegraphics[width=0.16\textwidth]{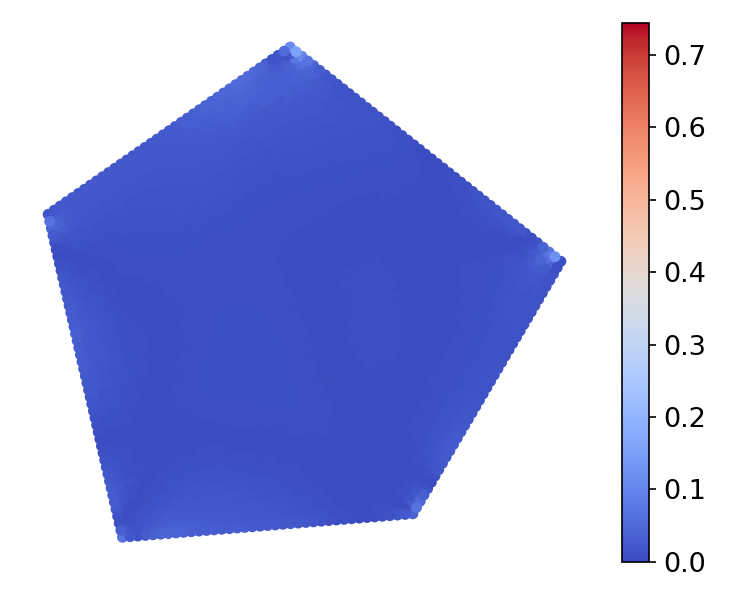}}  \\  
& \rotatebox[origin=l]{90}{\makecell{PI-DCON \\ Worst case}} & {\includegraphics[width=0.16\textwidth]{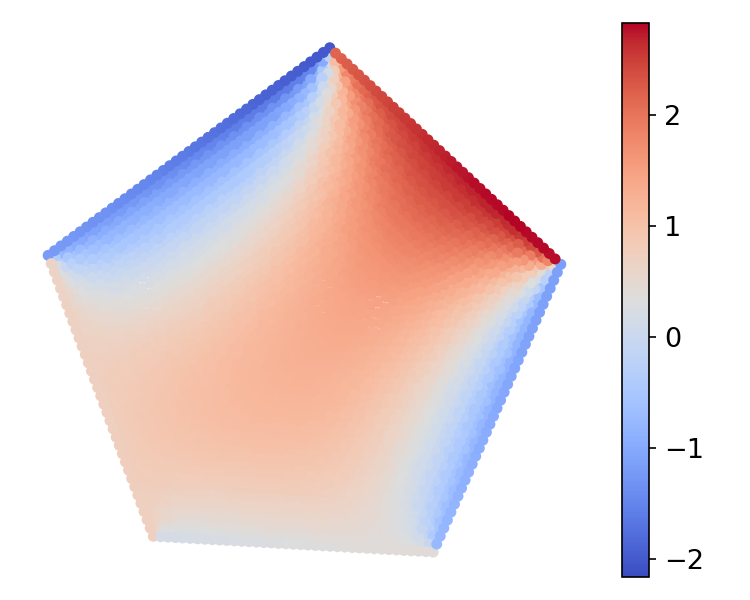}} & {\includegraphics[width=0.16\textwidth]{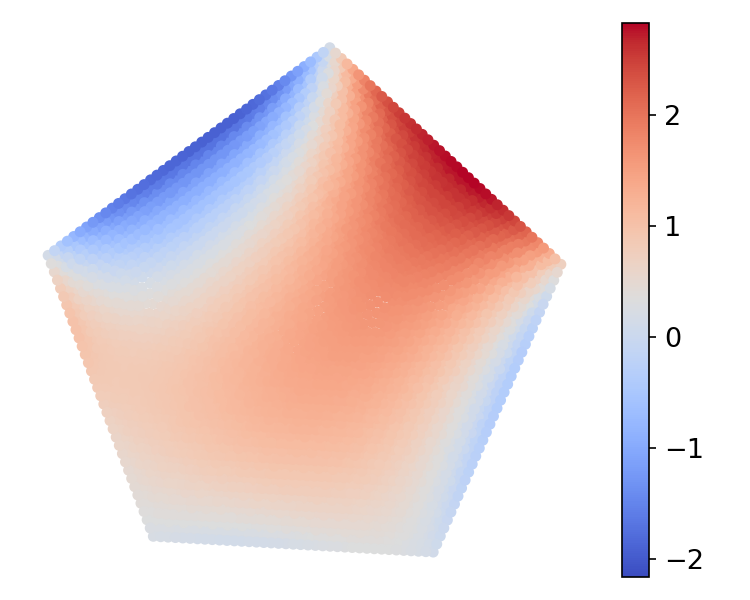}} & {\includegraphics[width=0.16\textwidth]{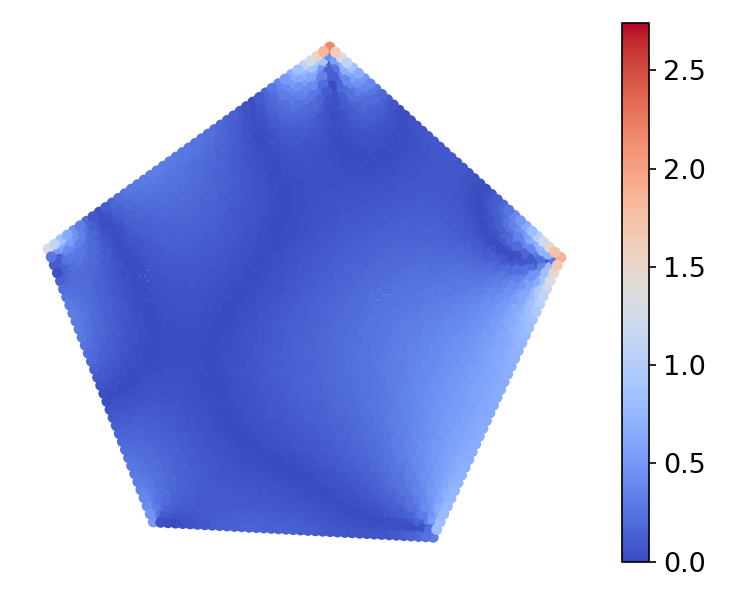}} & {\includegraphics[width=0.16\textwidth]{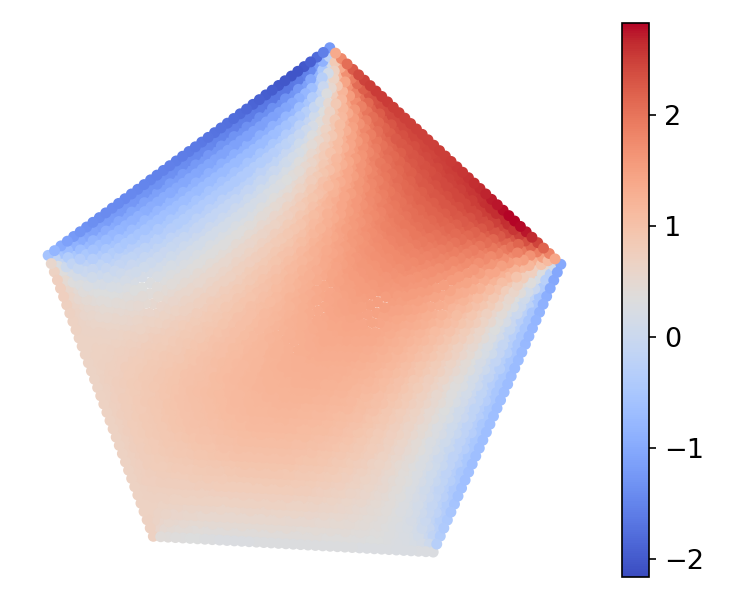}} & {\includegraphics[width=0.16\textwidth]{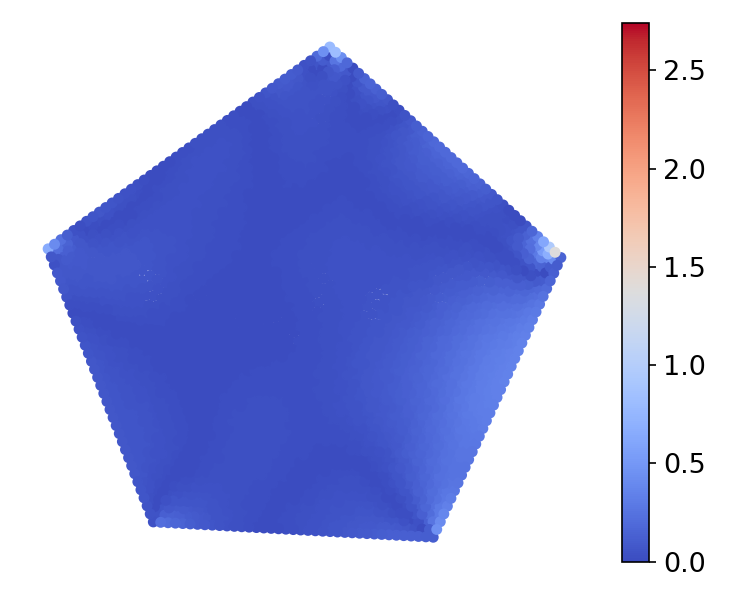}} \\
& \rotatebox[origin=l]{90}{\makecell{PI-GANO \\ Worst case}} & {\includegraphics[width=0.16\textwidth]{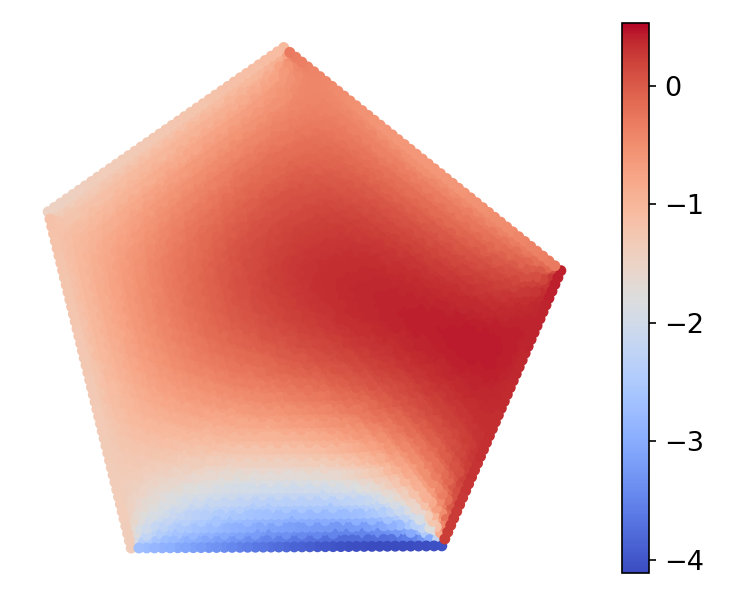}} &
 {\includegraphics[width=0.16\textwidth]{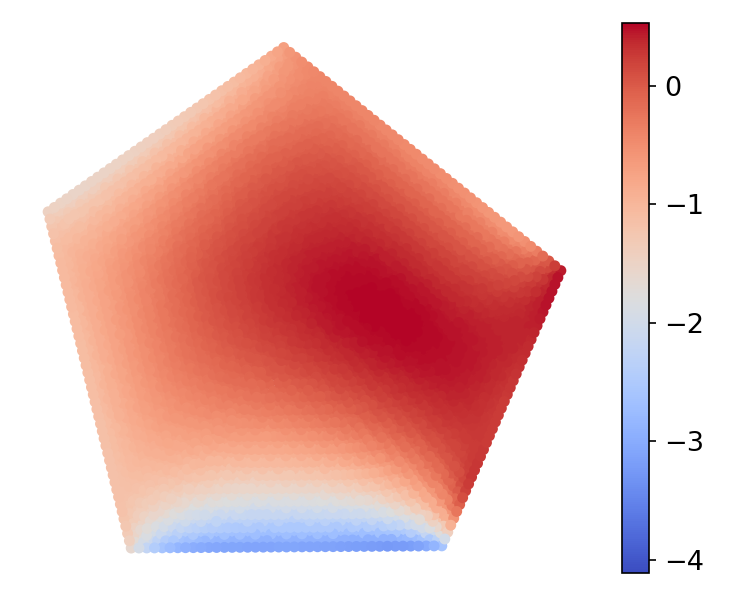}} & {\includegraphics[width=0.16\textwidth]{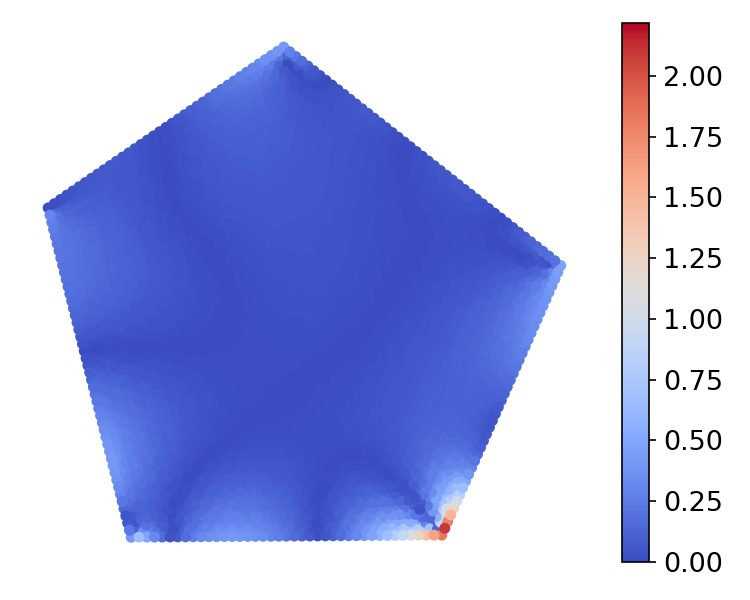}} &{\includegraphics[width=0.16\textwidth]{figs/samples/worst_DGKM_pred_Darcy_star_DG.png}} & {\includegraphics[width=0.16\textwidth]{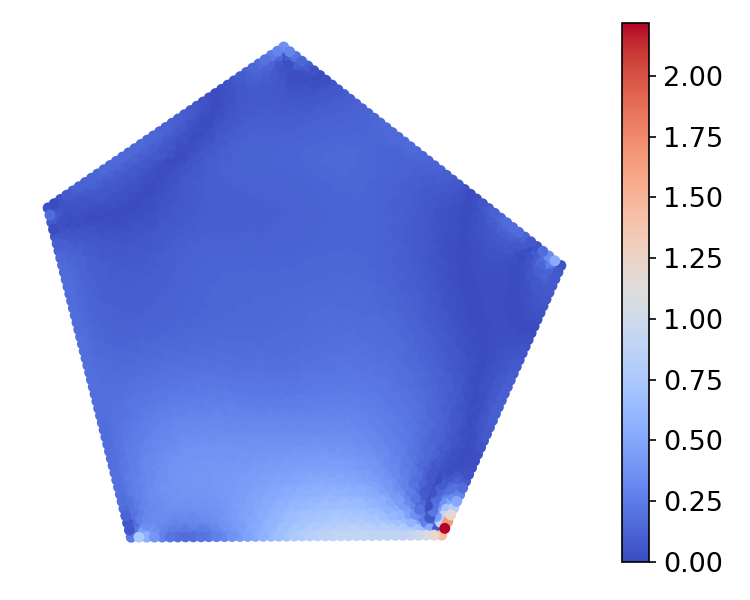}}  \\ 
\hline
\multirow{8}{*}{\rotatebox[origin=c]{90}{2D Plate}} & \rotatebox[origin=l]{90}{\makecell{PI-DCON \\ Best case}} & {\includegraphics[width=0.16\textwidth]{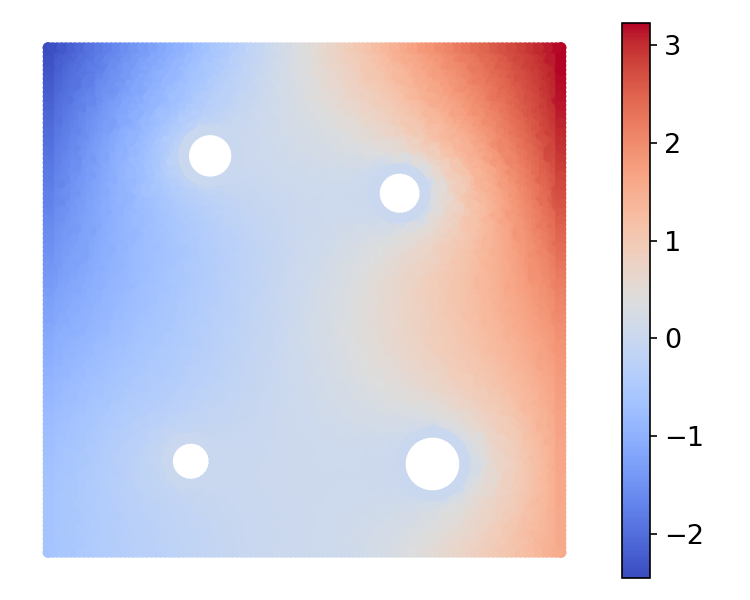}} & {\includegraphics[width=0.16\textwidth]{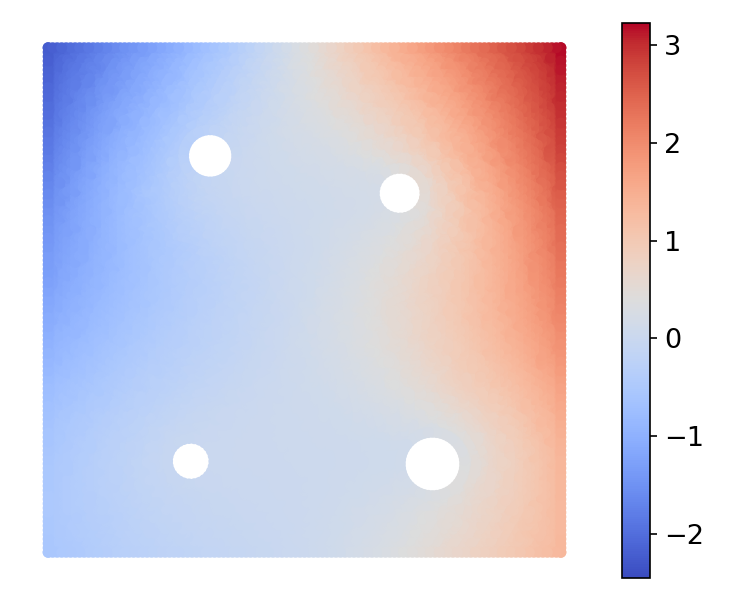}} & {\includegraphics[width=0.16\textwidth]{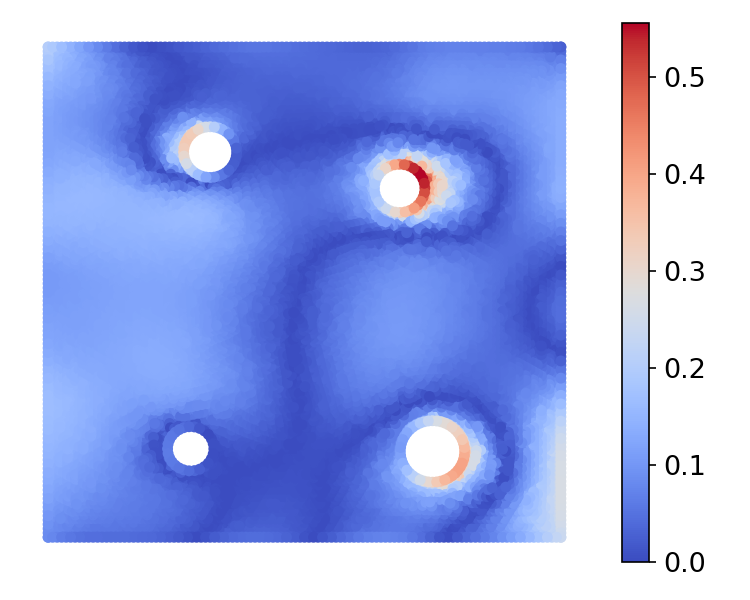}} & 
{\includegraphics[width=0.16\textwidth]{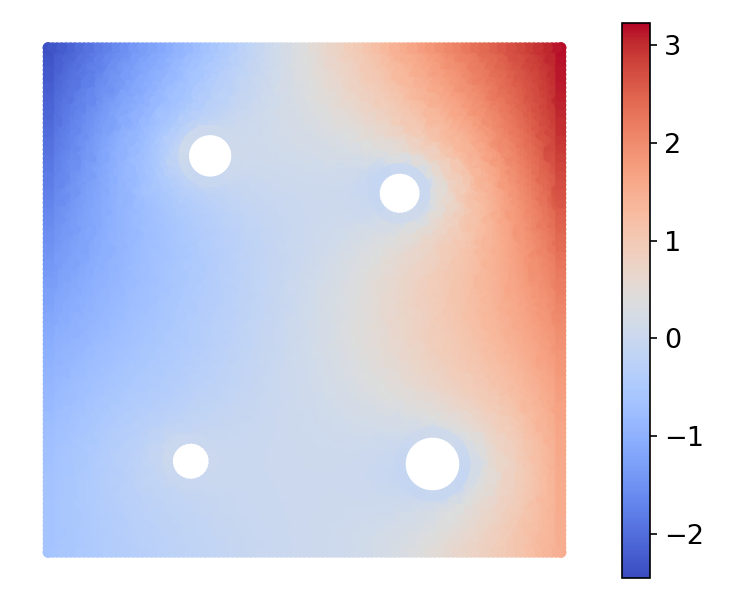}} & {\includegraphics[width=0.16\textwidth]{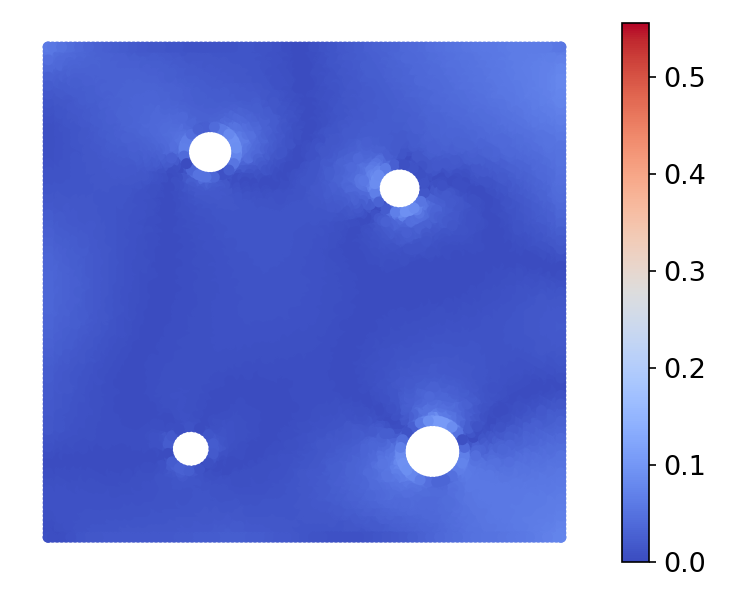}}\\
 & \rotatebox[origin=l]{90}{\makecell{PI-GANO \\ Best case}} & {\includegraphics[width=0.16\textwidth]{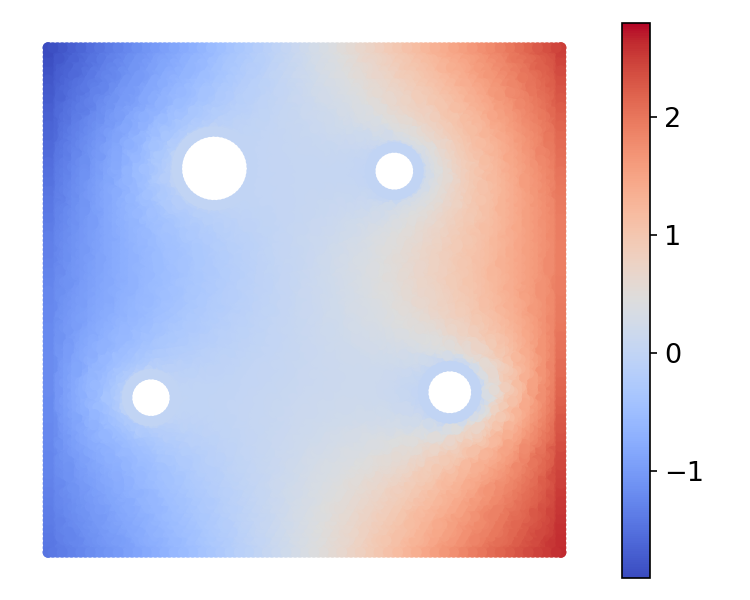}} & 
 {\includegraphics[width=0.16\textwidth]{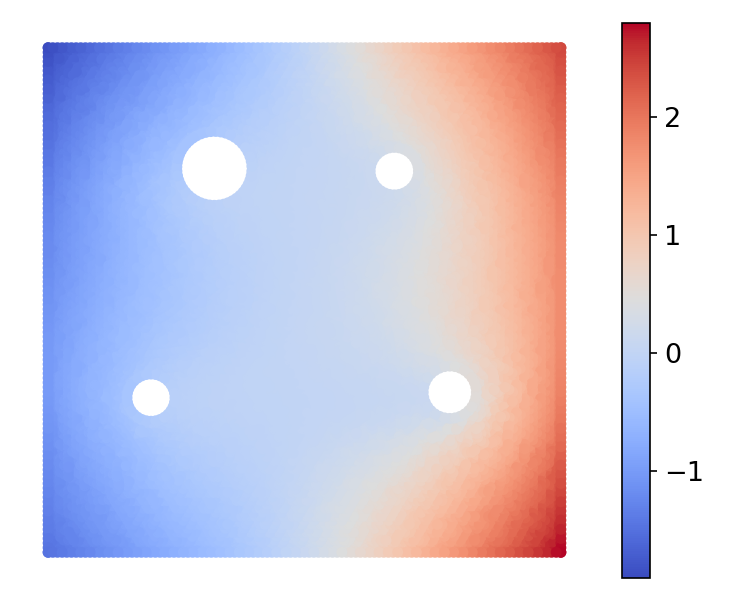}} & {\includegraphics[width=0.16\textwidth]{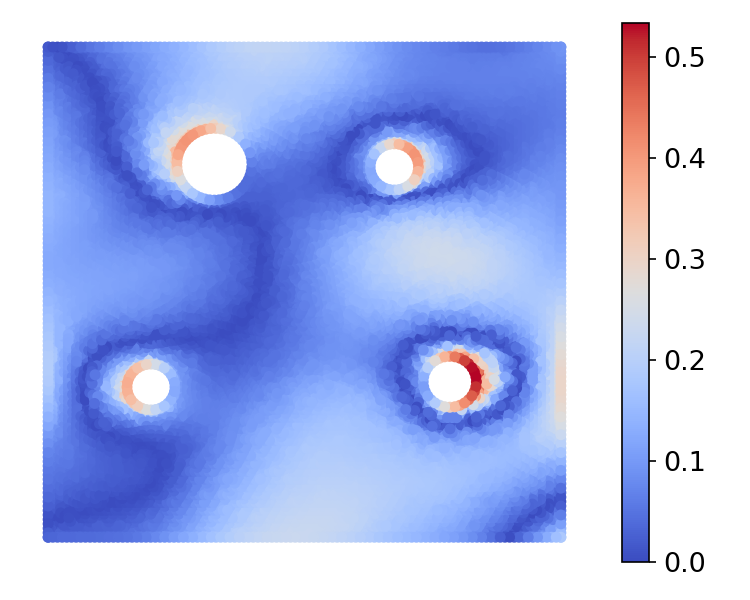}} &{\includegraphics[width=0.16\textwidth]{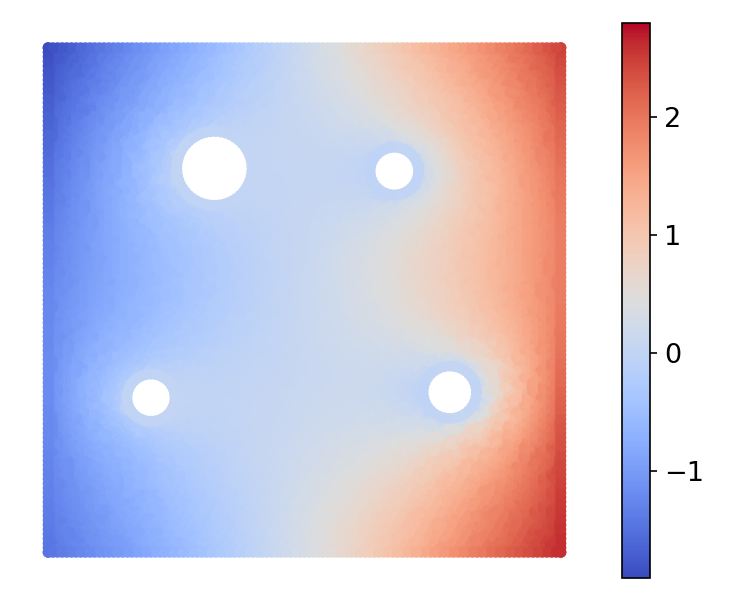}} & {\includegraphics[width=0.16\textwidth]{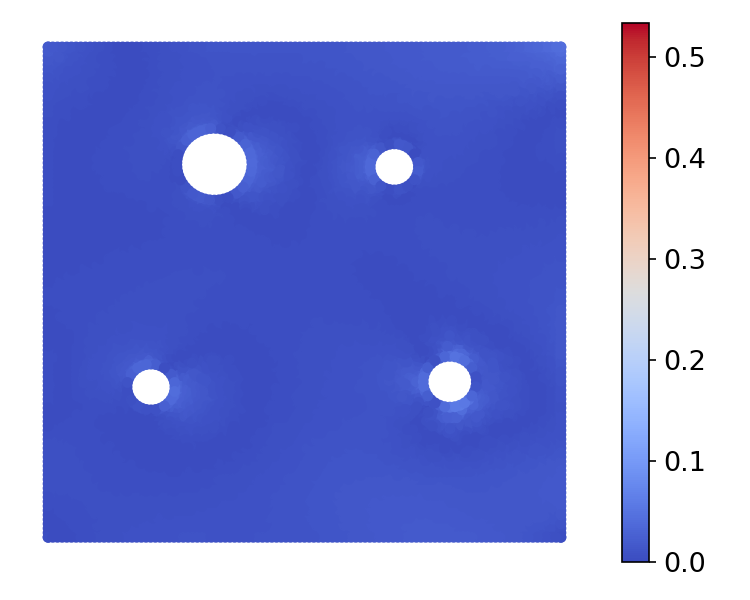}} \\ 
 & \rotatebox[origin=l]{90}{\makecell{PI-DCON \\ Worst case}}& {\includegraphics[width=0.16\textwidth]{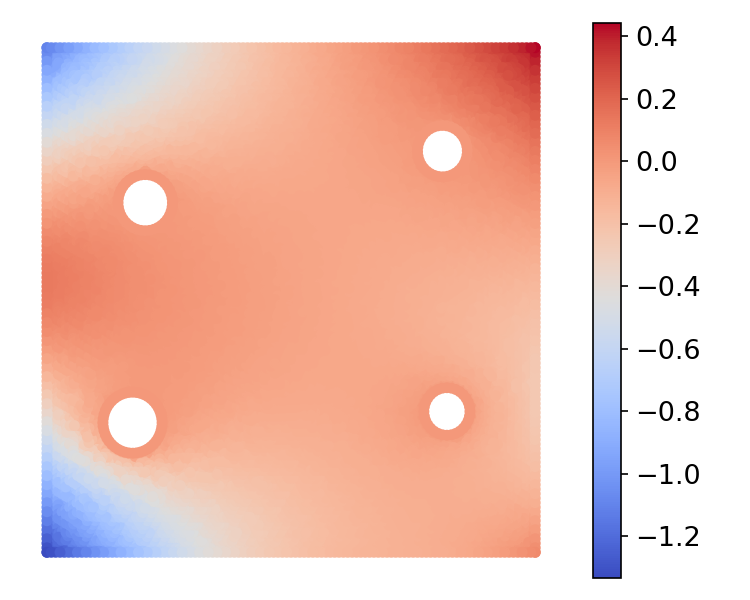}} & {\includegraphics[width=0.16\textwidth]{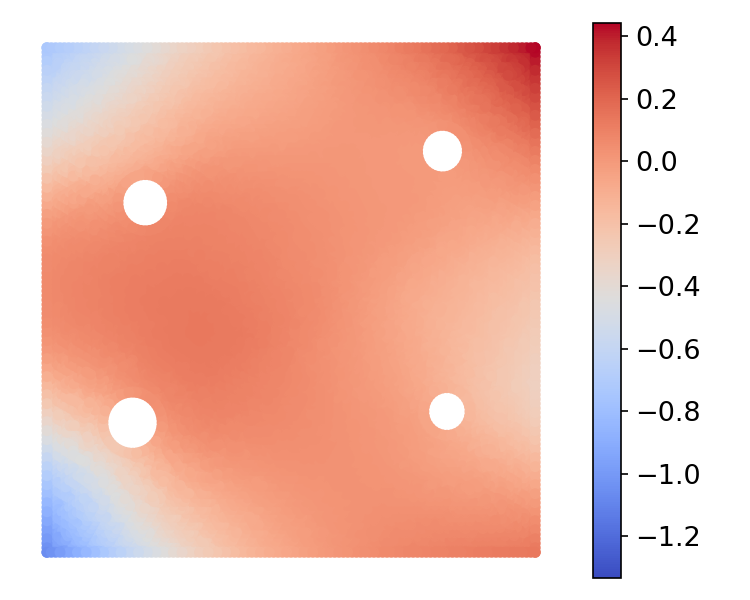}} & {\includegraphics[width=0.16\textwidth]{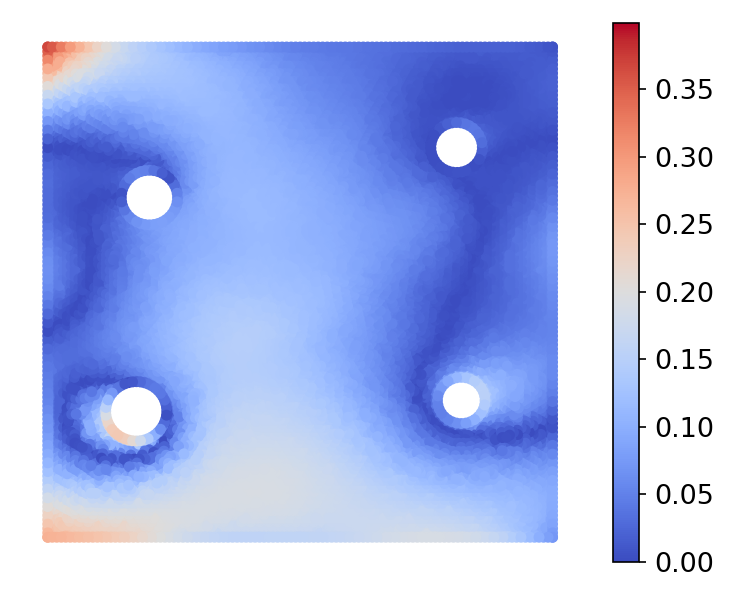}} & {\includegraphics[width=0.16\textwidth]{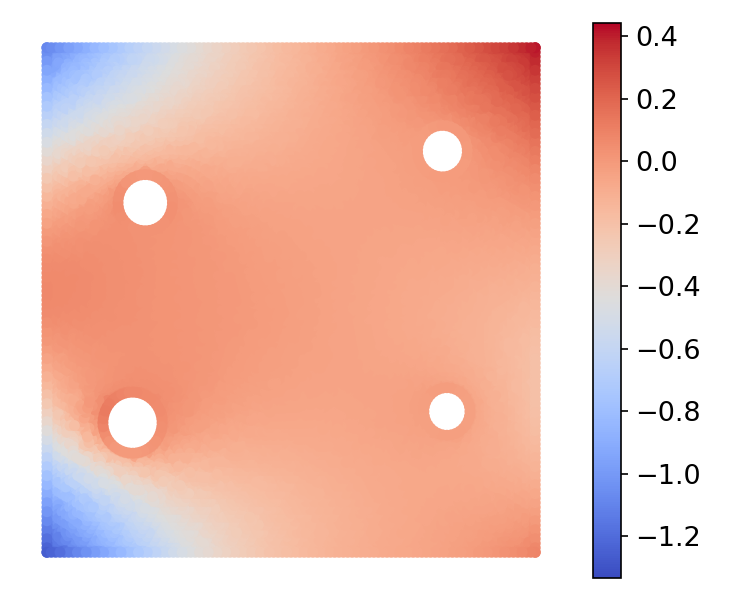}} & {\includegraphics[width=0.16\textwidth]{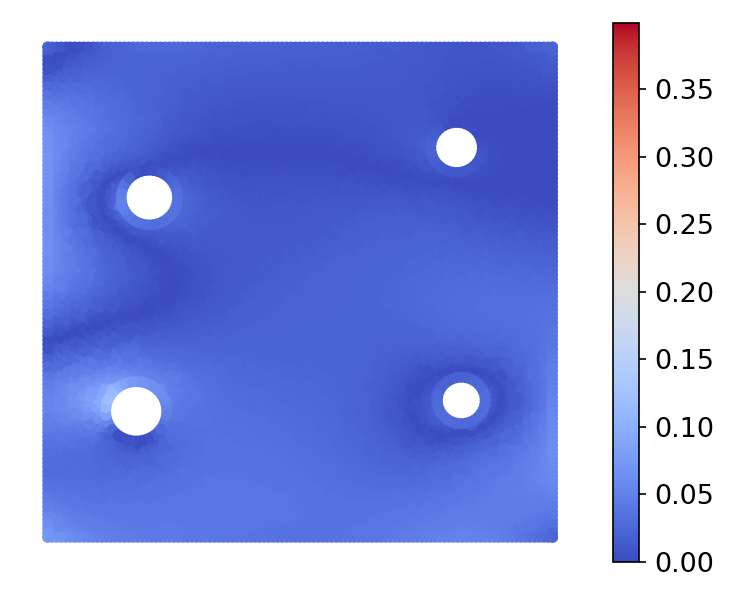}} \\
 & \rotatebox[origin=l]{90}{\makecell{PI-GANO \\ Worst case}} & {\includegraphics[width=0.16\textwidth]{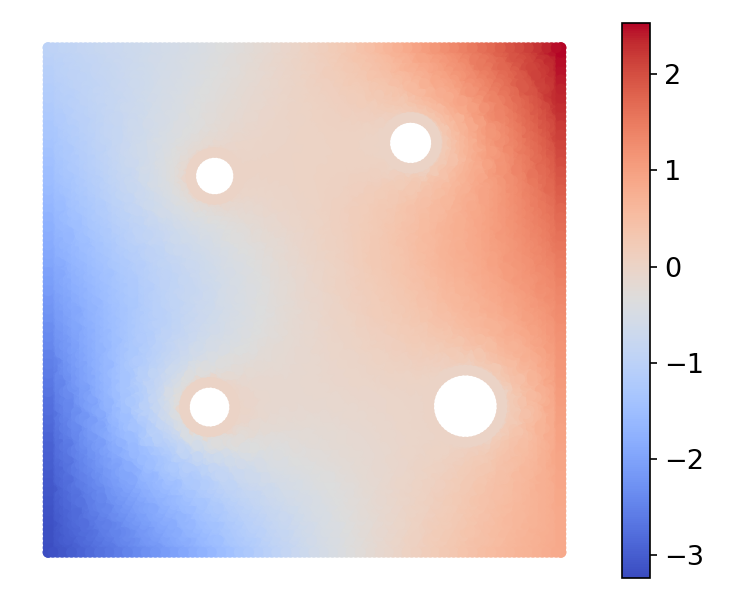}} & 
 {\includegraphics[width=0.16\textwidth]{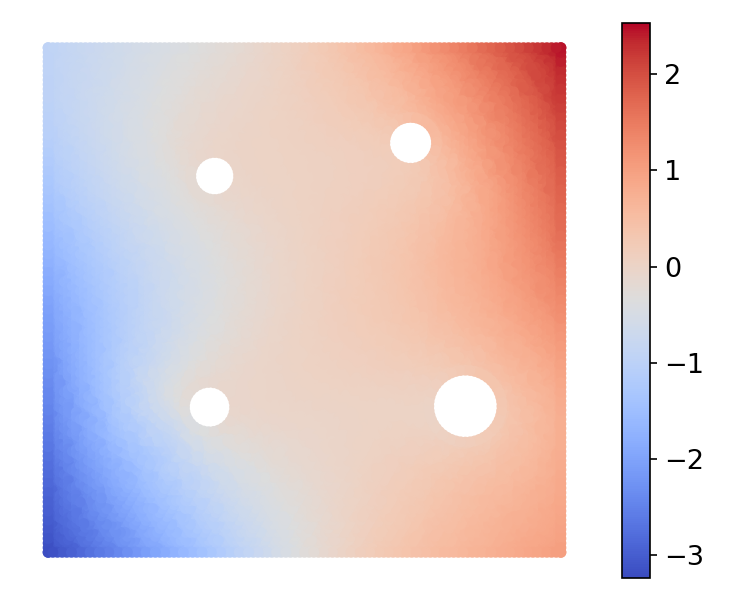}} & {\includegraphics[width=0.16\textwidth]{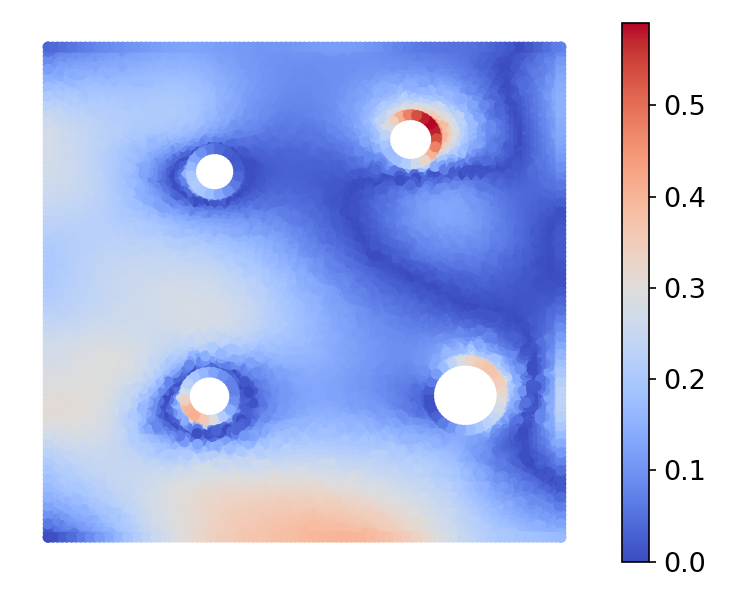}} &{\includegraphics[width=0.16\textwidth]{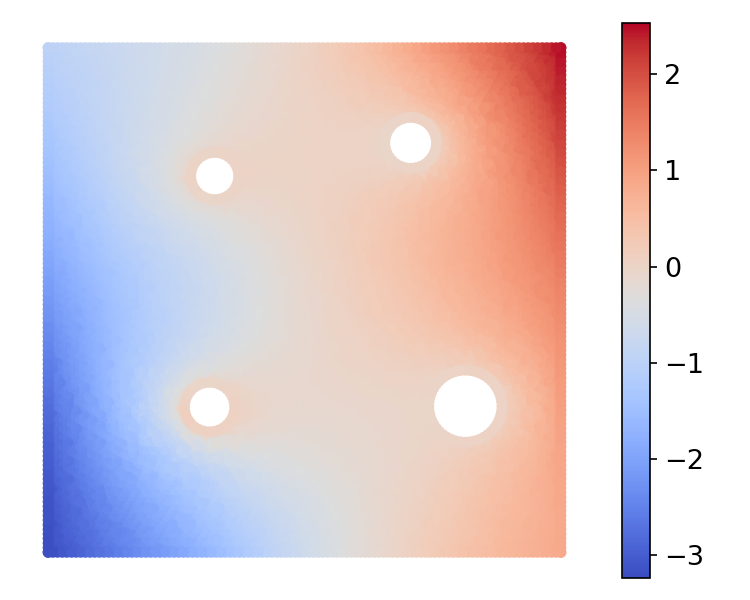}} & {\includegraphics[width=0.16\textwidth]{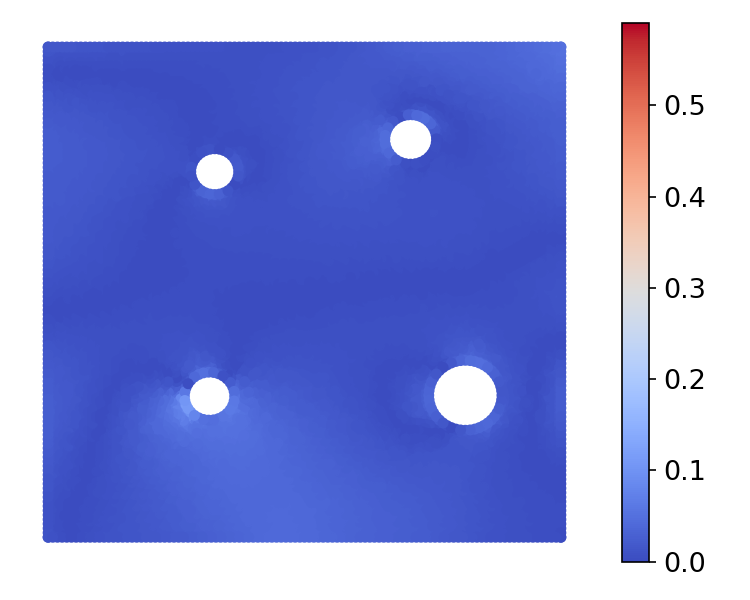}}  \\ 
 \hline
\end{tabular}
\label{fig.best_worst_cases}
\end{table*}

\subsection{Ablation study: Different geometry embeddings \label{Subsec.geo_embed_study}}

In  previous experiments, we used the collocation points on the boundary to infer important features of domain geometry. These collocation points can be collected from different parts of the domain.  In particular, in this section, we study four different ways to do so: (1) collecting collocation points on segments of the boundary that vary, (2) collecting collocation points on all boundary segments, (3) collecting collocation points in the entire domain. As an additional case, we also consider using directly the parametric representation of the geometry, which in our examples are available. 

The parametric geometry representation of the Darcy flow problem consists of the coordinates of the  five points that form the polygon, i.e. a total of 10 coordinates. Similarly, for the 2D plate problem, the parametric  geometry representation contains the locations and sizes of  the four holes, i.e. a total of 12 parameters. For cases where the parametric representation is used, the  geometry encoder will be simply a multi-layer perceptron.

As shown in Table \ref{table.different_geometry_embeddings}, the highest model accuracy is achieved when using collocation points from varying boundaries. Although using all collocation points provides the most comprehensive information about the domain geometry, this approach results in the poorest model performance. This can be due to the fact that many interior point locations are common across different geometries, rendering them redundant in capturing distinct geometric features leading to deterioration in effective geometry encoding. Furthermore, when comparing model performance using points from varying boundaries and parametric representations, the use of varying boundary points leads to 1.7\% accuracy improvement for the Darcy flow problem and 20\% improvement for the 2D Plate problem, indicating that parametric representations is not necessarily always the most effective approach. 

\begin{table}[H]
\begin{center}
\caption{Mean of relative error of different geometry embedding usage.}
\begin{tabular}{c c c c c}
\hline
\makecell{Geometry \\ input data} & \makecell{Interior and \\ boundary points}  & \makecell{Boundary points \\ (fixed and variable)} & \makecell{Parameters of \\ parametric representation} & \makecell{Boundary points \\ (variable only)} \\
\hline
Darcy flow & 11.1\% & / & 8.56\%  & \bf{8.40\%} \\
2D plate  & 21.17\% & 14.81\% & 9.02\% & \bf{7.21\%} \\
\hline
\label{table.different_geometry_embeddings}
\end{tabular}
\end{center}
\end{table}

In our proposed architecture, we used concatenation to combine the geometry embedding with the local coordinate embedding. Here, we study the performance of the model when these two embeddings are integrated using  other forms. In particular, we consider the following three cases
\begin{align}
    \text{Multiplication}: & \quad \bm{H}_j^i = \bm{h}^i_j \odot \bm{G}^i, \\
    \text{Addition}: & \quad \bm{H}_j^i = \bm{h}^i_j + \bm{G}^i, \\
    \text{Concatenation}: & \quad \bm{H}_j^i = [\bm{h}^i_j \| \bm{G}^i],
\end{align}
where $\bm{G}^i$ is the global feature of the $i$-th geometry, $\bm{h}^i_j$ is the local hidden embedding of the $j$-th coordinate in the $i$-th geometry, and $\bm{H}^i_j$ is the global hidden embedding of the $j$-th coordinate in the $i$-th geometry. To generate the results, we only use collocation points on the varying boundary, which was shown earlier to be the best approach. As can be seen in  Table \ref{table.different_geometry_embed_method}, concatenation offers the highest accuracy. 

\begin{table}[H]
\begin{center}
\caption{Mean of relative error of different geometry embedding methods.}
\begin{tabular}{c c c c}
\hline
 \multirow{2}{*}{Problem} & \multicolumn{3}{c}{Operations}\\
\cline{2-4}
 & Multiplication & Addition & Concatenation\\
\hline
Darcy flow & 11.13 \%  & 10.90\% & \bf{8.40\%} \\
2D plate & 15.17\%  & 13.81\% & \bf{7.21\%} \\
\hline
\label{table.different_geometry_embed_method}
\end{tabular}
\end{center}
\end{table}

 {We also investigate the impact of various types of pooling  on the model’s performance. By incorporating boundary points as geometry information and utilizing concatenation operations, we assess the prediction accuracy for different pooling types, as shown in Table \ref{table.different_pooling_layer}. Our findings reveal that employing an average pooling layer results in the slightly enhanced performance, compared to other pooling types. We believe this improvement is because that Average pooling maintains more consistent gradients during back-propagation compared to max-pooling and min-pooling, which can lead to smoother and more stable learning.}

\begin{table}[H]
\begin{center}
\caption{Mean of relative error of different pooling layers.}
\begin{tabular}{c c c c}
\hline
 \multirow{2}{*}{Problem} & \multicolumn{3}{c}{Operations}\\
\cline{2-4}
 & Min-pooling & Max-pooling & Avg-pooling\\
\hline
Darcy flow & 8.49 \%  & 8.44\% & \bf{8.40\%} \\
2D plate & 7.38\%  & 7.27\% & \bf{7.21\%} \\
\hline
\label{table.different_pooling_layer}
\end{tabular}
\end{center}
\end{table}

\subsection{Generalization for large  geometry variation}

In this section, we further explore the geometry generalization using a dataset with significantly higher geometry variation. This dataset is created under the same settings as those described in Section \ref{Subsec.2D_plate}, with the exception that the holes can be of any radius and they can be placed anywhere on the plate, as long as they do not protrude beyond the outer edges. Overlapping holes are also allowed in this data generation, merely to create a more challenging generalization test. To ensure comprehensive coverage of the highly variable geometric distribution, we generate 5,000 different geometry samples. The size ratio of the training, validation  and testing datasets are 70\%, 10\%, and 20\%, respectively. 

As indicated in Table \ref{table.PI_com_high_variation}, our proposed model achieves approximately a 70\% accuracy improvement on this challenging dataset, underscoring its robustness in handling highly varying geometries. By visually comparing the model's performances, Table \ref{fig.best_worst_cases_high_variation} shows that PI-GANO can offer more accurate predictions for various best and worst-case scenarios.

\begin{table*}[!ht]
\begin{center}
\caption{Accuracy comparison between PI-GANO and PI-DCON for the dataset of high geometry variation.}
\begin{tabular}{c c c}
\hline
Model & Mean of relative error & Standard deviation of relative error \\
\hline
PI-DCON & 23.72\% & 10.91\% \\
PI-GANO  & 7.29\% & 3.22\% \\
\hline
\label{table.PI_com_high_variation}
\end{tabular}
\end{center}
\end{table*}

\begin{table*}[!ht]
\centering 
\caption{Comparison between the performance of PI-DCON and PI-DeepONet. Out of all realizations of boundary conditions, the ones that causes best and worst performances of each model are shown.}
\begin{tabular}{|c|c|c | c c | c c|}
\hline
\multicolumn{2}{|c|}{ } & \multirow{2}{*}{Ground Truth} & \multicolumn{2}{c|}{PI-DCON} & \multicolumn{2}{c|}{PI-GANO} \\ 
\cline{4-7}
\multicolumn{2}{|c|}{ } &  & Prediction & Absolute Error & Prediction & Absolute Error \\ 
\hline
\multirow{8}{*}{\rotatebox[origin=c]{90}{2D Plate}} & \rotatebox[origin=l]{90}{\makecell{PI-DCON \\ Best case}} & {\includegraphics[width=0.16\textwidth]{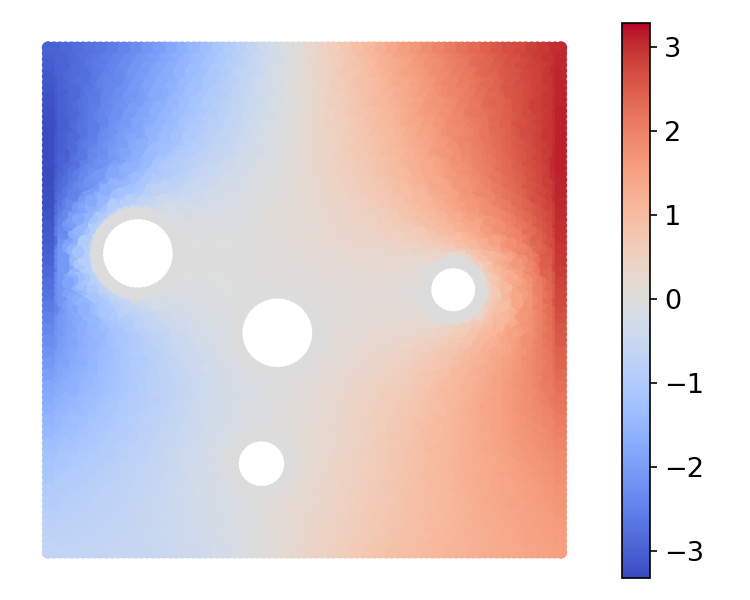}} & {\includegraphics[width=0.16\textwidth]{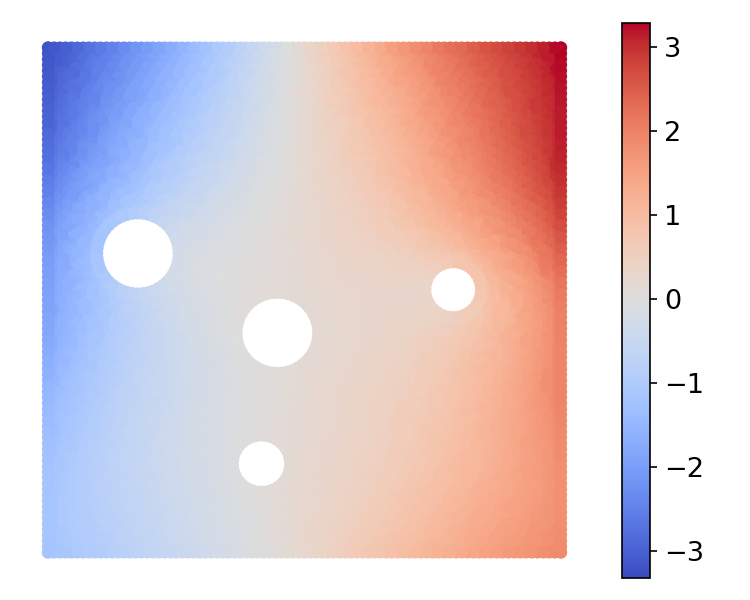}} & {\includegraphics[width=0.16\textwidth]{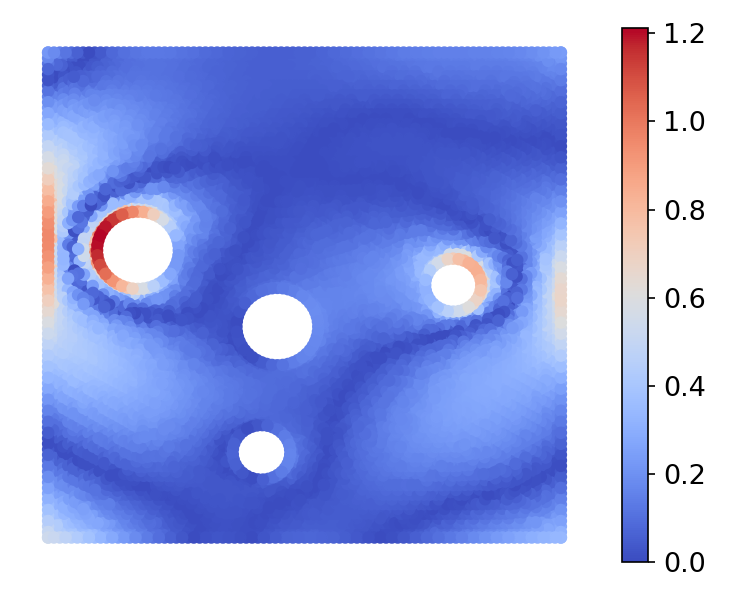}} & 
{\includegraphics[width=0.16\textwidth]{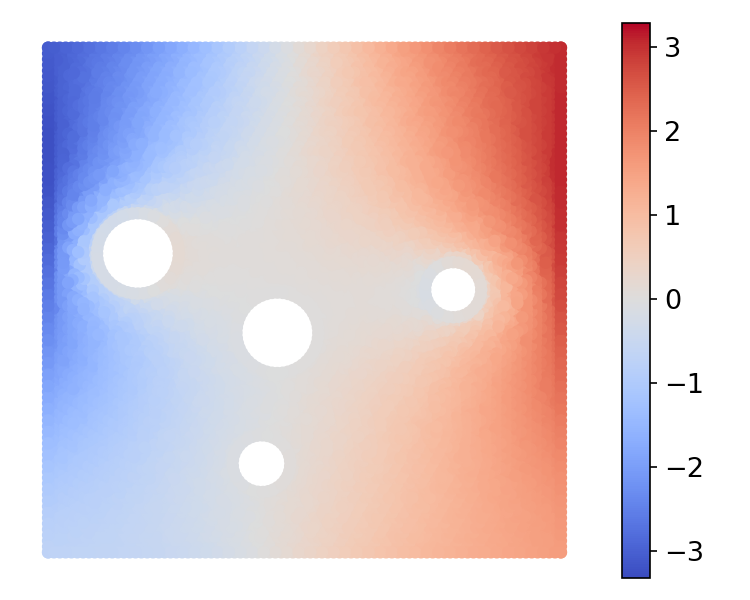}} & {\includegraphics[width=0.16\textwidth]{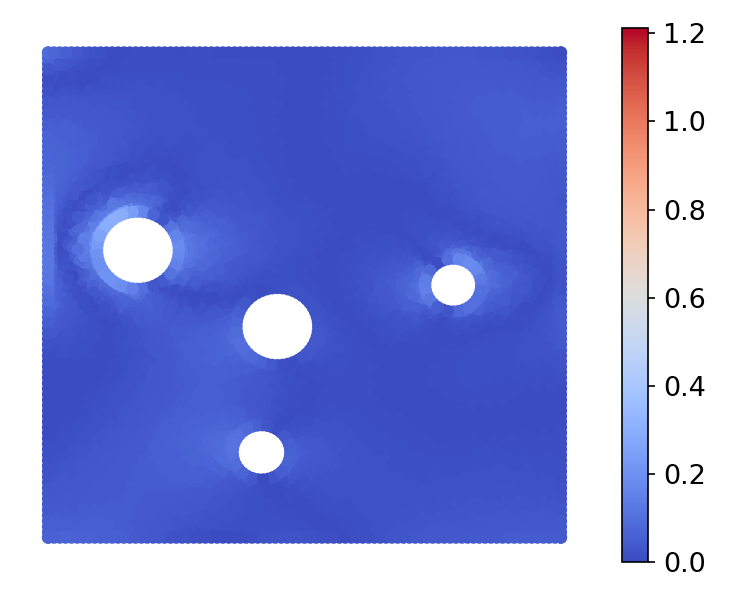}}\\
 & \rotatebox[origin=l]{90}{\makecell{PI-GANO \\ Best case}} & {\includegraphics[width=0.16\textwidth]{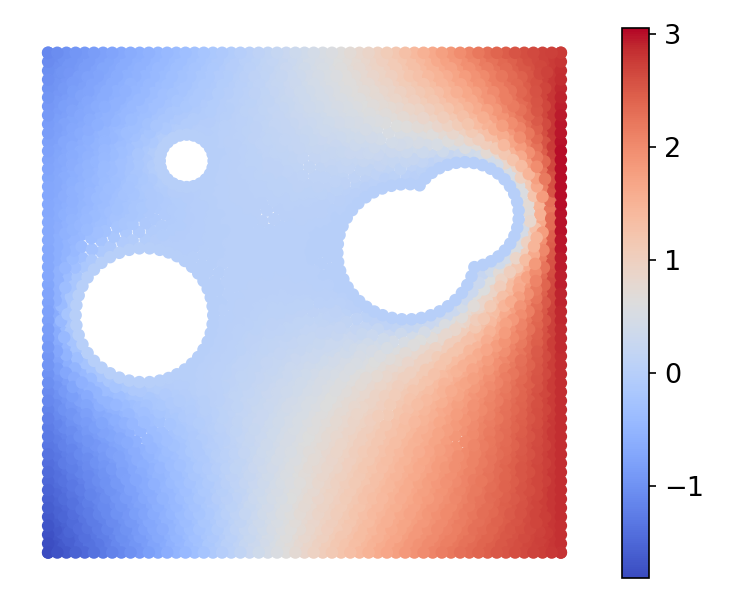}} & 
 {\includegraphics[width=0.16\textwidth]{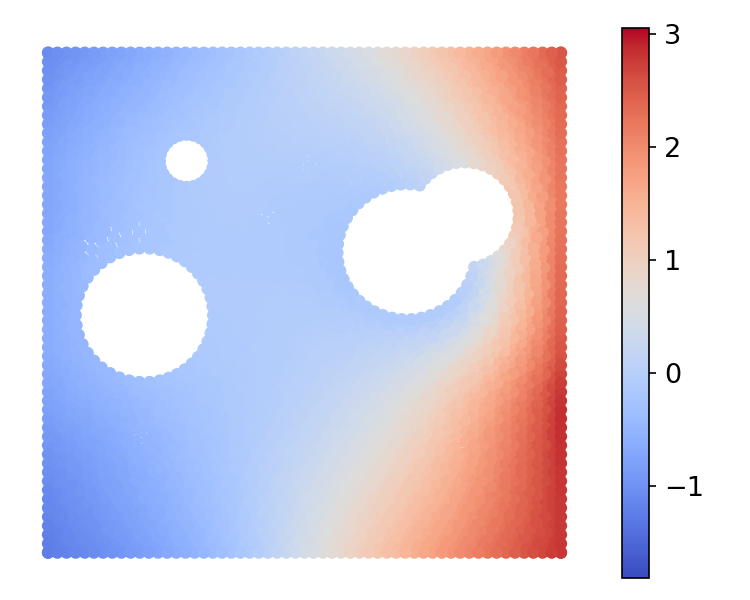}} & {\includegraphics[width=0.16\textwidth]{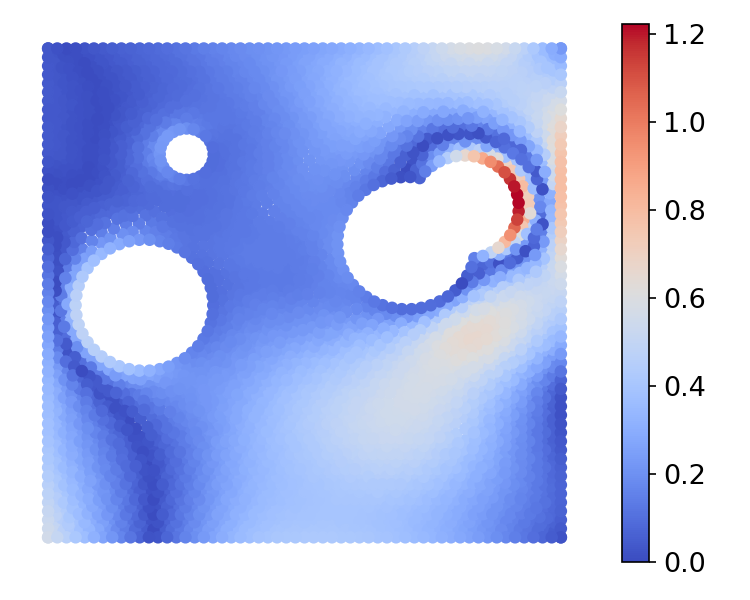}} &{\includegraphics[width=0.16\textwidth]{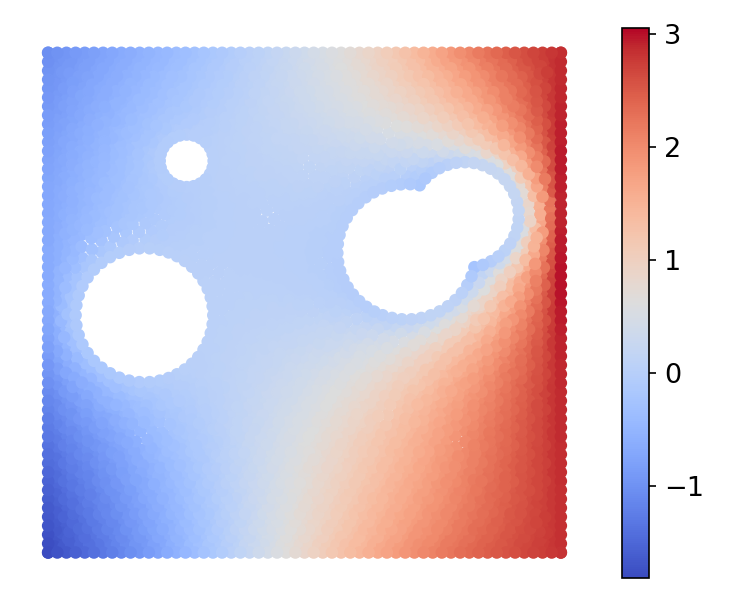}} & {\includegraphics[width=0.16\textwidth]{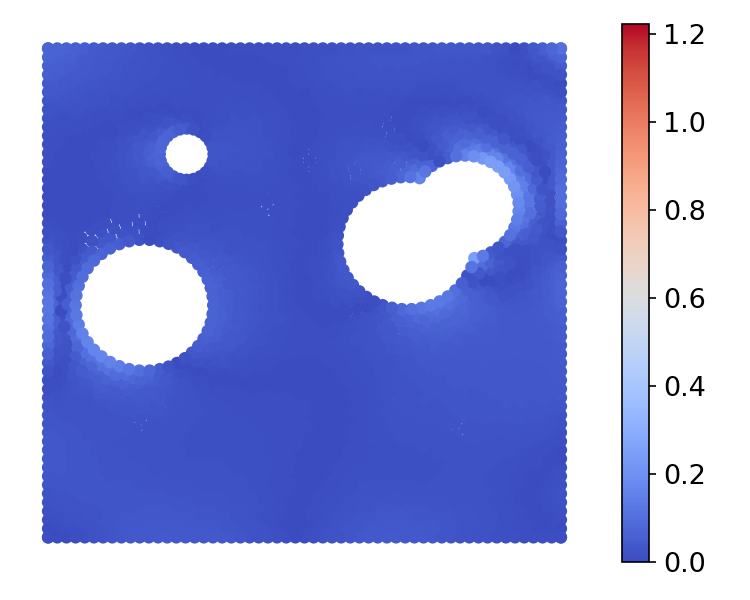}} \\ 
 & \rotatebox[origin=l]{90}{\makecell{PI-DCON \\ Worst case}}& {\includegraphics[width=0.16\textwidth]{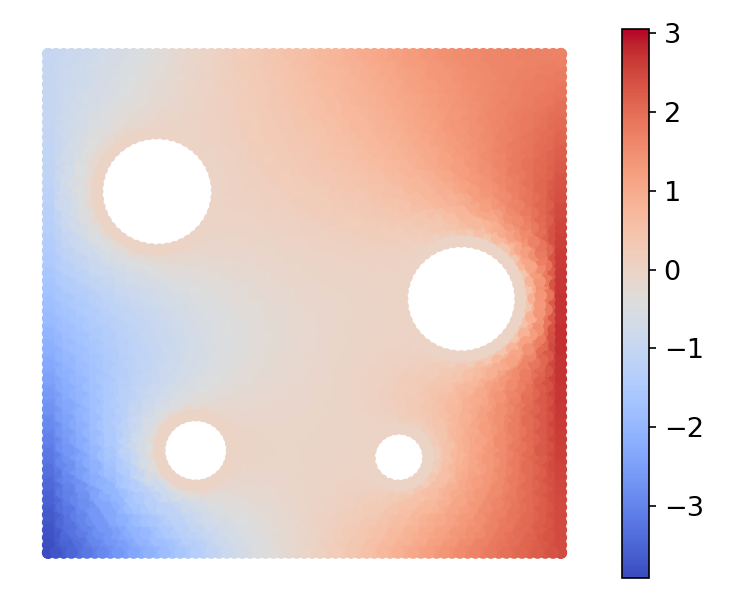}} & {\includegraphics[width=0.16\textwidth]{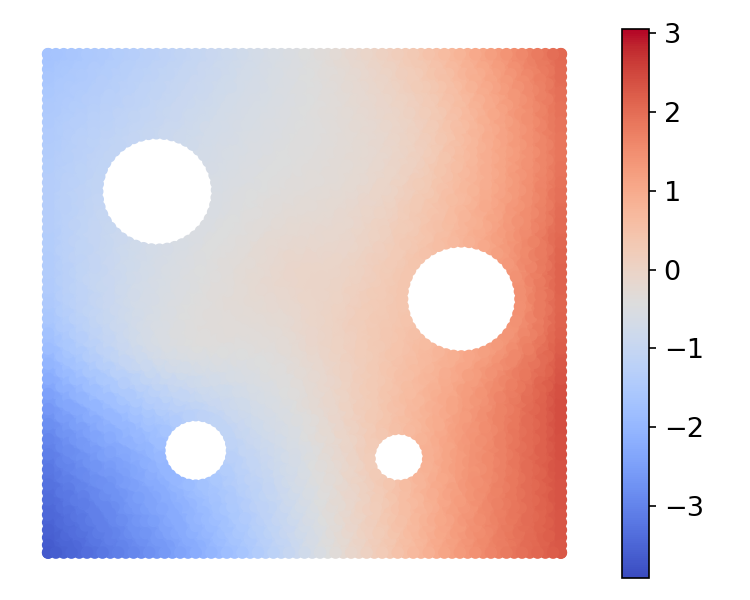}} & {\includegraphics[width=0.16\textwidth]{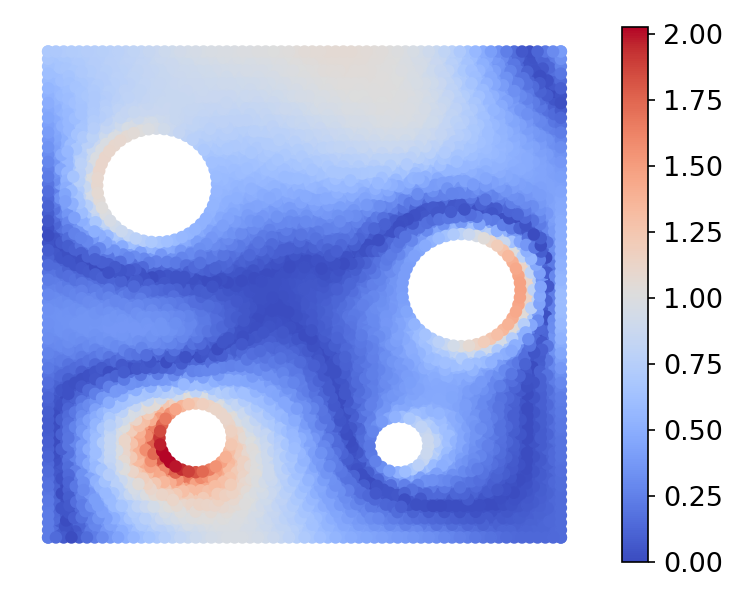}} & {\includegraphics[width=0.16\textwidth]{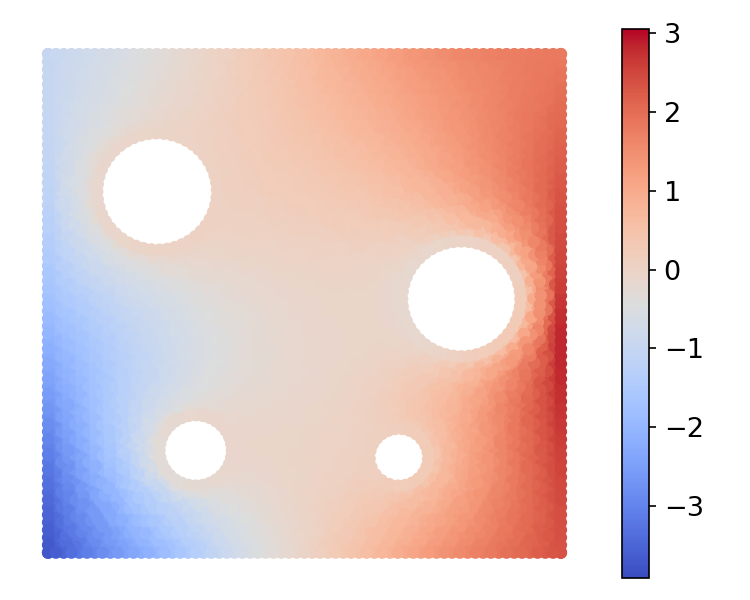}} & {\includegraphics[width=0.16\textwidth]{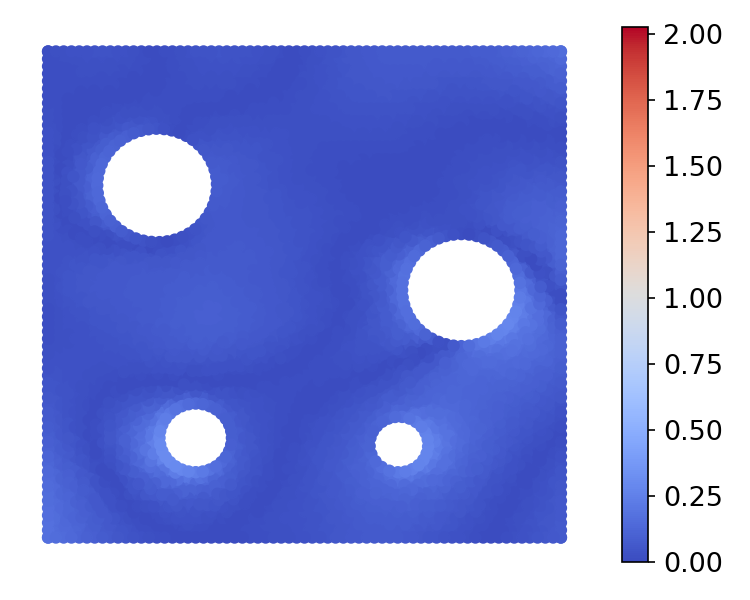}} \\
 & \rotatebox[origin=l]{90}{\makecell{PI-GANO \\ Worst case}} & {\includegraphics[width=0.16\textwidth]{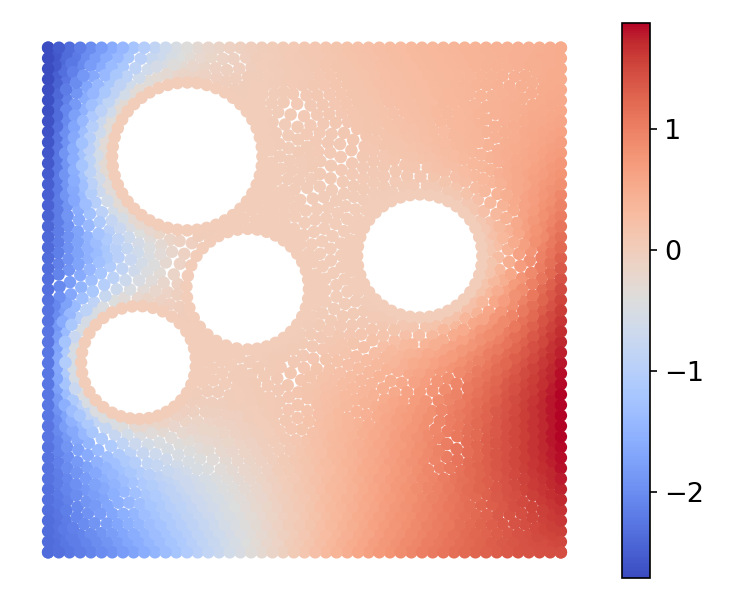}} & 
 {\includegraphics[width=0.16\textwidth]{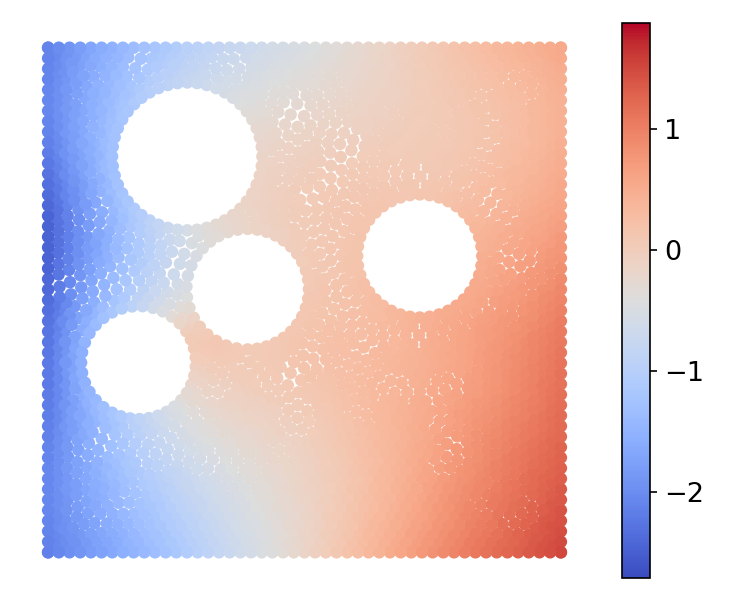}} & {\includegraphics[width=0.16\textwidth]{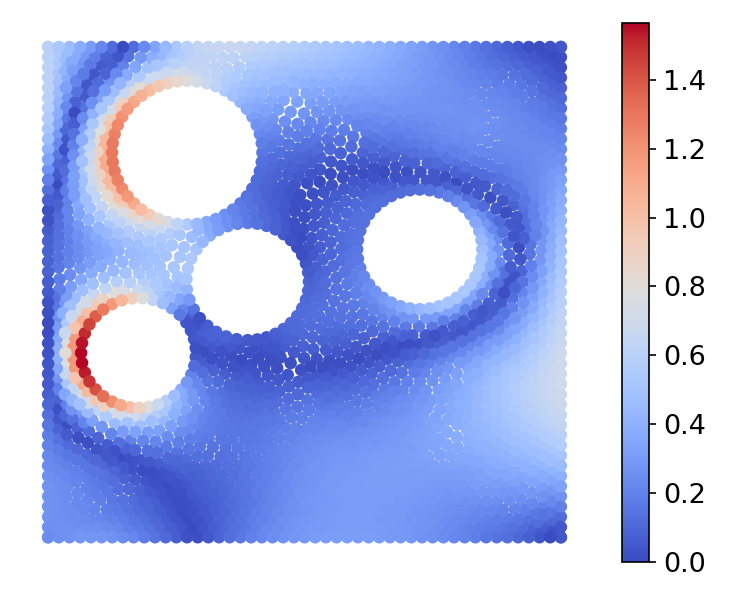}} &{\includegraphics[width=0.16\textwidth]{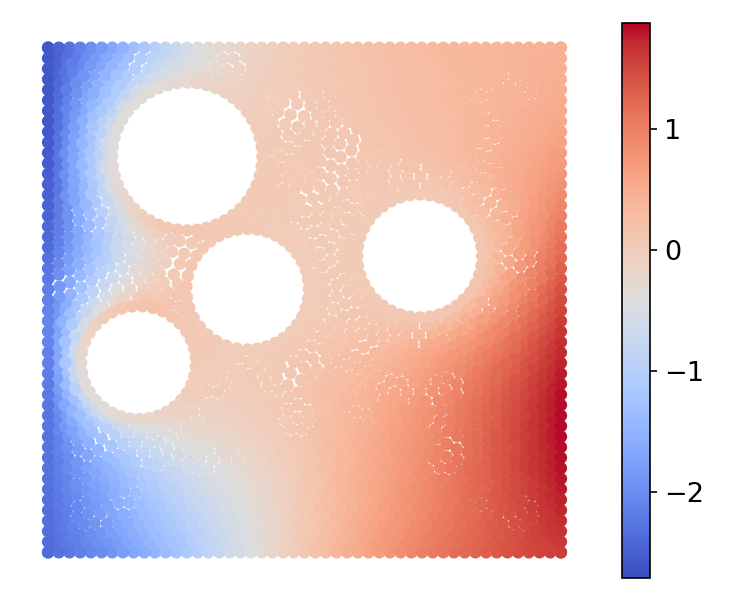}} & {\includegraphics[width=0.16\textwidth]{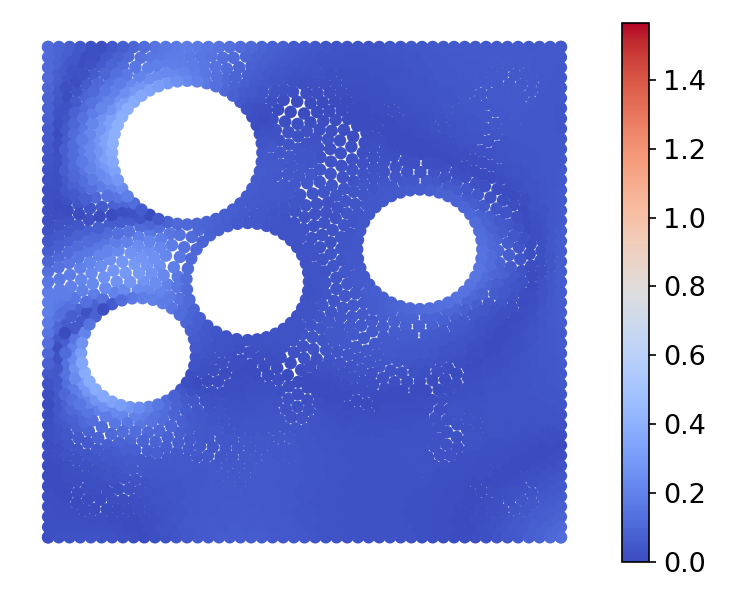}}  \\ 
 \hline
\end{tabular}
\label{fig.best_worst_cases_high_variation}
\end{table*}

 {To evaluate the model's capability to handle geometry test samples that are differently distributed compared to the training data, we performed several experiments using various training-testing dataset pairs. These includes transfer learning cases where the model was trained on low-variation geometry samples and tested on high-variation geometry samples, and vice versa, as well as another ensemble case where the model was trained on a set which consisted of both low- and high-variation geometry samples. As shown in Table \ref{table.shift_test_dist}, the model fails to provide accurate predictions when the training and testing datasets come from two distinct distributions. This is while training the model with both datasets leads to performance improvements. However, the performance is still inferior compared to the case where both training and testing data are from the same distribution, indicating that the model has limitations in handling out of distribution geometry variations.}

\begin{figure}[!ht]
 \centering
 \begin{subfigure}[b]{0.33\textwidth}
    \includegraphics[width=5.3cm]{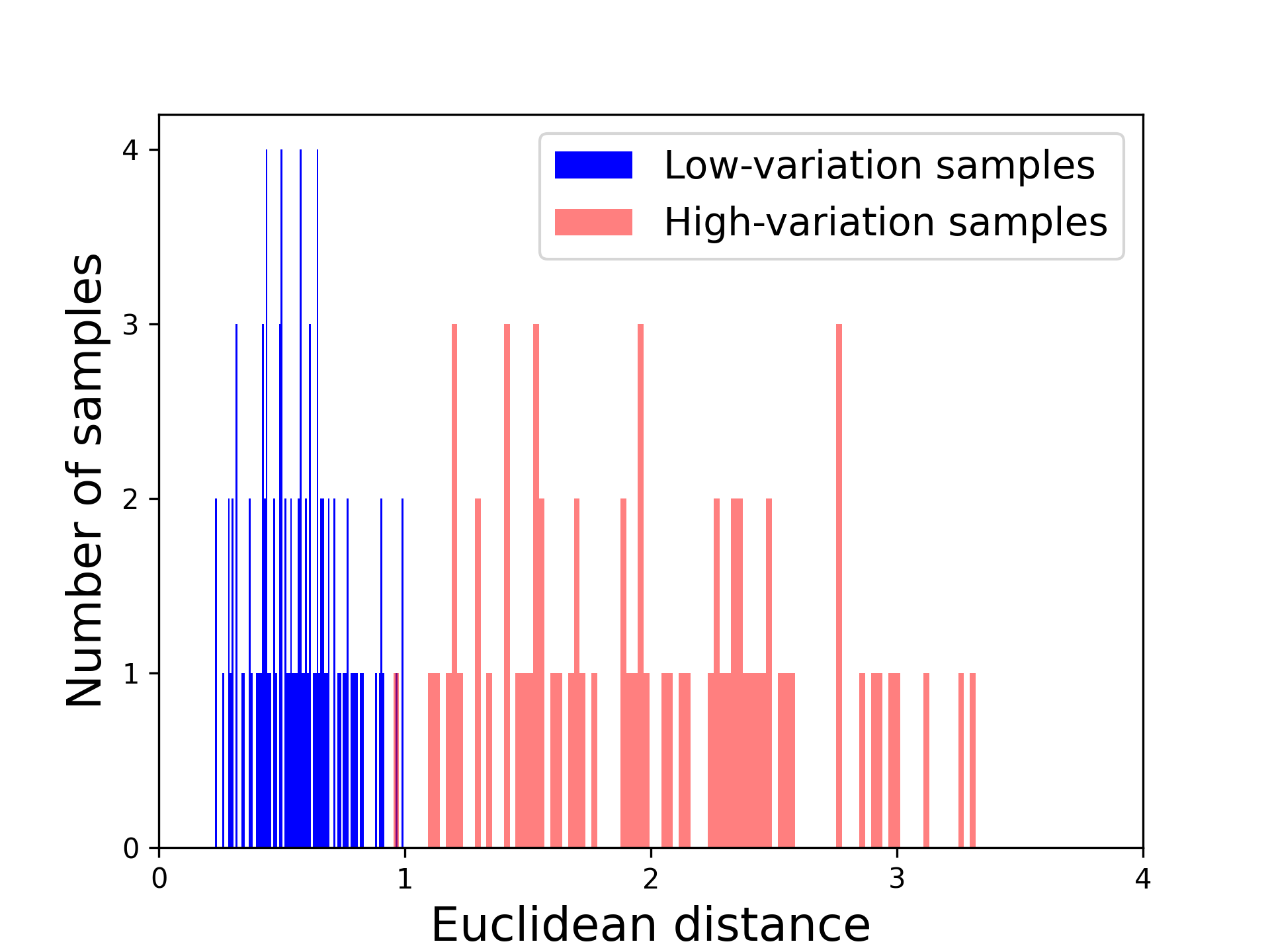}
    \caption{Case 1}
 \end{subfigure}
 \begin{subfigure}[b]{0.33\textwidth}
    \includegraphics[width=5.3cm]{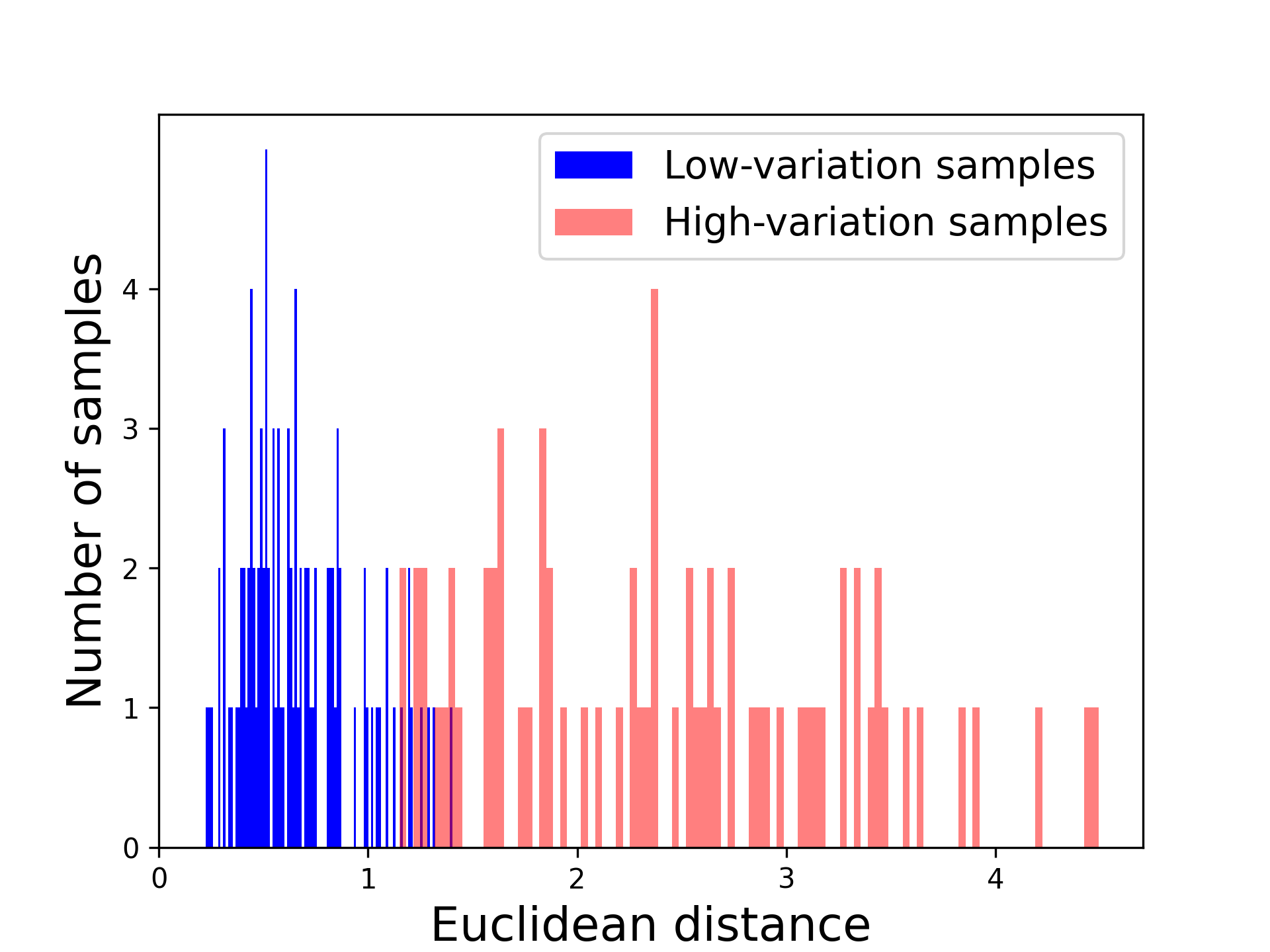}
    \caption{Case 2}
 \end{subfigure}
 \begin{subfigure}[b]{0.33\textwidth}
    \includegraphics[width=5.3cm]{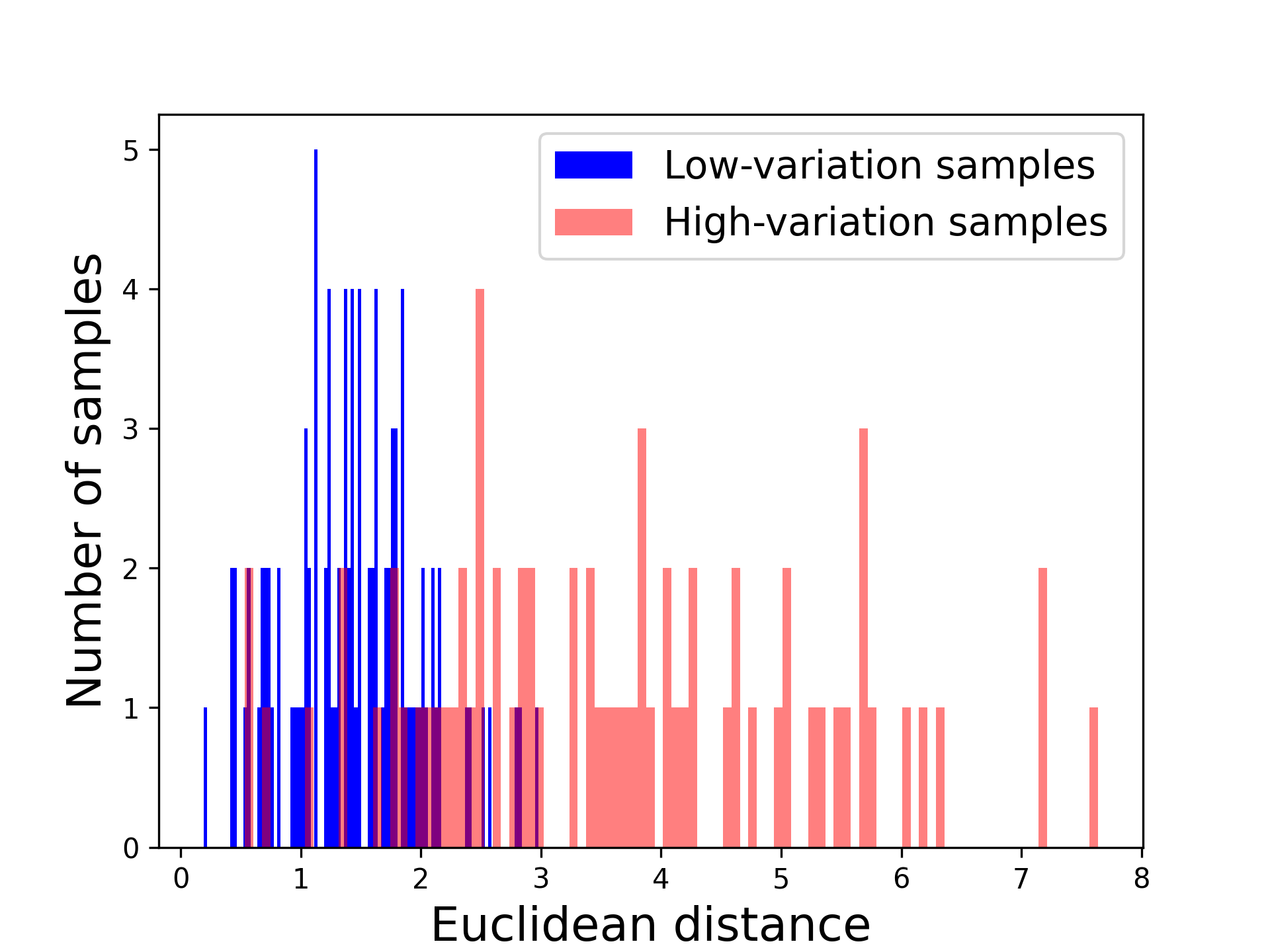}
    \caption{Case 3}
 \end{subfigure}
 \caption{\footnotesize The histograms of Euclidean distance between the embedding vector of various geometry samples. The distance is measured with respect to the  center of cluster of embedding vectors corresponding to the low-variation geometry samples. The three cases are defined in Table~\ref{table.shift_test_dist}.}
 \label{fig.embedding_studies}
\end{figure}

 {We also visualize the differences between the two geometry distributions that we studied, by analyzing their geometry embeddings. In Figure \ref{fig.embedding_studies}, we show the distance of the hidden embeddings of both low-variation and high-variation geometry samples, measured from the  center  of the cluster of low-variation geometry samples. Three different training-test cases shown in Table~\ref{table.shift_test_dist} were used. In Cases 1 and 2, it is evident that the average distance of the high-variation samples is nearly four times as that of the low-variation samples, emphasizing that the two geometry distributions are significantly different. Even when the model is trained on both low-variation and high-variation samples (Case 3), we can still see the significant difference between the learned embeddings of low-variation and high-variation samples.}

\begin{table*}[!ht]
\begin{center}
\caption{Accuracy comparison of PI-GANO between using low-variation training dataset and high-variation training dataset.}
\begin{tabular}{c c c c c}
\hline
{} & Training set & Test set & Mean of relative error & Standard deviation of relative error \\
\hline
 Case 1 & 500 low-var & 100 high-var & 59.72\% & 32.91\% \\
 Case 2 & 500 high-var & 100 low-var & 47.12\% & 22.91\% \\
\multirow{2}{*}{Case 3} & 500 low \& high & 100 low-var  & 23.8\% & 6.4\% \\
{} & 500 low \& high & 100 high-var  & 33.8\% & 12.5\% \\
\hline
\label{table.shift_test_dist}
\end{tabular}
\end{center}
\end{table*}

\subsection{Comparison with data-driven neural operators}

In this section, we evaluate the efficiency and accuracy of our proposed physics-informed model compared to the data-driven (supervised) versions of DCON and GANO architectures.  {In terms of inference time, solving a Darcy problem using FEM takes approximately 1 second, while a 2D plate stress problem requires 2.1 seconds. In contrast, predicting the PDE solutions with the neural operator takes only 0.0021 seconds and 0.0038 seconds, respectively, demonstrating an efficiency improvement of three orders of magnitude. However, beyond inference efficiency, our primary focus is on the computational cost associated with training the models and how a physics-informed can be advantageous.}

\begin{figure}[ht]
    \centering
            \begin{subfigure}[b]{0.45\textwidth}
            \centering
            \includegraphics[width=\textwidth]{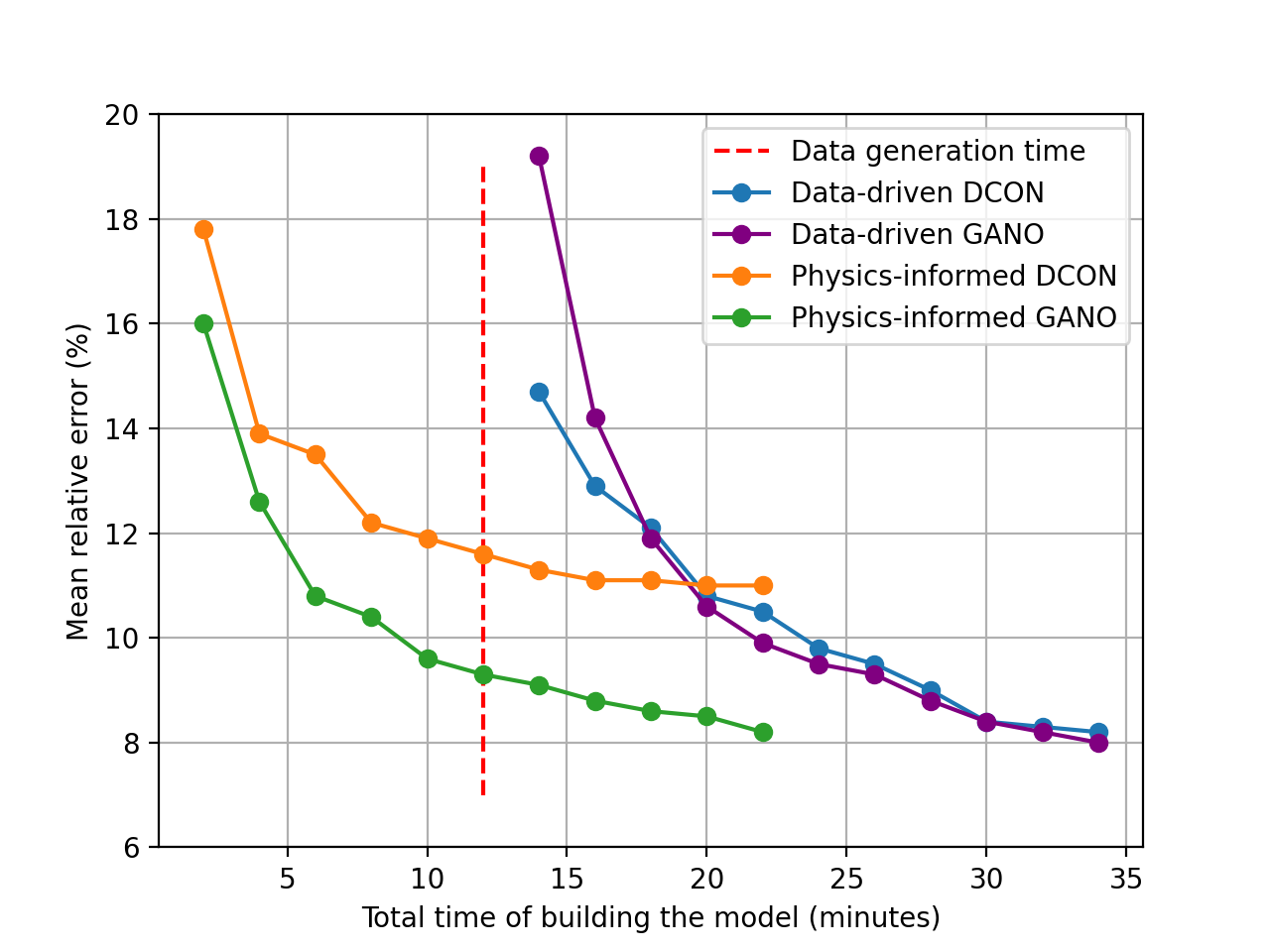}
            \caption{Darcy flow problem}    
            \label{fig:compare_DD_darcy}
        \end{subfigure}
        \hspace{0.1cm}
        \begin{subfigure}[b]{0.45\textwidth}  
            \centering 
            \includegraphics[width=\textwidth]{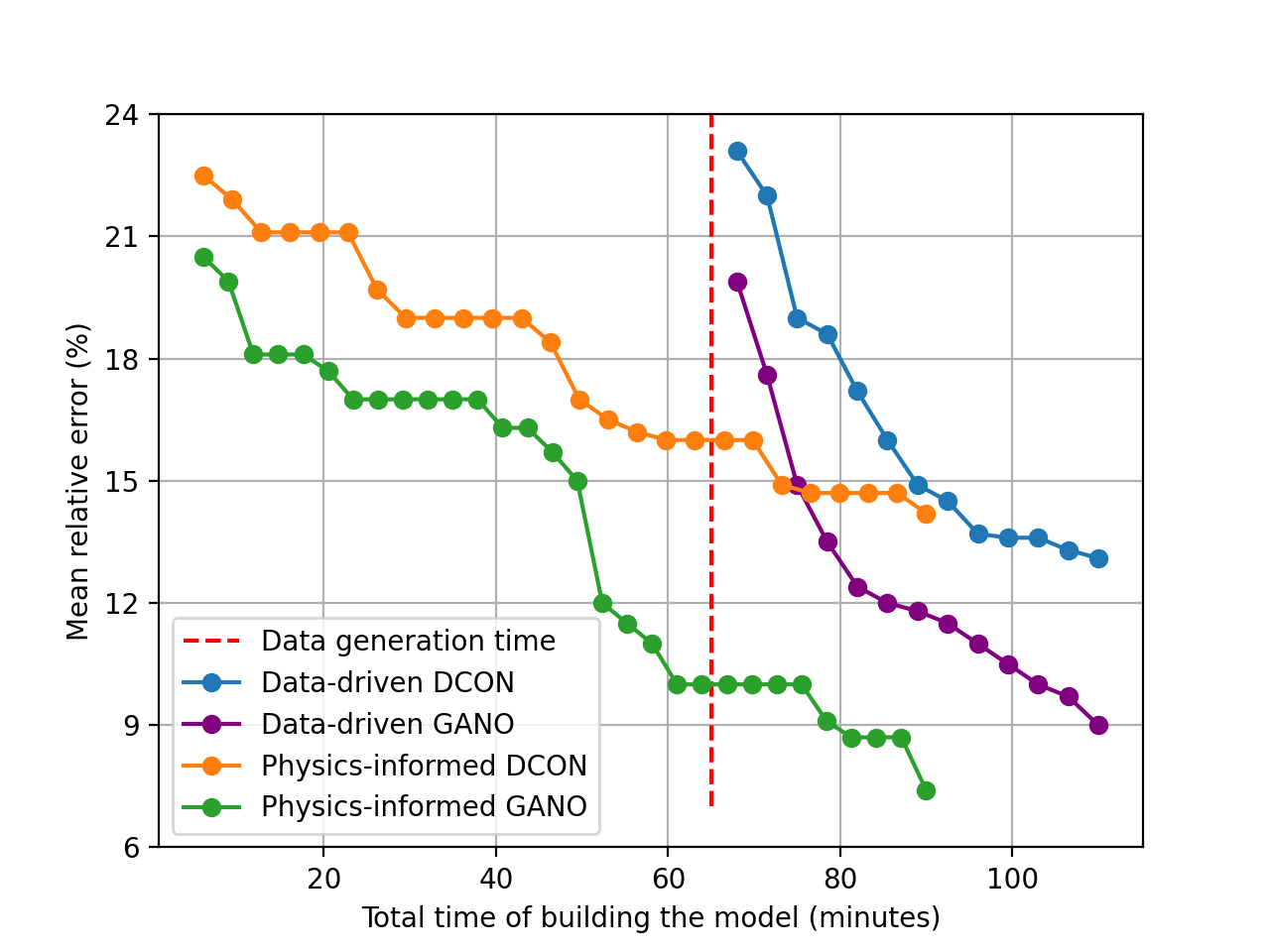}
            \caption{2D plate problem}    
            \label{fig:compare_DD_plate}
        \end{subfigure}
    \caption{\footnotesize Comparison between the total time needed to train different neural operators. For data-driven method, this time also includes the time needed to generate the training (FE simulation) data.}
    \label{fig.eff_acc_com}
\end{figure}

We use the same dataset of 500 samples, as in our previous experiments, with 70\% allocated for training, 10\% for validation, and 20\% for testing. The total time for FEM data generation is estimated based on 350 FEM computations.  We report the accuracy of the two data-driven models in Table \ref{table.data_driven_com}, where the proposed GANO architecture offers superior accuracy over  DCON. We also present the best model performance achieved versus the total time of building the model in Figure \ref{fig.eff_acc_com}. For the data-driven models, the total time includes the data generation and training times. It can be seen that for the Darcy flow problem, PI-GANO reaches a convergence accuracy comparable to that of data-driven GANO and DCON, while requiring  less time to achieve this accuracy level. This underscores the efficiency benefits of using physics-informed training to develop neural operators. Additionally, data-driven GANO and data-driven DCON exhibit similar prediction accuracy levels. We attribute this to the fact that the variable PDE parameter, which are the boundary conditions, in the Darcy flow problem, are imposed on the entire exterior boundary of the domain. This means that the PDE parameter input  also inherently contains geometry information about the entire geometry. Therefore, the addition of a separate geometry encoder does not significantly enhance the model performance.

On the other hand, for the 2D plate problem, data-driven GANO achieves over a 28\% accuracy improvement compared to data-driven DCON. This shows it is important to include a separate geometry encoder, because the variable  parameters in this problem are only imposed on  a subset of boundary segments, which do not fully characterize the entire domain.  Furthermore, it can be seen in this figure that our proposed PI-GANO outperforms the data-driven version of GANO, highlighting the potential of physics-informed training to offer efficiency and also to enhance prediction accuracy when the same model architecture is used. 

 {An interesting observation is that PI-GANO outperforms the data-driven GANO in the 2D plate stress problem. Typically, we would expect data-driven training to yield better performance than physics-informed training when using the same model architecture, as it is a more straightforward and easily implementable approach. However, as shown in works related to PI-DeepONet \cite{PI-deepOnet, improved_don}, physics-informed models can sometimes surpass data-driven models when the model architecture and training algorithm are appropriately combined. Additionally, this comparison is constrained by equal training times, suggesting that the data-driven model might outperform the physics-informed model with extended training time. }

We also observe that, in the 2D plate problem, the best model prediction accuracy of the physics-informed models does not always decrease with more training epochs. Instead, it may remain unchanged during parts of the training process. This phenomenon can be attributed to the small size of the collocation point sampling used for PDE residual computation to handle meshes of high resolutions. Using a small set of points for PDE residual computation can cause the model to overly focus on the accuracy of a small region in the domain, at the expense of overall approximation accuracy across the entire domain. This issue highlights the need for more advanced physics-informed training algorithms to enhance the training of neural operators

\begin{table*}[!ht]
\begin{center}
\caption{Accuracy comparison between data-driven DCON and data-driven GANO.}
\begin{tabular}{c c c c c}
\hline
\multirow{2}{*}{Dataset} & \multicolumn{2}{c}{\makecell{Mean of relative error}} & \multicolumn{2}{c}{\makecell{Standard deviation of relative error}} \\
\cline{2-5} 
& \makecell{Data-driven \\ DCON} & \makecell{Data-driven \\ GANO} & \makecell{Data-driven \\ DCON} & \makecell{Data-driven \\ GANO}\\
\hline
\makecell{Darcy flow} & 8.21\% & \bf{8.00\%} & 3.27\% & \bf{2.93\%}\\
\makecell{2D plate} & 13.5\% & \bf{9.01\%} & 4.56\% & \bf{3.17\%} \\
\hline
\label{table.data_driven_com}
\end{tabular}
\end{center}
\end{table*} 

\subsection{Hyper-parameter studies}

\begin{figure}[ht]
    \centering
            \begin{subfigure}[b]{0.4\textwidth}
            \centering
            \includegraphics[width=\textwidth]{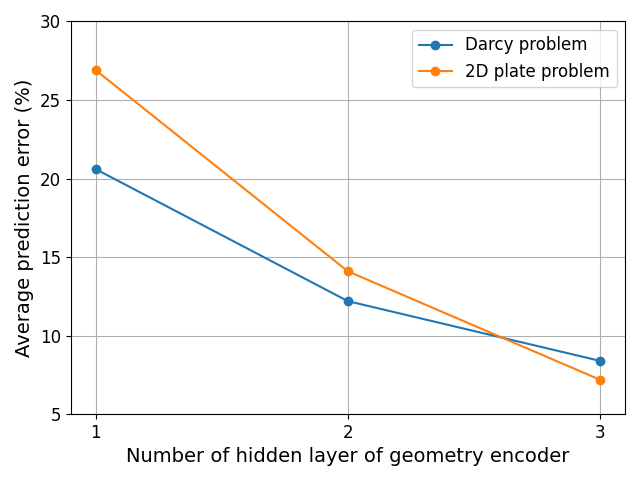}  
            \label{fig:hyper_geo_hidden}
        \end{subfigure}
        \hspace{0.1cm}
        \begin{subfigure}[b]{0.4\textwidth}  
            \centering 
            \includegraphics[width=\textwidth]{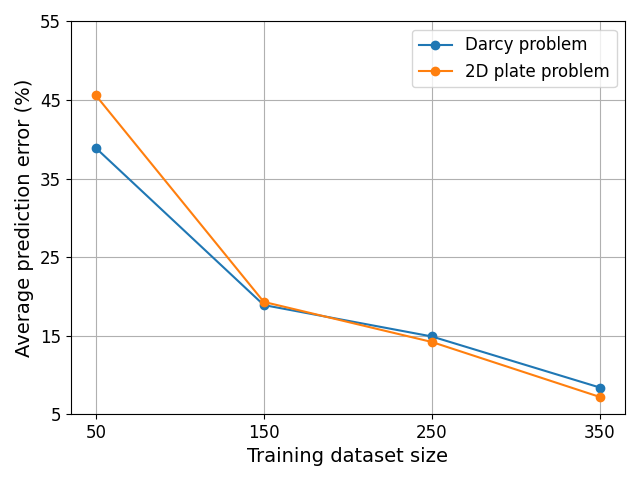}  
            \label{fig:hyper_datasize}
        \end{subfigure}
    \caption{\footnotesize Hyper-parameter studies of PI-GANO architecture.}
    \label{fig.hyper_studies}
\end{figure}

 {In this section, we analyze the impact of various hyperparameters on the model's performance in terms of geometry generalization. Specifically, we focus on the number of hidden layers in the geometry encoder and the size of the training dataset. As illustrated in Figure \ref{fig.hyper_studies}, model performance improves with increasing complexity of the geometry encoder and a larger training dataset. This result is intuitive, as a more complex geometry encoder allows for better feature extraction from the domain geometry, and a larger dataset provides the model with more opportunities to learn and generalize. However, we also observe a significant drop in performance when the training dataset is insufficient. For instance, when the number of training samples decreases from 150 to 50, the model's average prediction error rises from approximately 17\% to 40\%.}

\section{Conclusion}\label{sec.conclusions}
In this study, we introduced PI-GANO, a physics-informed neural operator inspired by PI-PointNet and PI-DCON, marking the first attempt to construct a neural operator capable of generalizing to both variable PDE parameters and variable domain geometries without the need for FEM training data or a regular meshing of domains. Our results demonstrate that PI-GANO can effectively predict PDE solutions for variable boundary conditions and domain geometries. Moreover, comparing the proposed PI-GANO with the data-driven version of our proposed architecture, we showed the potential of physics-informed training to offer efficiency and accuracy benefits in developing neural operators.

Despite these advancements, our approach has its limitations. Primarily, our evaluation focuses solely on the model's capability to compute steady-state solutions. Further research is needed to adapt PI-GANO to handle dynamic responses, thereby enabling it to address time-dependent PDEs. Additionally, we observed instability in physics-informed training, which leads to requiring more training epochs to identify optimal model parameters. To address this, we seek to adopt a more intelligent collocation point sampling strategy, importance sampling \cite{importance}, for physics-informed training. Moreover, this approach can be investigated in more challenging problems, such as those with cracks and highly nonlinear response.

 {Furthermore, the current model is restricted to solving PDE problems within a specific underlying distribution of geometries and PDE parameters, as is common in many design problems with limited variation. However, to tackle scenarios where domain geometries and problem settings display greater variability or multiple varying PDE parameters, future work will focus on developing a more robust foundation model capable of generalizing across diverse distributions of design parameters. This will require designing a more advanced architecture with increased complexity and capacity, while also supporting physics-informed training. For example, using multiple PDE parameter encoders to capture different PDE parameter information will be one possible improvement. To address more complex problems, we also advocate for the use of advanced physics-informed training algorithms to enhance the training of existing complex data-driven neural operators.}

\bibliographystyle{plainnat} 

\bibliography{references}  

\end{document}